\documentclass[a4, 12pt]{article}
\pdfoutput=1

\usepackage{array}
\usepackage{verbatim}
\usepackage{pifont}
\usepackage{calc}
\usepackage{siunitx}
\usepackage{url}
\usepackage{graphicx}
\usepackage{amsmath}
\usepackage{chngpage}
\newcommand{\xmark}{\ding{55}}%
\newcommand{\picheight}{3.2} 
\newcommand{\widepicheight}{2.5} 
\newcommand{\hardpicheight}{3.8} 
\newcommand{\spherepicheight}{4.0}
\newcommand{\monkeypicheight}{10.0}

\newcommand\blfootnote[1]{%
  \begingroup
  \renewcommand\thefootnote{}\footnote{#1}%
  \addtocounter{footnote}{-1}%
  \endgroup
}

\renewcommand{\vec}[1]{\boldsymbol{\mathrm{#1}}}
\newcommand{\mat}[1]{\boldsymbol{\mathrm{\MakeUppercase{#1}}}}
\usepackage{subfigure}
\title{Tree Reconstruction using Topology Optimisation}
\author{Thomas Lowe$^{1,\dagger}$ and Josh Pinskier$^{2,\dagger}$\\
}
\date{%
    \footnotesize
    $^1$CSIRO Data61, Brisbane Australia\\ thomas.lowe@csiro.au\\ ORCID 0000-0001-8932-018X\\%
    $^2$CSIRO Data61, Brisbane Australia\\ josh.pinskier@csiro.au\\ ORCID 0000-0002-8878-9012\\
    $\dagger$ Both authors contributed equally to this research.
}

\begin{document}
\maketitle

\begin{abstract}
Generating accurate digital tree models from scanned environments is invaluable for forestry, agriculture, and other outdoor industries in tasks such as identifying fall hazards, estimating trees' biomass and calculating traversability. Existing methods for tree reconstruction rely on sparse feature identification to segment a forest into individual trees and generate a branch structure graph, limiting their application to easily separable trees and uniform forests. However, the natural world is a messy place in which trees present with significant heterogeneity and are frequently encroached upon by the surrounding environment. We present a general method for extracting the branch structure of trees from point cloud data, which estimates the structure of trees by adapting the methods of structural topology optimisation to find the optimal material distribution to interpolate the input data. We present the results of this optimisation over a wide variety of scans, and discuss the benefits and drawbacks of this novel approach to tree structure reconstruction. Our method generates detailed and accurate tree structures, with a mean Surface Error (SE) of 15 cm over 13 diverse tree datasets.
\end{abstract}

\section{Introduction}
\label{sec:introduction}
Automatically converting a 3D point cloud of a natural scene into a geometric description of its trees is important in both digital and outdoor industries. It is valuable in forestry to measure biomass and growth of a managed forest, it is needed by power companies to identify branches encroaching on power lines, and it is valuable in bush fire management and post-fire fall risk assessments. In general, it is important in any form of digital twinning, for an accurate digital clone of vegetated environments.
\begin{figure}[h]
\centering
\includegraphics[width=0.7\columnwidth]{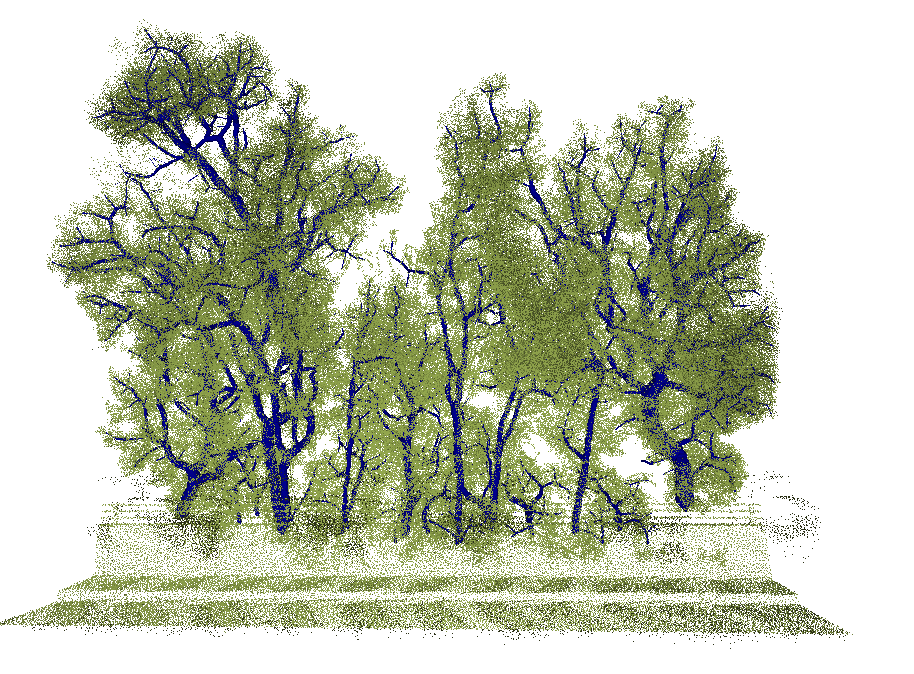}
\caption{A reconstruction of a challenging, cluttered environment where all tree bases are occluded and approximately 2m below the overpass. It shows the overlaid view of point cloud (green) and generated trees (blue)}
\label{fig:overlaid}
\end{figure}

Digital tree reconstruction is challenging because vegetated environments are extremely varied. For example, trees may be tall pines or wide figs, they may be foliated or bare, overlapping or separated, topiarised or unmanaged. The trunks may be obscured by vegetation, and the ground could be sloped or vegetated. In addition to the variety of vegetation, the acquisition can also be varied. 3D point clouds can be collected from ground or aerial lidar, depth cameras or reconstructed from sets of 2D images. Depending on the sensing mode and collection method, the density of points in the point cloud, viewing angle and accuracy can vary significantly. Lastly, dense vegetation often obscures or occludes the scene behind it, resulting in missing branches, trunks and low point density in obscured regions. At its most severe it may be only the outer canopy that is observed, and for aerial data it may be only the upper canopy with no visible trunks. 

This variability makes the task difficult for traditional tree reconstruction methods, which rely on uniquely and accurately identifying common features of trees, such as crowns, trunks, and cylindrical branches. These features are often poorly distinguished or obscured. As a result, robust tree reconstruction in general environments remains elusive, and so leading methods provide tree reconstruction solutions for only special-case environments. Whether that be specialising to conifer forests~\cite{trochta20173d}, to high resolution data~\cite{aiteanu2014hybrid}, or to tree environments with clearly resolved trunks~\cite{du2019adtree}. 

We investigate tackling the tree reconstruction problem from the perspective of topology optimisation, in particular using topology optimisation as a dense reconstruction method. The proposed method acts on all sensor observations equally to interpolate the lidar points with contiguous acyclic tree structures that avoid all observed free space. Rather than searching for sparse features such as crowns, trunks or cylindrical branches, it finds an efficient distribution of material (wood) that matches the input data under a small number of assumptions. These are:
\begin{itemize}
    \item the input data represents ground, low lying vegetation and trees only
    \item trees are solid acyclic structures that connect to the ground, with no gaps
    \item unobserved trees have a similar branching angle
    \item unobserved structures use their material `efficiently', having approximately straight branches
\end{itemize}
The meaning of `efficiently' here depends on the choice of cost function. We employ a minimal and standard topology optimisation cost function: compliance under linear scalar loads. This is described in Sections~\ref{sec:SIMP} and~\ref{sec:problem}.

We believe that this approach makes fewer assumptions about the nature of the data than traditional reconstruction approaches, and that its failures are likely to be less extreme, giving results that are tree-like and connected even where significant trunk and branch data is missing. Reducing the severity of any failures is important because bulk measures such as number of trees, number of branches and biomass, can remain reasonable. 
This robustness is shown in Figure~\ref{fig:overlaid}, in which a set of trees (blue) are extracted from a 3D point cloud (green) using our method. Despite several intertwined trees and a barrier occluding the tree-trunks, the tree branches closely match the measured point cloud and it plausibly estimates unobserved regions. We validate this claim in Sections~\ref{sec:qualitaive} and~\ref{sec:quantitaive}.

To clarify the tree reconstruction problem we first specify the input and output data structures that we will use.  The input data can come from a number of sensors, most commonly lidar (mobile or static, ground based or aerial), but also depth cameras and binocular cameras. For brevity, we will refer to the input sensor as a lidar in this paper as it is the most commonly used solution for these environments. The lidar data is then processed with a registration scheme (such as Simultaneous Localisation and Mapping, SLAM) to transform the observations into a common coordinate frame. The result is a set of 3D points and a set of rays from each sensor location to the observed 3D point. The set of 3D points represent physical surfaces and the set of rays represents free space. The data may also contain rays that have no end point, such as those pointing towards the sky. 

If the set of rays is not provided directly from the registration scheme then it can be generated from the set of 3D points and the calculated trajectory of the sensor, provided that they both contain time stamps in order to find the corresponding ray start and end locations. For static lidars there may be no explicit trajectory provided, but the origin of each scan is typically available, so the set of rays can also be constructed in this case. 

The set of all of these rays and points represents the principle input data that has been observed from the scene, and is referred to as a ray cloud, as opposed to a point cloud, which is only the set of points. 

The output data is a piecewise-cylindrical acyclic graph, representing the surface geometry of the trees in the scene. This Branch Structure Graph (BSG) is the set of nodes defined by: a 3D node location, a reference to the parent node, and a radius for the cylinder between the node location and the parent node's location. The root node of each tree contains no parent. Unlike in~\cite{livny2010automatic} we allow the graph to be disconnected, so a single graph represents the set of trees, and so the output of our method is a single BSG.

The problem that we address in this paper is how to robustly transform a ray cloud of a natural scene into a BSG representation of its trees. To the best of the authors' knowledge this paper presents the first time use of topology optimisation as a method for physics-informed geometric reconstruction.
Our main contributions are:
\begin{enumerate}
    \item The development and investigation of a novel tree reconstruction method based on topology optimisation, which allocates biomass to minimise a specified cost function
    \item The development of a branch structure generation algorithm, which converts the optimised voxel trees into a graph of their branch structures
    \item Its application to a dataset of realistic natural scenes chosen to assess common failure cases systematically
\end{enumerate}

\section{Existing Work}
\subsection{Tree Reconstruction}
The most common approach to tree structure reconstruction from point clouds is a two stage approach. The first stage segments the data into individual trees, and the second stage extracts the branches of a single-tree point cloud. 

The segmentation phase has many variants, which reflect the variety of features that may be used to separate trees. Trunk identification methods~\cite{li2010segmentation,zhong2016segmentation,liu2021point,burt2019extracting} look for near-vertical cylindrical shapes in the point cloud as the basis for locating the individual trees. Crown methods~\cite{li2012new} look for conical or dome-shaped crowns within the point cloud. More recently, gap-based methods~\cite{ayrey2017layer,heinzel2018constrained} use spectral clustering or agglomerative clustering~\cite{wang2018scalable} to separate trees based on any gap or thinning of foliage between adjacent trees. Lastly, branch direction methods~\cite{luo2021individual} look at average branch directions in order to estimate a central trunk location, and assign trunk ownership to each branch in a forest point cloud.

These methods are effective in special cases, such as managed pine forests, or sparse woodland. But in general point clouds, each feature has many failure cases: trunks are typically unobserved from aerial data, crowns are poorly observed from ground-based scanning, and gap-based methods lose reliability on densely overlapping trees. Lastly, branch direction methods fail on highly foliated trees, where the branches are occluded by leaves. 
Deep learning based methods have recently emerged that are able to semantically segment point clouds with high accuracy in woody forests \cite{Krisanski2021}. However these require large amounts of labelled training data to perform well, and have not been shown to generalise to more complex and diverse environments.

There are branch reconstruction methods available that construct the tree skeleton using medial axis methods~\cite{tagliasacchi20163d,schilling2014automatic}, thinning~\cite{jiang2020skeleton}, region growing~\cite{raumonen2013fast} and cylinder slices~\cite{aiteanu2014hybrid}, see~\cite{wang2018lidar} for a comparative review. The most prominent of these is the distance segmentation method~\cite{tagliasacchi20163d, hu2017efficient} first described in 2007~\cite{xu2007knowledge}. This method finds the shortest path tree from root to all cloud points, and groups each point by path distance from root. The result is a set of approximately cylindrical sections, which can be converted into the tree structure (BSG) nodes. This general approach is available in the software treeQSM~\cite{treeQSM}, AdTree~\cite{du2019adtree}, and AdQSM~\cite{fan2020adqsm} which is based on AdTree. These methods require well-sampled branches in order for the shortest path algorithm to connect the points appropriately, and a well-defined trunk in order to ascertain the root position of the path. Moreover, these and the other cited branch reconstruction algorithms require the point cloud of a single tree. They therefore depend on tree segmentation and so inherit the discussed failure cases of the existing segmentation methods.

\subsection{Topology Optimisation}
Topology optimisation (TO) is a computational design method which finds the material distribution that minimises a specified cost function. 
Typically it is used to generate stiff, lightweight structures, such as bridges and mounting brackets. However, it has been applied to a vast range of problems across numerous physical domains, including thermal conductivity problems (heat sinks) \cite{Ikonen2018b}, compliant mechanisms \cite{Pinskier2020,Pinskier2018b,Clark2018}, soft robots \cite{Pinskier2022,Zhang2018f}, electrothermal actuators \cite{Sigmund2001}, fluidic actuators \cite{Kumar2020}, and fluidic mixers \cite{Munk2017}, and biomimetic design \cite{Zhang2022a}.
Regardless of application, the design space is parameterised by a set of elements, loads, and boundary conditions which enables their governing physical equations to be evaluated using a finite element solver. Iteratively evaluating a candidate design and updating it with a gradient-based or heuristic non-gradient solver then allows a high-performing design to be identified. 
For a detailed review of topology optimisation methods and applications see~\cite{Bendsoe2003,Zhang2018e,Sigmund2013,Munk2015,VanDijk2013}.

Compared to the previously discussed methods, topology optimisation offers a generalised, dense approach to tree structure reconstruction, rather than the sparse identification of features of previous methods. 

In this work, we use TO as an inverse design tool, to reconstruct a tree given an incomplete set of measurements. That is, it estimates the branch geometry wherever it is not directly observed.
To our knowledge, TO has not previously been employed as an inverse design tool, and has not previously been used for tree reconstruction from point clouds. 

In this paper, we use the open source software PANSFEM2~\cite{Pansfem2019} as the basis for the topology optimisation solver (section~\ref{sec:SIMP}). The software does not have an accompanying paper, but its topology optimisation method is very similar to Andreassen et al~\cite{andreassen2011efficient}, including the Heaviside projection filter option. 

\section{Method}
\begin{figure*}[h]
\centering
\includegraphics[width=1\linewidth]{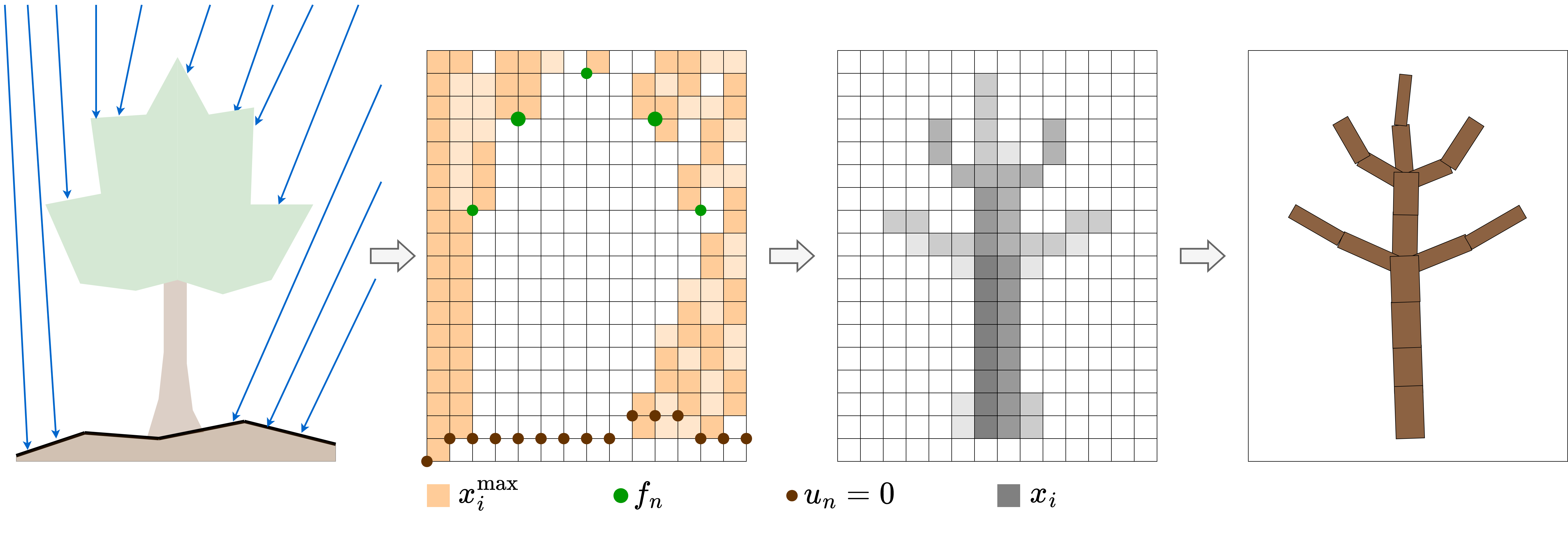}
\caption{Two-dimensional depiction of our method. From left to right: ray cloud and ground mesh of a natural scene, optimiser constraints, optimiser output (per-voxel wood density), and BSG. The optimiser constraints are the Dirichlet condition (red dots), Neumann condition (green dots), and the maximum volume condition (white=1, orange=0) as defined in Eq~\eqref{eq:formulation}.}
\label{fig:process}
\end{figure*}

Our approach to tree reconstruction from lidar data is to interpolate the lidar samples and the ground with an efficient distribution of woody material. This optimal material distribution problem is what is solved using topology optimisation. We now describe this standard topology optimisation formulation, followed by our set of adaptations for application to tree reconstruction in sections~\ref{sec:problem} to~\ref{sec:branchstructuregraph}. The main stages of this process are depicted in Figure~\ref{fig:process}.

\subsection{SIMP Topology Optimisation}
\label{sec:SIMP}

The standard approach to topology optimisation is based on iteratively solving a structural problem using the finite element method (FEM). 
The design space is discritised into a 3D grid of cubic elements (voxels) and the grid nodes (vertices). The stiffness $E_i$ of each element index $i$ is taken as a function of the `pseudodensity' design parameter $x_i$ using the  Solid Isotropic Material with Penalization (SIMP) method:
\begin{equation}
\label{eqn:element_modulus}
E_i(x_i)=E_{V}+x_i^p(E_{S}-E_{V})
\end{equation}  
giving physical values of $E_i$ between solid $E_S$, and void $E_V$ which is small but non-zero to avoid singularities. The penalty exponent $p$ drives convergence to a binary final solution, where $p\geq3$.

The optimal design is formed as a constrained minimisation of the system's total compliance $c(\vec{x})$:
\begin{equation}
\begin{aligned}
    \min_{\vec{x}} :\:&c(\vec{x})=\mat{U}^T\mat{K}\mat{U}=\sum_{i=1}^N E_i(x_i)\vec{u}_i^T\vec{k}_i\vec{u}_i \\
    \text{subject to}:\:&\mat{K}\mat{U}=\mat{F}\\
    &v(\vec{x})\le v^\mathrm{max} \\
    &0\le x_i\le x_i^\mathrm{max} \le 1
\end{aligned}
\label{eq:formulation}
\end{equation}
where the total volume of woody material $v(\vec{x})=\sum_{i=1}^N x_i$ has a prescribed upper limit $v^\mathrm{max}$, and $N$ is the total number of voxels. 

The global displacement matrix $\mat{U}$ is composed of per-element vectors $\vec{u}_i$, which are in turn composed of eight nodal displacements $u_n$. Likewise, the global load matrix $\mat{F}$ is composed of per-element load vectors $\vec{f}_i$ that are composed of eight nodal loads $f_n$. The global stiffness matrix $\mat{K}$ is composed of per-element stiffness matrices $\vec{k}_i$.

To avoid `checkerboard' distributions of pseudodensity, a spatial filtering and projection scheme is implemented, similar to \cite{Guest2004,Liu2021}.
First it smooths the elemental density accord to the average density of elements in the surrounding neighbourhood:
\begin{equation}
\tilde{x_i}=\dfrac{\sum_{j \in N_e} W_{ij}x_j}{\sum_{j \in N_e} W_{ij}}
\end{equation}
where $N_e$ is the set of elements within distance $R=1.75$ voxel widths of $i$, $W_{ij}$ is the filter weighting between elements $i$ and $j$, given by $W_{ij}=R-r_{ij}/R$, and $r_{ij}$ is the Euclidean distance between elements i and j.
The smoothed density is then `projected' through a smooth approximation of the Heaviside step function to obtain a more binary pseudodensity according to:
\small
\begin{equation}
\bar{x_i}(\beta)=\dfrac{\tanh{(0.5\beta)}+\tanh({\beta(\tilde{x_i}-0.5}))}
{2\tanh{(0.5\beta)}}
\end{equation}
\normalsize
where $\beta$ is the filter steepness parameter.

Eq~\eqref{eq:formulation} is solved using the CONLIN solver \cite{Fleury1989}, which solves a sequence of linearised subproblems and uses the sensitivities (Jacobian) of the system to increase solver efficiency. The sensitivities of the cost with respect to the densities can be found in Appendix \ref{appendix:sensitivity}.


\subsection{Application to Tree Reconstruction}
\label{sec:problem}
The topology optimisation is subject to two boundary conditions. The Dirichlet condition sets the displacements $u_n$ to zero on the nodes closest to the ground, as estimated in section~\ref{sec:ground}. The Neumann boundary condition fixes the nodal load $f_n$ in proportion to the number of lidar points that are closest to this node. The solved pseudodensity distribution $\vec{x}$ then represents the distribution of wood in the scene, which is the set of branches.

The resulting voxel tree is therefore the structure connected to the ground that is most efficient at resisting the loads applied at the lidar points. Unlike typical structural optimisation, we do not use \textit{vector} loads and displacements, representing gravitational force and sag, instead we use \textit{scalar} loads and displacements. This is a non-directional representation that is mathematically equivalent to the thermal model of topology optimisation, only using structural terms (load, displacement, stiffness) rather than the thermal equivalents (heat, temperature and thermal conductivity). 

This scalar model guarantees that the resulting structures are topologically trees (acyclic graphs), unlike the (cyclic graph) truss structures that are a signature of gravitational structural optimisation. The structures are also always contiguous and connected to the ground (the Dirichlet boundary), and the requirement to minimise compliance preferences straight branches between lidar points. This is because the compliance at the end of a curved branch is integrated over a greater length than that of a straight branch, and the compliance is being minimised. All of these properties fit with our assumptions about the input data that we listed in Section~\ref{sec:introduction}.

Because this method is not identifying sparse features, the problem does not become underdetermined when such features are missing. This is because the state equation $\mat{K}\mat{U}=\mat{F}$ ensures that there is an underlying physical model (a prior) wherever data is missing. For this reason, tree structures occur even when there only 5 points in the cloud and no ray information (see Figure~\ref{fig:gilbert}), or even if the cloud were a set of randomly sampled points, or contained walls or cyclic structures like pylons. In every case this scalar model of topology optimisation will generate the lowest compliance acyclic structures to connect the available data to the ground.
\subsection{Controlling Branch Angle}
\label{sec:branchangle}
The above system generates a tree structure of woody material to match the specified point sources. The closer this tree structure is to the shape of real trees, the more accurate its interpolation will be in unobserved regions. Unfortunately the generated tree structure has a fairly narrow branch angle and it has been demonstrated~\cite{yan2018non} that this angle reduces to zero in the highest fidelity solutions. It is therefore important to be able to preference larger branch angles, in particular to control this angle through a parameter, according to the type of trees in the scene. 

We do this by including an additional spatial low pass filtering kernel, applied for each element $i$:
\begin{equation}
    A_i=\sum_{j} \dfrac{x_j}{\max(r_{ij}, 1)}
    \label{eq:convolution}
\end{equation}
where $r_{ij}$ is the distance in voxels between $i$ and $j$ in the 3D grid. This is used as a multiplier on the stiffness:
\begin{equation}
    \tilde{E}_i(x_i)=A_i^\alpha E_i(x_i)
    \label{eq:multiplier}
\end{equation}
for some $\alpha$, where $\alpha=0$ represents the standard solution and $\alpha>0$ produces larger branch angles in unobserved regions.

\begin{figure}[h]
\centering
\includegraphics[width=1\columnwidth]{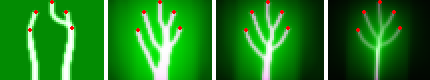}
\caption{Convolution with radial reciprocal kernel $A_i^\alpha$ shown in green, for branch angle parameter $\alpha=0,1/4,1/2,1$ (left to right). The resulting tree from five point loads is shown in white. }
\label{fig:gilbert}
\end{figure}

Excluding discretisation artefacts $A_i$ is proportional to branch cross-sectional area for any particular branch shape. So Eq~\eqref{eq:multiplier} scales stiffness as a power of the branch's thickness. Due to the higher stiffness of the thicker branches, the smaller ones can connect to them farther from the sink, which makes the branching angle larger, as seen in Figure~\ref{fig:gilbert}.

\subsection{Estimating Ground}
\label{sec:ground}
We estimate the ground surface height from the ray cloud by interpolating the set of `lowest points'. We define these lowest points as each point with gradient $\frac{\Delta z}{\Delta x}$ to every other point less than some gradient $g$. Specifically the set of point indices $\{i: \forall_j\; z_i-z_j < g\mid x_i-x_j\mid\}$. This can be thought of as a sand model of the terrain height, since dry sand settles up to a maximum gradient. The interpolation is achieved by removing the $z$ component of the set of lowest points, calculating its Delauney triangulation, then restoring the $z$ component to the triangle vertices. This triangular mesh allows a ground height to be calculated for any horizontal location within the Delauney triangulation. We use the open source library raycloudtools~\cite{lowe2021raycloudtools} for a fast implementation of this ground surface reconstruction.  
\subsection{Free Space}
\label{sec:freespace}
In addition to lidar points, the lidar rays themselves serve an important role in mapping out the free space (the air) in the scene. This is important in preventing the algorithm from generating branches in regions that are observed to be free space. The free space is mapped by using quarter-width subvoxels to record the presence of any ray that passes through the subvoxel. This is achieved by `walking' each ray through the subvoxel grid using a line drawing algorithm, if any ray passes through a subvoxel then it is flagged as unoccupied. The 64 unoccupied statuses can be encoded in a single long integer per voxel, and the number of occopied subvoxels divided by 64 provides the maximum density value $x_i^\mathrm{max}$ for each voxel $i$.

\subsection{Branch Structure Graph}
\label{sec:branchstructuregraph}
The output of the optimisation process is a grid of voxels containing density values from zero to one. These densities form a treelike structure, with voxel density representing the percentage of coverage of the underlying tree. This is a useful data structure in its own right, and can be used for physical simulation, biomass estimation and other metrics, but it is also valuable to convert the results into the Branch Structure Graph (BSG) referred to in Section~\ref{sec:introduction}. This provides a canonical description of the topology and geometry of the trees. 

In order to extract the graph, we employ a variation of the distance bucketing method used on point clouds~\cite{hu2017efficient, xu2007knowledge, calders2015nondestructive}. Our method uses Dijkstra's algorithm~\cite{dijkstra1959note} to generate the shortest path forest between the lowest non-empty voxels and all other non-empty voxels, considering the nearest 26 voxels (the Moore neighbourhood) for each voxel. Rather than use Euclidean distance between voxel centres, we divide the Euclidean distance to each neighbour by its density value. This rewards paths that follow the densest set of voxels. We then use the Euclidean path distance from the ground to bucket the voxels into distance intervals that are two voxels wide.

For each bucket, all voxels less than a density threshold are removed and the bucket is split into one tree segment for each connected set of voxels $S$. Each tree segment's parent is recorded, then its centroid $c_s$ and radius $r_s$ are calculated according to:
\begin{align}
    c_s&=\sum_{i\in S}x_i \vec{p}_i \\
    r_s&=w\sqrt{\frac{1}{2\pi}\sum_{i\in S}x_i},
\end{align}
where $\vec{p}_i$ is the centre of voxel $i$, and $w$ is the voxel width. The density threshold is set at one quarter of the maximum density value of the parent segment. This threshold therefore reduces on the thinner branches, ensuring that these low density structures are also represented.

\begin{figure}[h]
\centering
\includegraphics[width=.4\columnwidth]{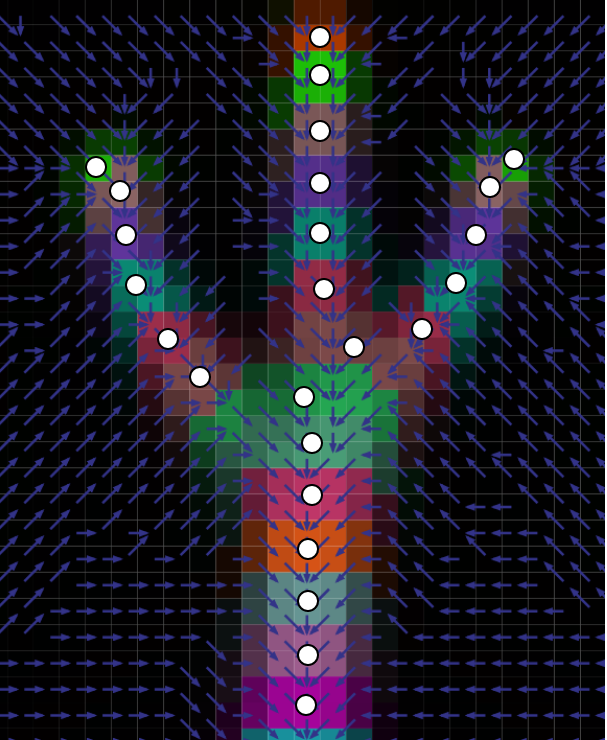}
\caption{Dijkstra's algorithm on a 2D three-point tree, using density-scaled distance function, and segmented by Euclidean path distance every two element widths. Brightness is pseudodensity. White points are the density-weighted centroids of each contiguous segment, which form the nodes of the Branch Structure Graph.}
\label{fig:generate_BSG}
\end{figure}

This process of splitting the buckets into segments and connecting to the parent segment is repeated until a full graph of centroids and radii is formed, which is the BSG. Figure~\ref{fig:generate_BSG} illustrates this process on a two-dimensional tree, with one colour per bucket and the resulting BSG centroids shown as white disks. The branches below a certain radius can be cropped to avoid reconstructing branches that are too small compared to the voxels that they are representing. We crop branches when $r_s<w/4$ in our experiments.

\section{Experiments}\label{sec:experiments}
We assess our method using 21 acquired datasets from our research site in Pullenvale, Australia. These were acquired using scanners based on a rotating Velodyne VLP16 lidar, and registered into a 3D point cloud using the Wildcat SLAM software~\cite{Ramezani2022}. The datasets cover a wide range of vegetated environments, including multiple tree species such as figs, eucalypts, palms and a monkey puzzle tree; and a wide range of heights from 1.5 m up to 28 m. They have been captured specifically to assess our method under the typical failure cases of existing tree segmentation and reconstruction algorithms, and are labelled as (failure) case \#1 to \#21. 

The aim of this data is to test our hypothesis that the topology optimisation approach has fewer or less extreme failure cases than the existing methods that rely on sparse features. The experiments are therefore a study of the ability to generate trees that fit to poorly-defined and varied input data. We consider this to be an assessment of the robustness of the algorithm to these varied environments. Here we explain the choice of failure cases:

\begin{description}
\item[Non-problematic:] these are included for completeness and are cases \#1 and \#2
\item[Disconnected canopies:] these are sections of canopy that do not connect to the ground due to cropping the scan, or due to the tree being only partially observed. They are included in cases \#3 and \#4 as they are a common boundary issue that affects all reconstruction methods.
\item[Hidden trunks:] some existing algorithms rely on clearly distinguishable trunks in order to segment~\cite{li2010segmentation,zhong2016segmentation,liu2021point,burt2019extracting} or reconstruct trees~\cite{treeQSM,du2019adtree,fan2020adqsm}, we have scanned scenes with hidden or overgrown trunks in cases \#6, \#7 and \#8.
\item[Non-tree objects:] the majority of reconstruction algorithms assume that there are no non-tree objects in the scan, we have therefore included scans with a shelter (\#5), walls (\#9, \#18) and lamp posts (\#7, \#14 and \#17)
\item[Canopy close to ground:] in case \#10 we include a scene with canopy reaching almost to ground, as this can be falsely interpreted as a trunk location by reconstruction methods that ignore ray information~\cite{tagliasacchi20163d, hu2017efficient,xu2007knowledge,treeQSM,du2019adtree,fan2020adqsm}
\item[Overlapping trees:] several segmentation methods rely on gaps or thinning between \\ trees~\cite{ayrey2017layer,heinzel2018constrained,wang2018scalable}, we therefore include overlapping trees in cases \#11, \#12 and \#16
\item[V-shaped trunk:] some methods assume that trunks are singular at the base~\cite{li2010segmentation,zhong2016segmentation} so we consider a V-shaped trunk in case \#13
\item[Hidden branches:] most branch reconstruction methods and some segmentation \\ methods~\cite{luo2021individual} rely on clearly observable branches so we include scenes with hidden branches in cases \#14 to \#21
\item[Ambiguous crowns:] for those that rely on crown features~\cite{li2012new} we include poorly identifiable crown scans in cases \#11 and \#15. Trees tucked under other trees are a failure case for crown extraction algorithms that consider only the upper canopy, so we include this situation in case \#18
\item[Uneven ground:] sloped and uneven ground needs to be considered in natural environments, this is represented in case \#19
\item[One-sided scan:] this is included in order to observe biases in tree structures due to the scan being acquired from principally one direction, and occurs in cases \#9 and \#21 
\end{description}

In cases \#2 to \#4 we removed reflections off the ground, which cause erroneous underground points that disturb the ground surface estimation. There was no other pre-processing of the 3D ray clouds. 

Ground height was estimated using the described method with gradient $g=1$ everywhere, but $g=2$ to cater for the single steep gradient case \#19. All points less than 30 cm above the ground were excluded from applying a load. This is to avoid the algorithm generating branches along the ground, and to reduce the presence of undergrowth in the final result. All scans are processed with the same design parameters: $\alpha=1$, $N=5,000,000$ and wood volume $v_\mathrm{max}=v_\mathrm{filled}/5$ where $v_\mathrm{filled}$ is the volume of voxels containing lidar points. We run for 20 iterations, over which we ramp up the parameters $p$ from 1 to 3, and $\beta$ from 1 to 4, as a form of annealing to discourage early local minima. 

The experimental results for all of the scenes that we scanned are presented in Tables~\ref{tab:Qualitativeresults1} to~\ref{tab:sphereexperiment} and Figure~\ref{fig:voxel_sensitivity}, and they represent in order: a qualitative study, a quantitative comparison, an ablation study, an aerial comparison, an occlusion study and a voxel size sensitivity study.

The qualitative study (Tables~\ref{tab:Qualitativeresults1} and \ref{tab:Qualitativeresults2}) visually compares the output BSG to the input scan. The left image is the set of ray cloud end points (the point cloud). The next column image is a rendering of all voxel pseudodensities $x_i\ge 0.5$, which indicates the main wood regions in the raw solver output, but excludes the thinnest branches where $x_i<0.5$. Any differences between this and the wood structure in the point cloud image represent inaccuracies in the topology optimisation part of the algorithm (Sections~\ref{sec:SIMP} to~\ref{sec:freespace}). The next column shows the BSG output, rendered as blue cylinders, this rendering does show the thinner branches. Any other differences between this and the voxelised result represent inaccuracies in the BSG reconstruction (Section~\ref{sec:branchstructuregraph}). The exposed branches and trunks in the left image can be compared directly to the BSG branches, and we encourage the reader to view this paper's accompanying data for further comparison. 

In order to compare against existing reconstruction methods, we recall that available methods rely on a two-stage approach of tree segmentation followed by single-tree reconstruction. We test against just the first stage with the understanding that poor tree segmentations will necessarily generate poor branch reconstructions, due to the single-tree requirement of the second stage. In the right hand column we compare to the tree segmentation in the TreeSeparation method~\cite{Jinhu-Wang,wang2018scalable}, each colour represents a separate identified tree, so a correct segmentation would show a unique colour for each tree in the ray cloud.

The quantitative comparison (Table~\ref{tab:comparison}) quantifies the accuracy of these 13 cases, and compares them against two other existing methods. The accuracy measurement is described in Section~\ref{sec:accuracy}. We assess the tree segmentation accuracy on two established methods Treeseg~\cite{andrew_burt_2021_4884923,burt2019extracting} and TreeSeparation~\cite{Jinhu-Wang,wang2018scalable}, and compare that to the number of principle trees that our method reconstructs. We calculate this as the number of trees with trunk diameter more than 30\% of the maximum trunk diameter to avoid counting small pieces of undergrowth. Lastly, we have manually inspected the point clouds and counted the number of principle trunks visible in these clouds, using the same criterion. 

Both segmentation libraries require a pair of parameters to be set, reflecting the average scale of the trees in the datasets. In both cases the parameters were within expected ranges and performed better than the default values. For Treeseg the minimum and maximum stem diameters were set to 0.05 m and 1 m respectively. For TreeSeparation the voxel height was set to 1.5 m, while the minimum tree diameter was set to 5 m for the widely spaced cases (\#1 to \#5) and 3 m for the remaining denser cases.

The ablation study (Table~\ref{tab:hardcases2}) compares the reconstruction results for two different values of the convolution kernel parameter $\alpha$. This parameter is only significant when whole branches are occluded. So we assess the performance for two values of this parameter ($\alpha=0$ and $\alpha=1$) for the six most occluded cases. The results match the behaviour in Figure~\ref{fig:gilbert}. 

In the aerial comparison (Table~\ref{tab:aerialscan}) we compare a dense handheld scan to a sparse sparse aerial scan of the same Morton bay fig tree. This aerial scan does not observe the trunk at all, and neither scan have identifiable observations of the tree's internal branches, due to its thick foliage. 

An occlusion study (Table~\ref{tab:sphereexperiment}) assesses our algorithm's robustness to occluded branches inside a tree's canopy. For this we have taken a tree with visible branches (case \#5) and have manually removed the features that do not fit the assumptions of the algorithm, such as the shelter and a lamp post, so that they are not part of the accuracy assessment. We have then generated an occluded version of this ray cloud, removing all points and ray sections within a radius of each tree centre. Lastly, we calculate the accuracy of the reconstructed BSG of this occluded ray cloud against the original unoccluded ray cloud. This is performed for multiple different radius values. We have chosen to employ this spherical occlusion because it is simple and reproducible, and because it represents a worst-case form of occlusion where central observations are completely absent. The reconstruction accuracy measure is described in the following section. 

Finally, the sensitivity study (Figure~\ref{fig:voxel_sensitivity}) evaluates the reconstruction accuracy for different choices of voxel size $w$. It uses the same dataset as the occlusion study (case \#5 with artificial objects removed), and assess the Surface Error for voxel counts of 6 million and 4 million voxels, and descending powers of two for each. The cubic voxel grid width $w$ (rounded to the nearest centimetre) that best fits these counts is then used for each sample. It results in slightly altered voxel counts from 6.69 million down to 66,096 voxels, approximately a 100-fold range.

\section{Accuracy}\label{sec:accuracy}
In order to measure how well the tree reconstruction fits the input point data, we begin by manually removing points that are not part of the reconstructed trees. This includes ground points, undergrowth, fences, points from the moving user, and tree canopy from trunks outside of the bounding box. Whilst this process is performed manually, the features are distinct and relatively unambigious. 

The measure that we then use represents the average surface error of the reconstructed tree branches relative to the set of remaining points $P$. We will refer to it as Surface Error or SE, it is calculated by finding the set of points $P_i$ closest to each cylindrical branch segment $i$: 
\begin{align}
P_i = \{p\in P\; |\;  \underset{j\in 1..n}{\mathrm{argmin}}\;d(p,c_j) = i \},
\end{align}
where $d(p,c_j)$ is the unsigned distance between point $p$ and the surface of cylinder $c_j$, and $n$ is the number of cylinders in the BSG. The mean distance error for cylinder $i$ is therefore:
\begin{align}
e_i = \underset{p\in P_i}{\sum}{d(p,c_i)}/|P_i|
\end{align}
We then take the surface-area weighted average of the mean distance errors: 
\begin{align}
\mathrm{SE} =\frac{\sum_i{A_i e_i}}{\sum_i{A_i}},
\end{align}
where $A_i$ is the surface area of cylinder $i$ (excluding end disks).  

We set the weight $A_i$ to the cylinder surface area because the BSG is a surface representation of the observed surface points of the branches, so the surface is the object that we are measuring the accuracy of. It is possible to set the weight to the cylinder's length or its volume, which roughly correspond to the accuracy of the tree skeletons and their volumes respectively. Being dimensionally intermediate between a length and a volume, the cylinder surface area is also the better single choice to approximate all three accuracy metrics.  

Setting the weight to $A_i=|P_i|$ would give the mean distance between points and the reconstructed surface, as has been used previously in assessing tree branch reconstruction~\cite{du2019adtree}. Unlike this choice, SE is invariant to local variations in point density. It is also invariant to changes in branch segment resolution, such as if a single cylindrical segment is split into a chain of shorter segments. Since we are estimating the accuracy of a reconstructed surface, this surface-area weighting ensures that larger surfaces are weighted accordingly, and it also minimises the impact of false matches to foliage points, which are impractical to remove from the point clouds. 

The accuracy is evaluated on all datasets where branches are sufficiently visible, which are those in Tables~\ref{tab:Qualitativeresults1} and \ref{tab:Qualitativeresults2} and~\ref{tab:sphereexperiment}. In the latter table the reconstruction is measured against the uncropped set of points, to assess the algorithm's ability to reconstruct unobserved branches deep within a canopy. 

Manual measurements of branch lengths, diameters and branch angles is another way to assess reconstruction accuracy. We prefer the Surface Error metric in this paper because there are approximately 40,000 branches across the 21 datasets, so manual assessment would be impractical and heavy random sampling would counter the accuracy gained by using manual measurements over the Velodyne lidar, which has a 3 cm range noise. Additionally, Surface Error measures the tilt of trunks, the position of the branches and how well they fit to a curved branch, this all matters in general-case tree reconstruction.

Two other methods for assessing accuracy are destructive harvesting and the use of simulated data from realistic virtual tree models~\cite{westling2021simtreels,esmoris2022virtual}. We have not employed these methods in this paper, but they are both valuable techniques that could be applied in future to help assess the reconstruction accuracy. 

\begin{table*}[]
\begin{adjustwidth}{-3cm}{-3cm}
\begin{center}
\caption{Qualitative results (Part 1), showing the original point cloud, topology optimised voxel representation of the tree, and the reconstructed branch structure graph. The segmented trees produced by TreeSep \cite{wang2018scalable} are presented for comparison.}
\begingroup
\setlength{\tabcolsep}{3pt} 
\renewcommand{\arraystretch}{0.0} 
\begin{tabular}{c|c|c|c|c}
    Case & Point Cloud & Voxelised Result & BSG & TreeSep \\
    \shortstack{1\\Single tree} & 
    \includegraphics[height=\picheight cm]{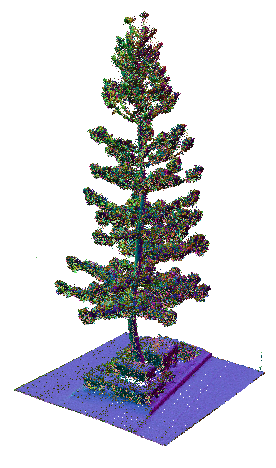} & 
    \includegraphics[height=\picheight cm]{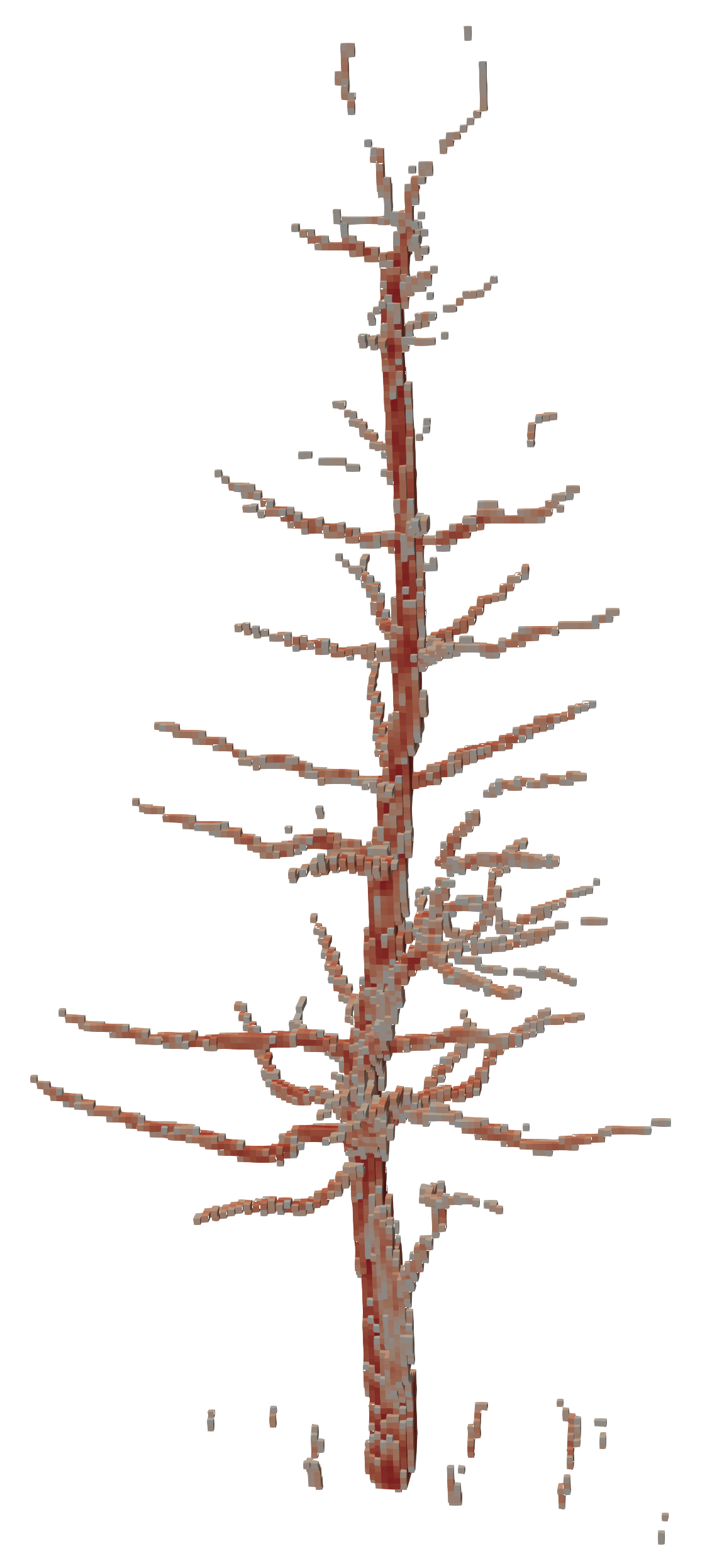} & 
    \includegraphics[height=\picheight cm]{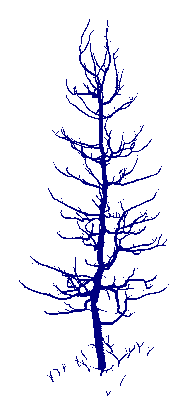} &
    \includegraphics[height=\picheight cm]{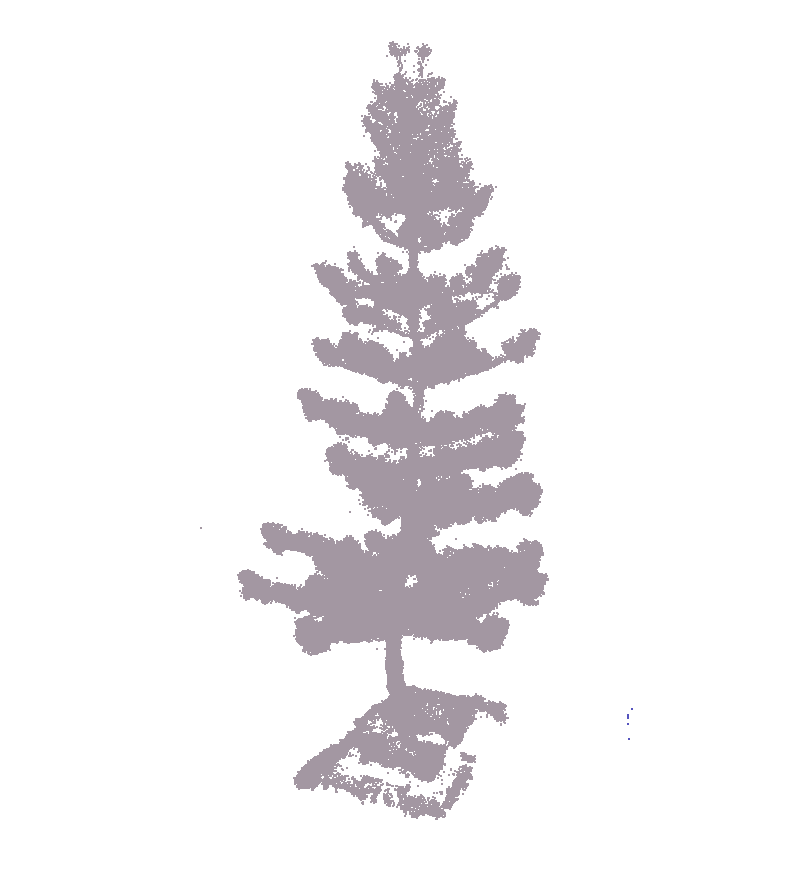} \\
    \shortstack{2\\Two trees} & 
    \includegraphics[height=\picheight cm]{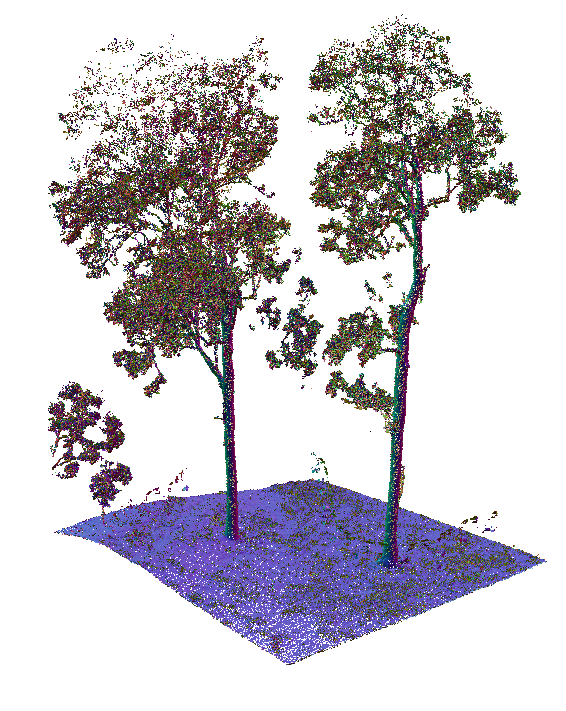} & 
    \includegraphics[height=\picheight cm]{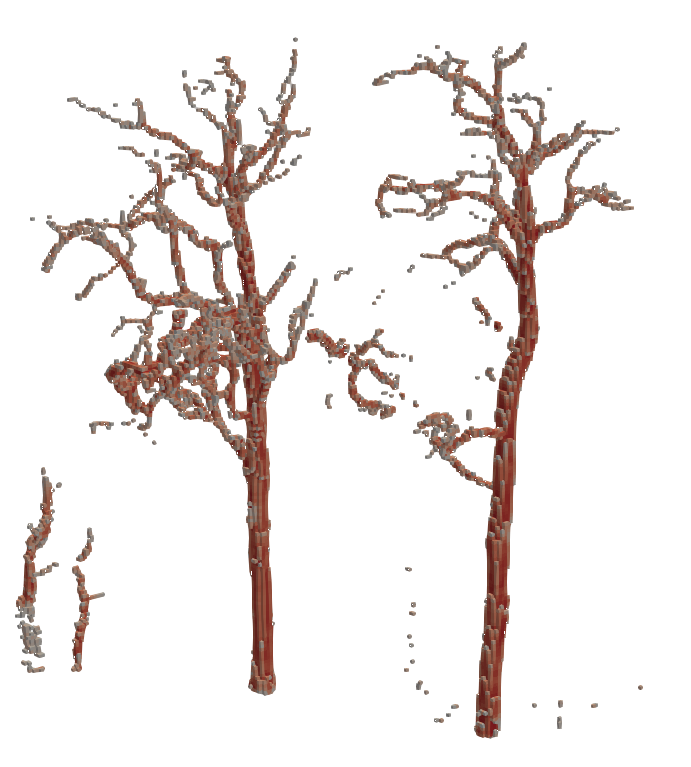} & 
    \includegraphics[height=\picheight cm]{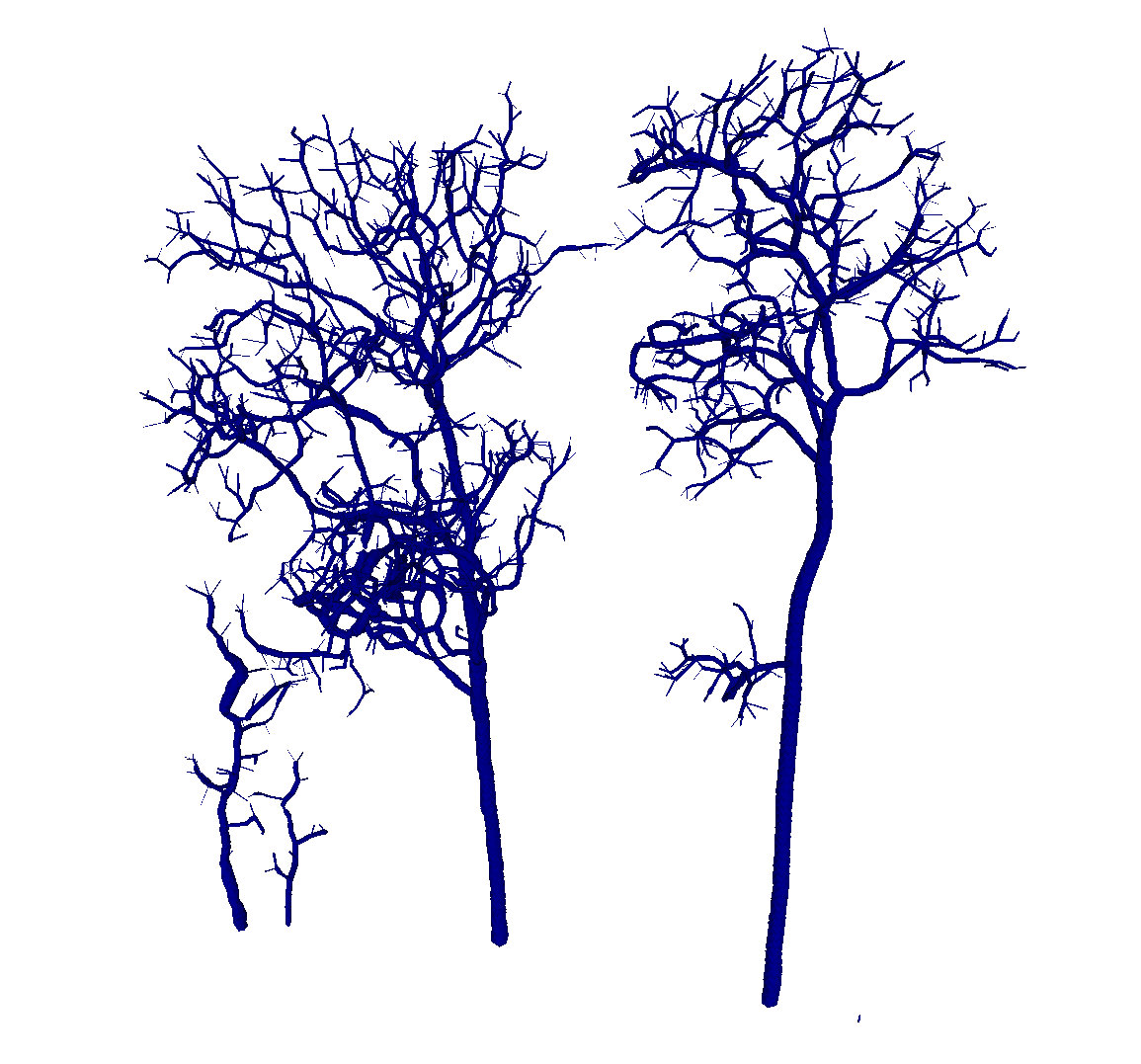} &
    \includegraphics[height=\picheight cm]{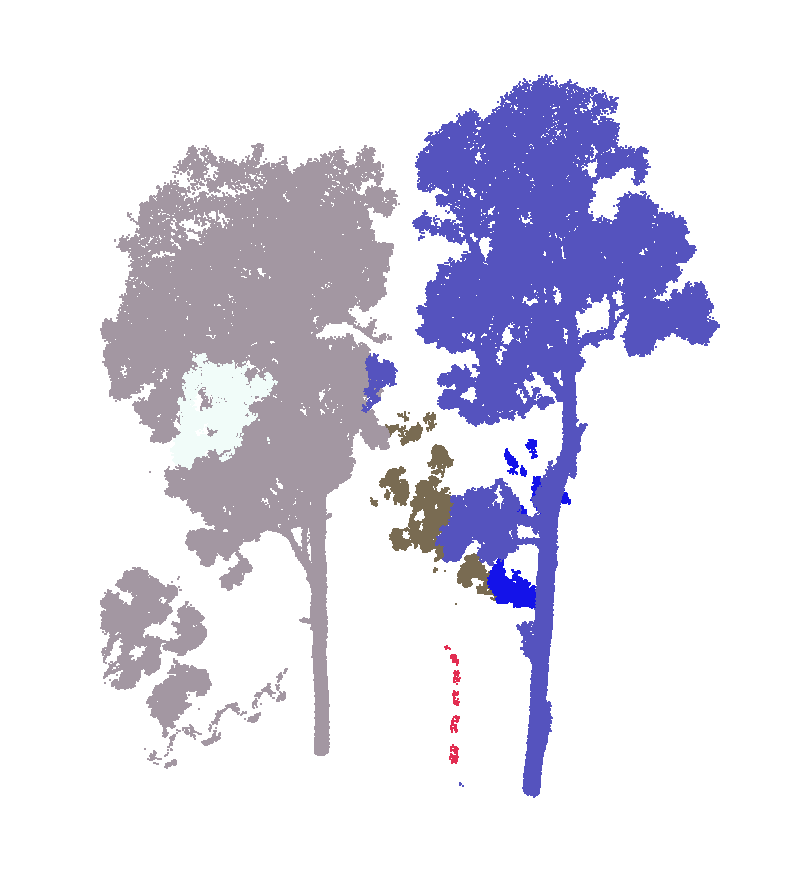} \\
    \shortstack{3\\Partial tree \\ canopy on left} & 
    \includegraphics[height=\picheight cm]{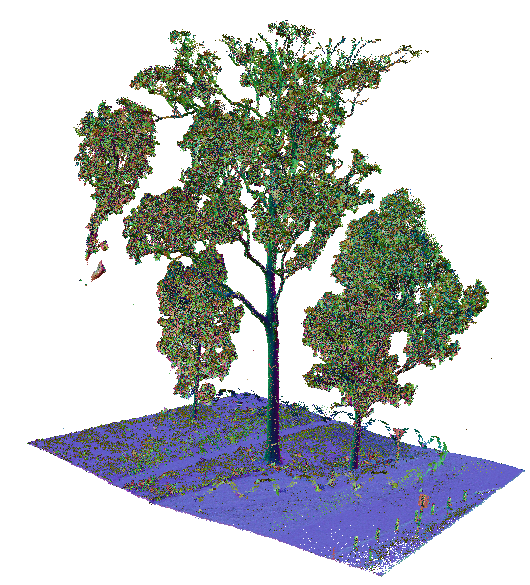} & 
    \includegraphics[height=\picheight cm]{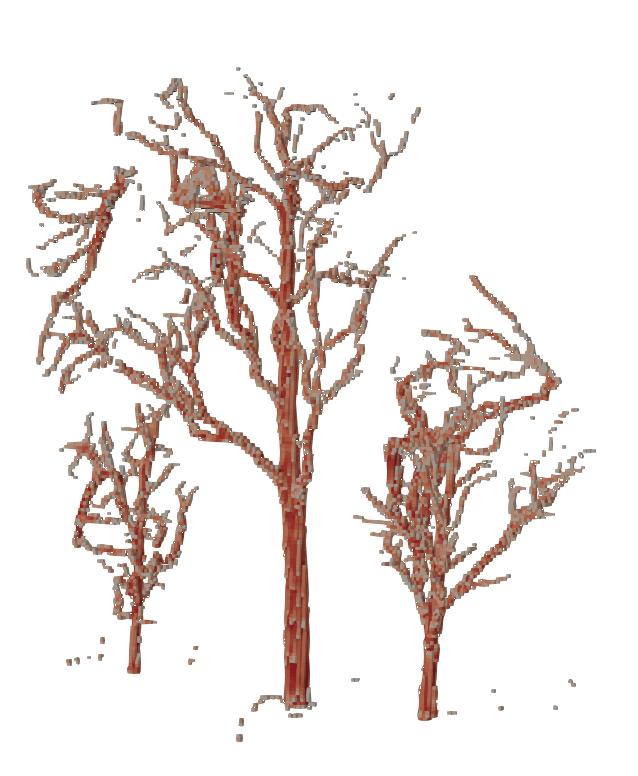} & 
    \includegraphics[height=\picheight cm]{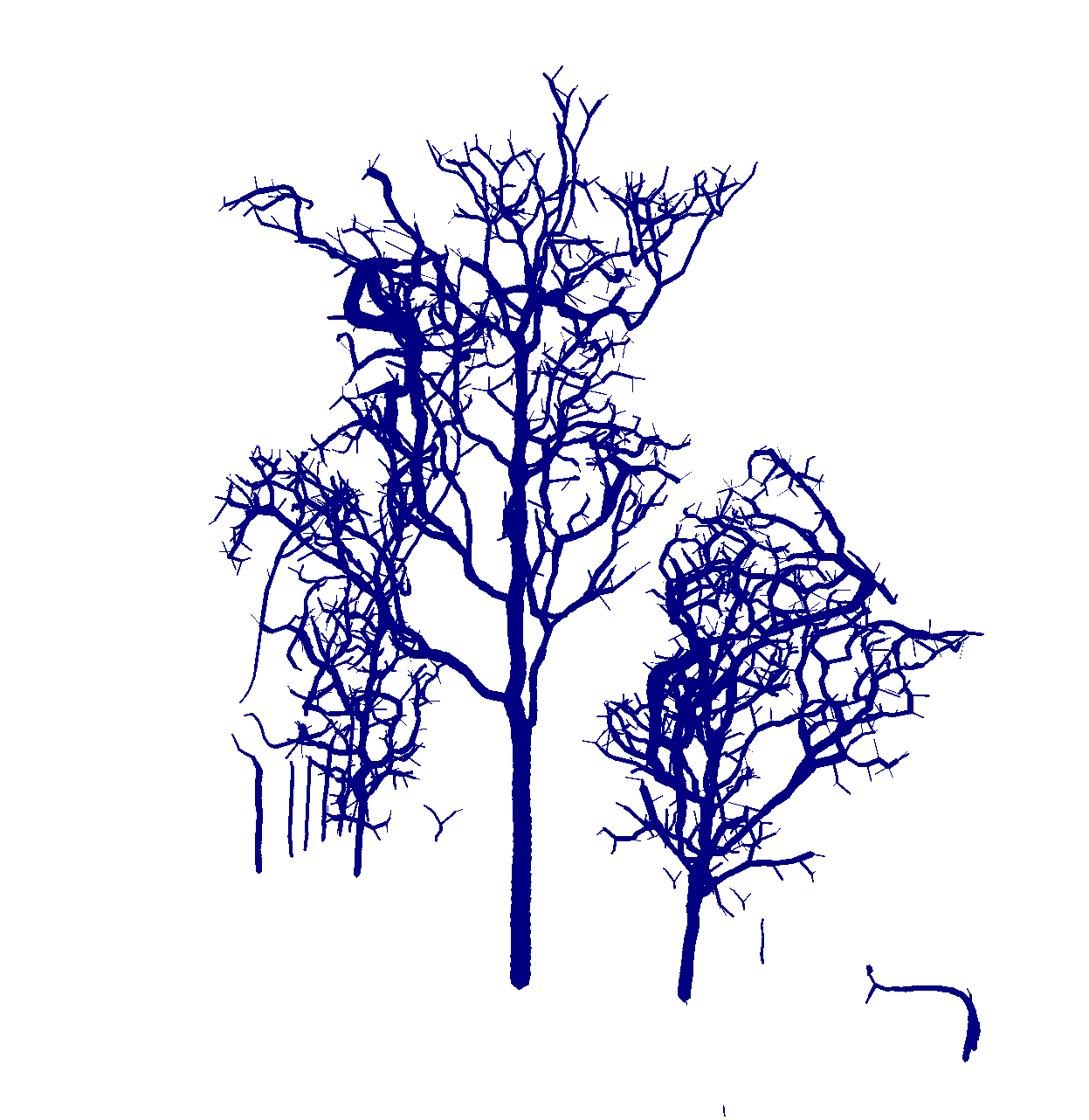} &
    \includegraphics[height=\picheight cm]{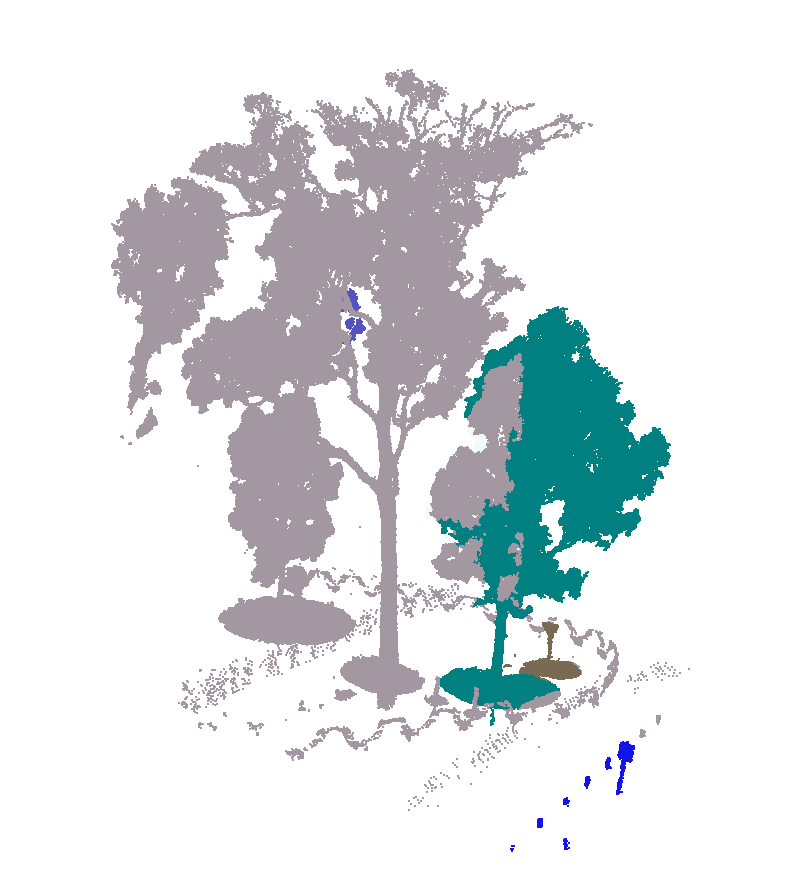} \\
    \shortstack{4\\Partial tree \\ canopy on right} & 
    \includegraphics[height=\picheight cm]{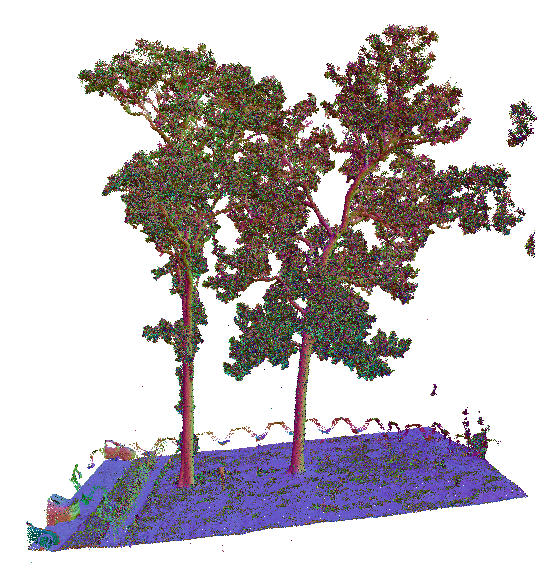} & 
    \includegraphics[height=\picheight cm]{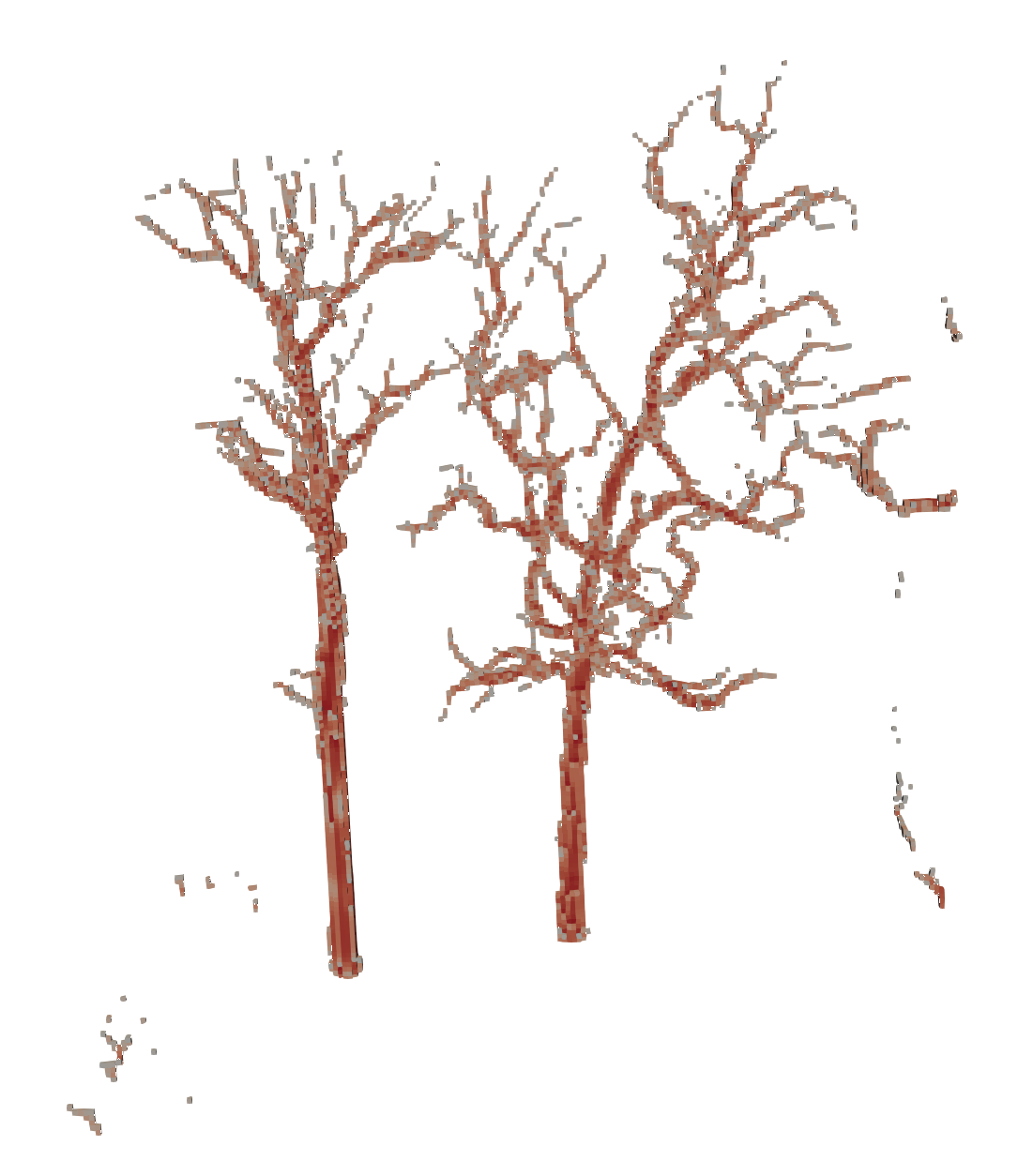} & 
    \includegraphics[height=\picheight cm]{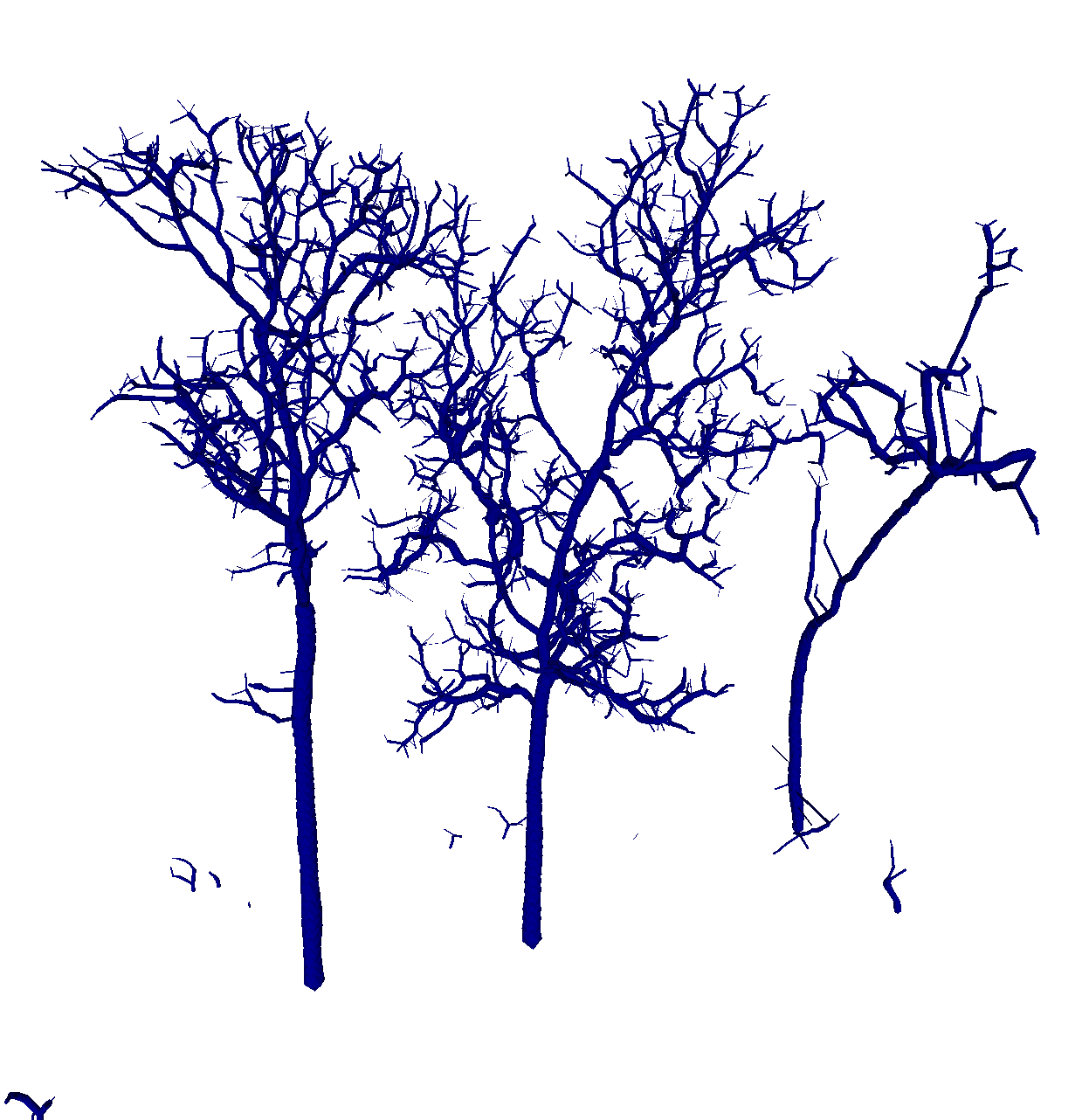} &
    \includegraphics[height=\picheight cm]{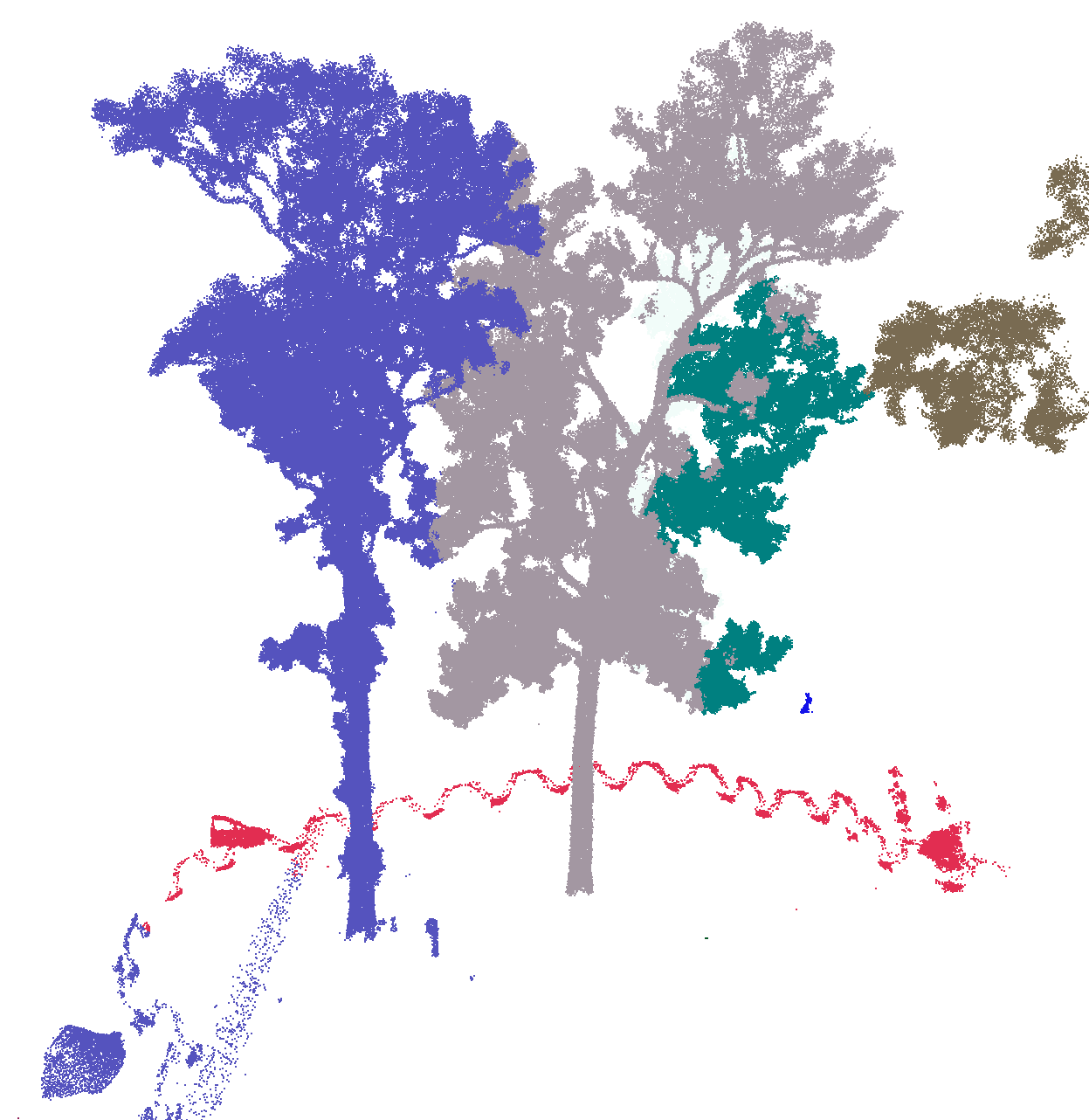} \\
    \shortstack{5\\Trees and \\shelter} & 
    \includegraphics[height=\picheight cm]{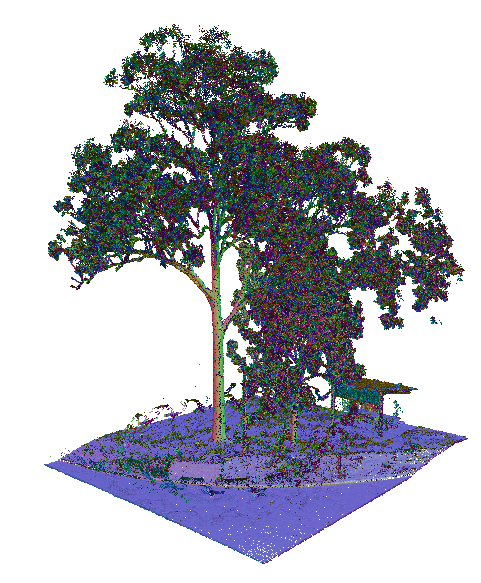} & 
    \includegraphics[height=\picheight cm]{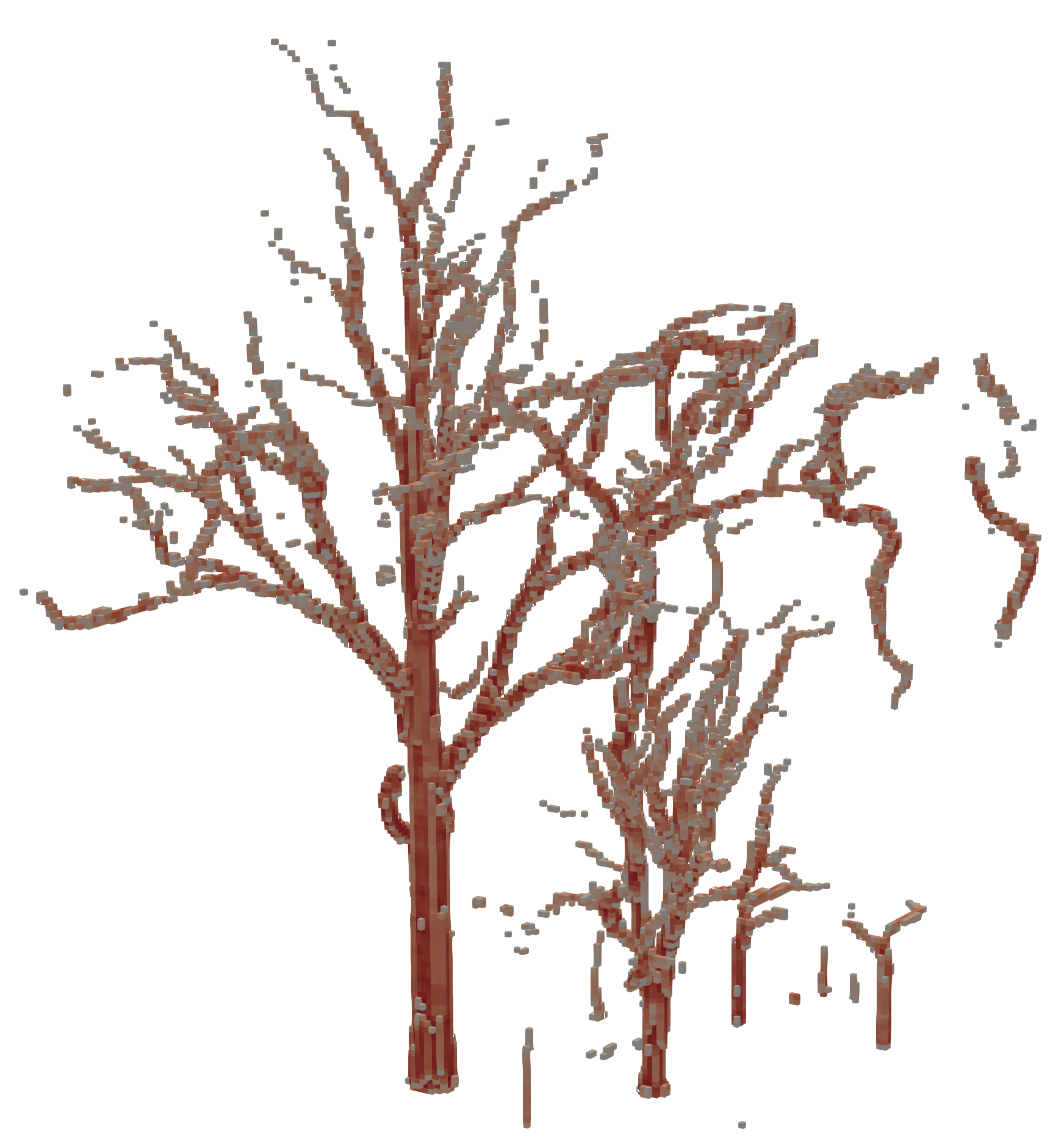}& 
    \includegraphics[height=\picheight cm]{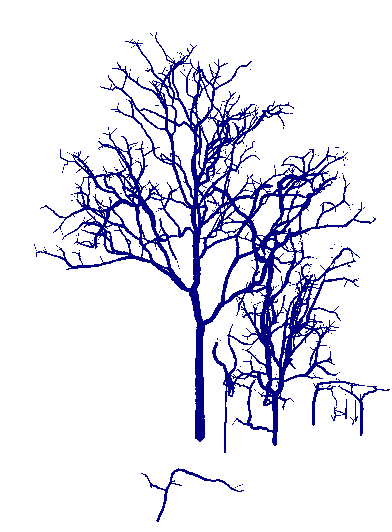} &
    \includegraphics[height=\picheight cm]{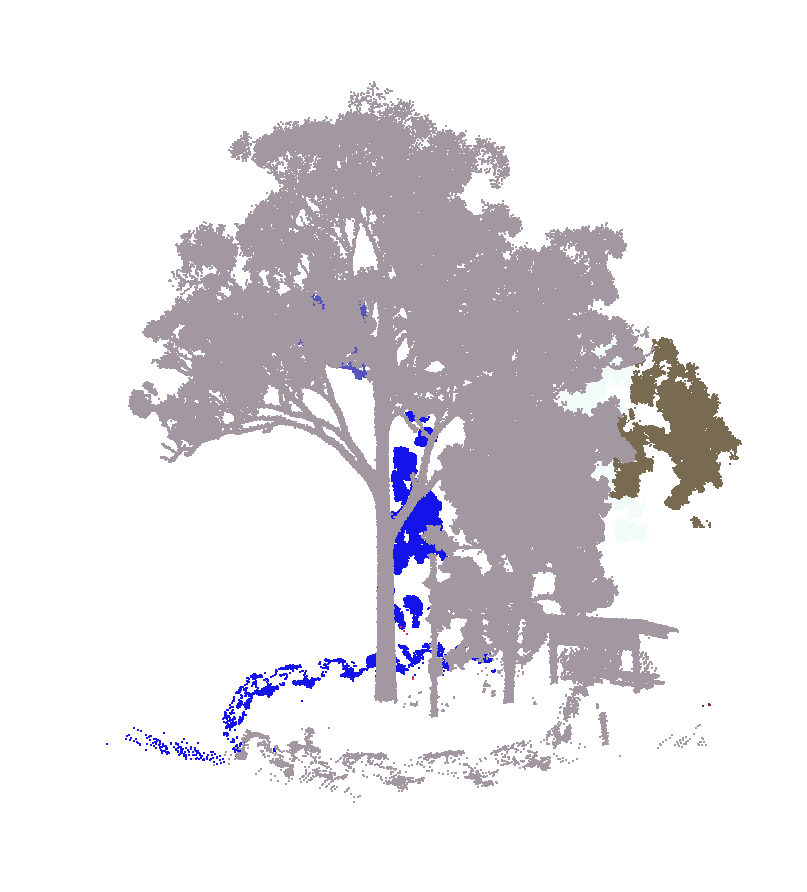} \\
    \shortstack{6\\Trunks occluded\\behind overpass} & 
    \includegraphics[height=\picheight cm]{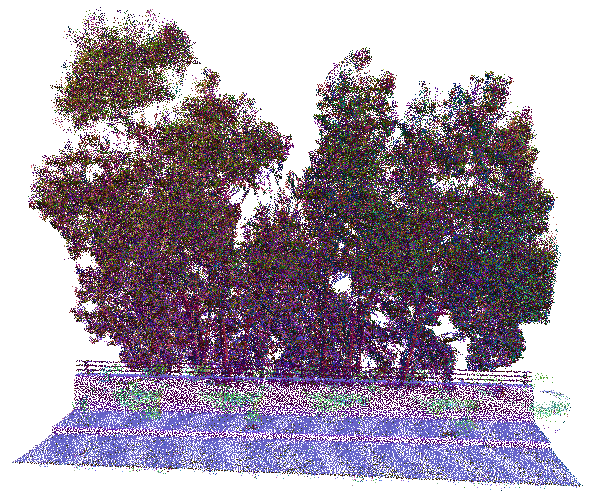} & 
    \includegraphics[height=\picheight cm]{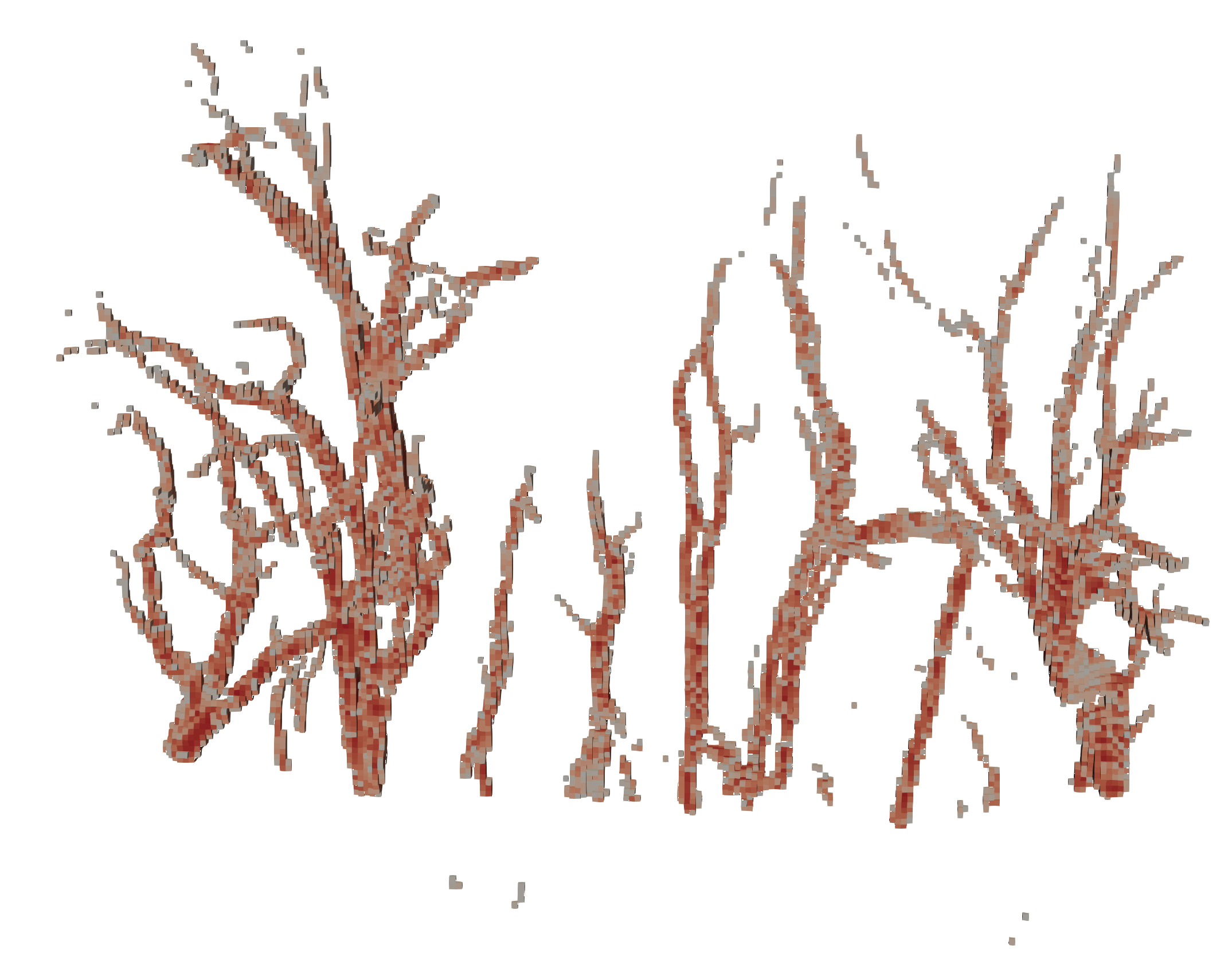} & 
    \includegraphics[height=\picheight cm]{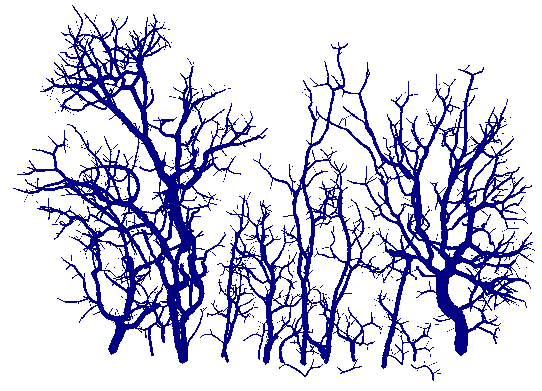} &
    \includegraphics[height=\picheight cm]{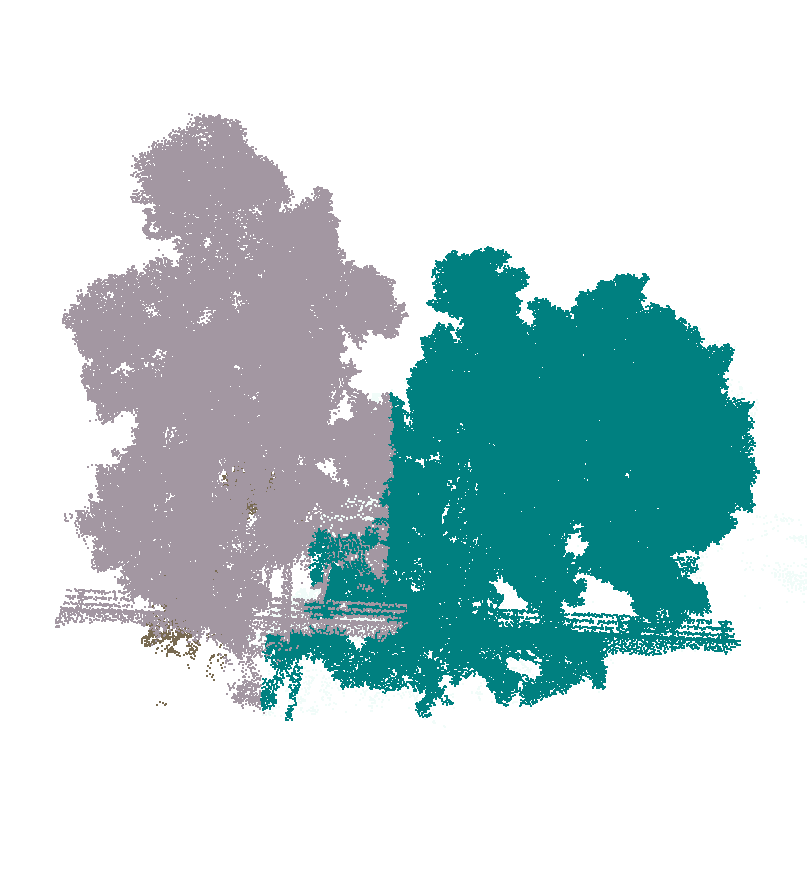} \\
    \end{tabular}
    \endgroup
    \label{tab:Qualitativeresults1}
    \end{center}
    \end{adjustwidth}
\end{table*}

\begin{table*}[]
\begin{adjustwidth}{-3cm}{-3cm}
\begin{center}
\vspace{-3cm}
\caption{Qualitative results (Part 2), showing the original point cloud, topology optimised voxel representation of the tree, and the reconstructed branch structure graph. The segmented trees produced by TreeSep \cite{wang2018scalable} are presented for comparison.}
\begingroup
\setlength{\tabcolsep}{3pt} 
\renewcommand{\arraystretch}{0.0} 
\begin{tabular}{c|c|c|c|c}
    Case & Point Cloud & Voxelised Result & BSG & TreeSep \\
    \shortstack{7\\Trunk covered\\in vegetation} &
    \includegraphics[height=\picheight cm]{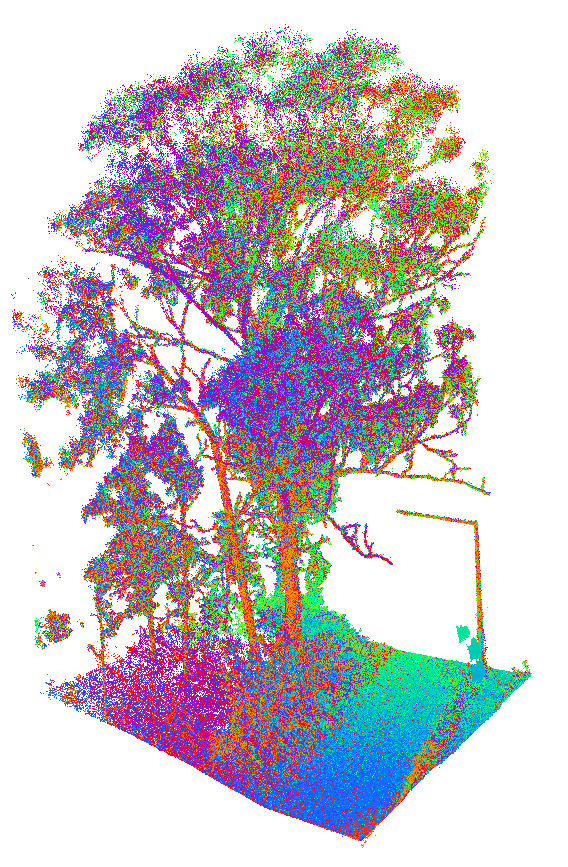} & 
    \includegraphics[height=\picheight cm]{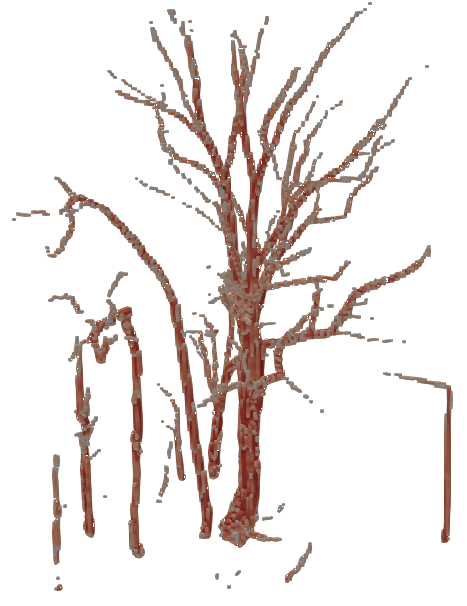}& 
    \includegraphics[height=\picheight cm]{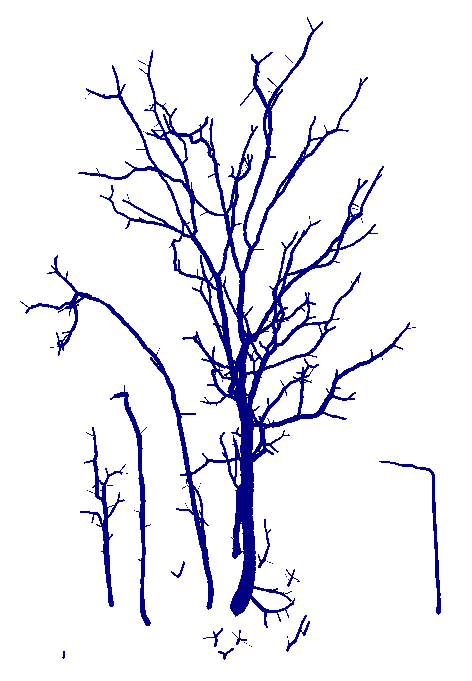} &
    \includegraphics[height=\picheight cm]{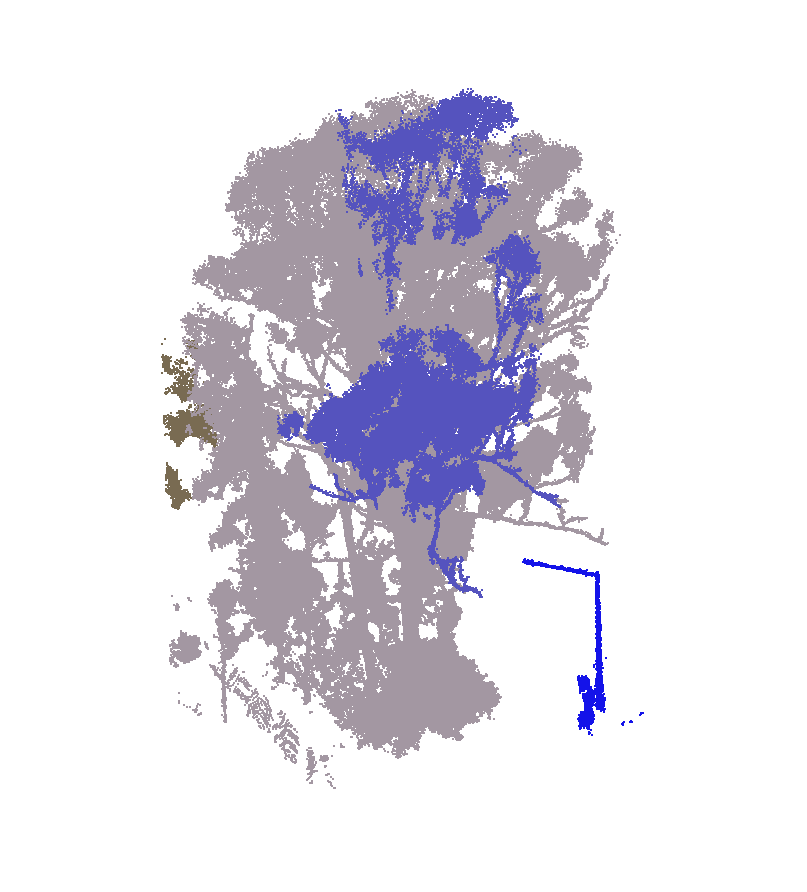} \\
    \shortstack{8\\Trunk occluded\\by vegetation} &
    \includegraphics[height=\picheight cm]{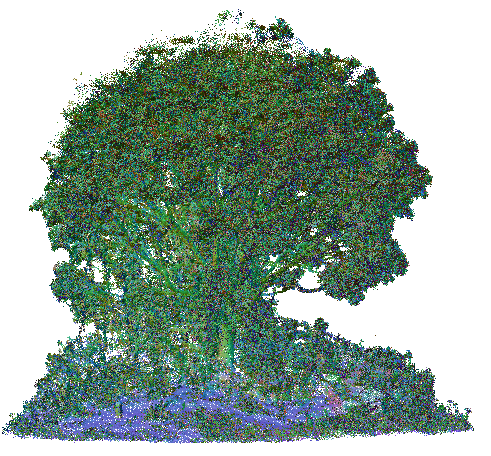} & 
    \includegraphics[height=\picheight cm]{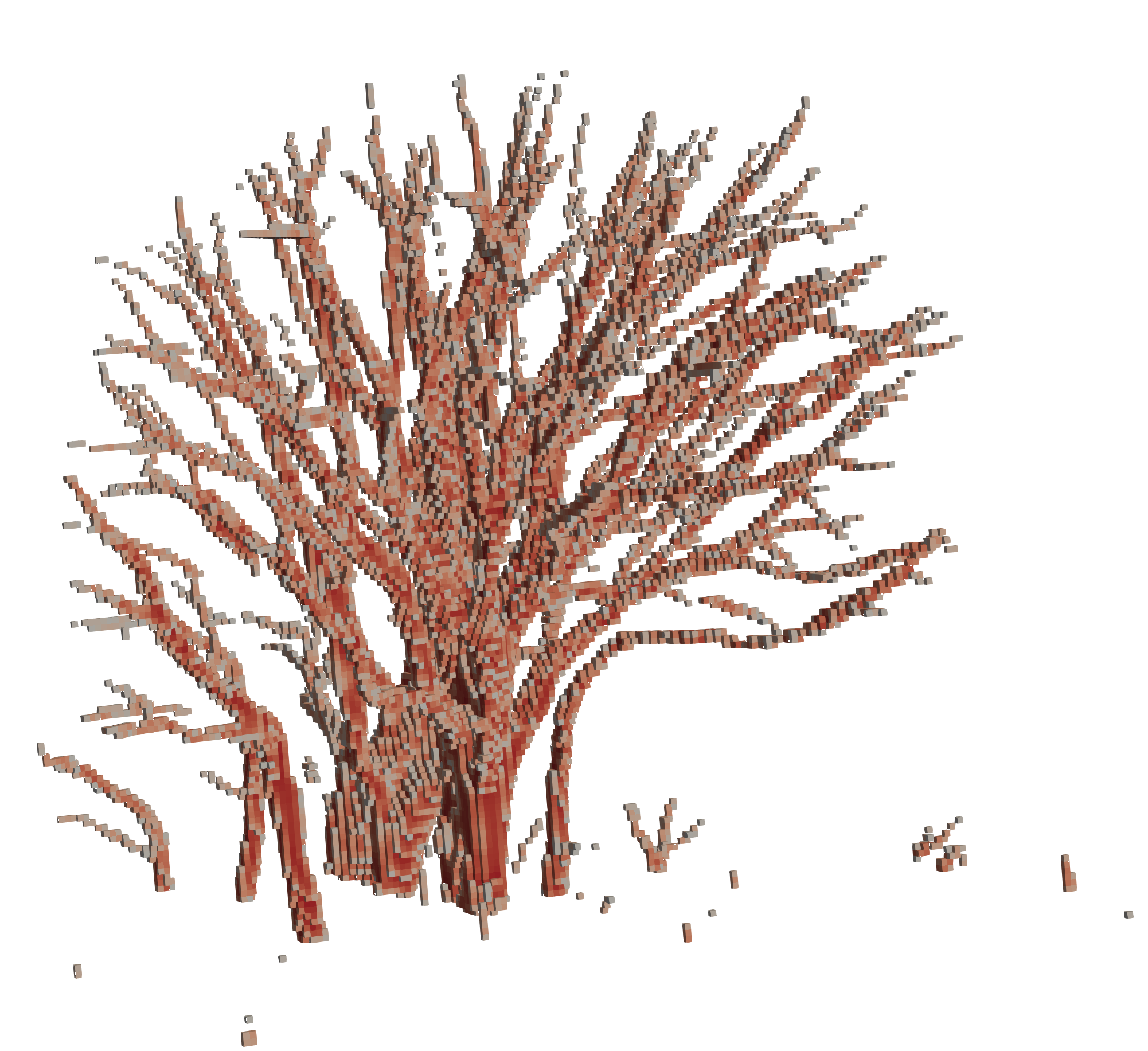} & 
    \includegraphics[height=\picheight cm]{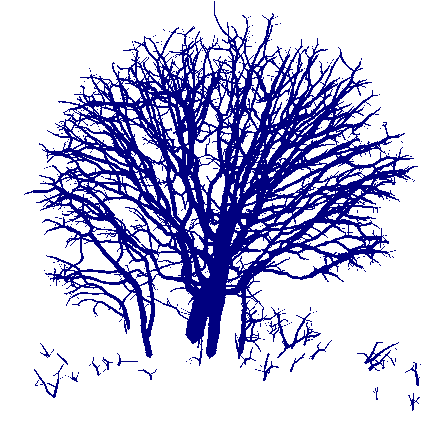} &
    \includegraphics[height=\picheight cm]{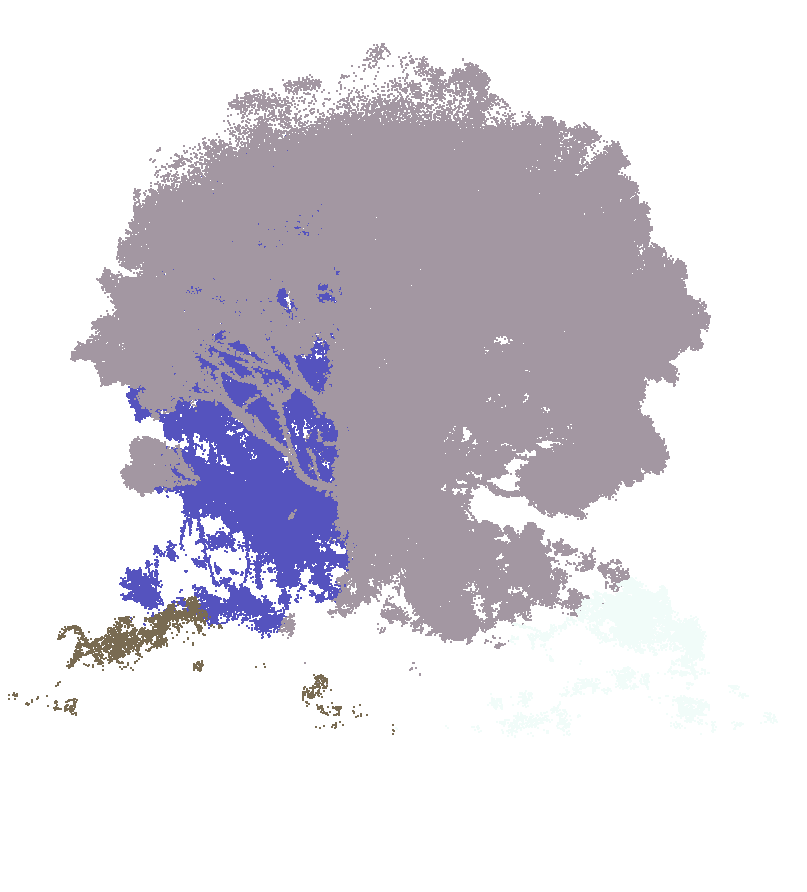} \\
    \shortstack{9\\Viewed on one\\side. Wall visible} & 
    \includegraphics[height=\picheight cm]{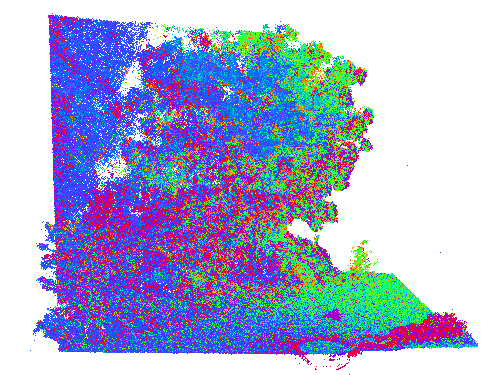} & 
    \includegraphics[height=\picheight cm]{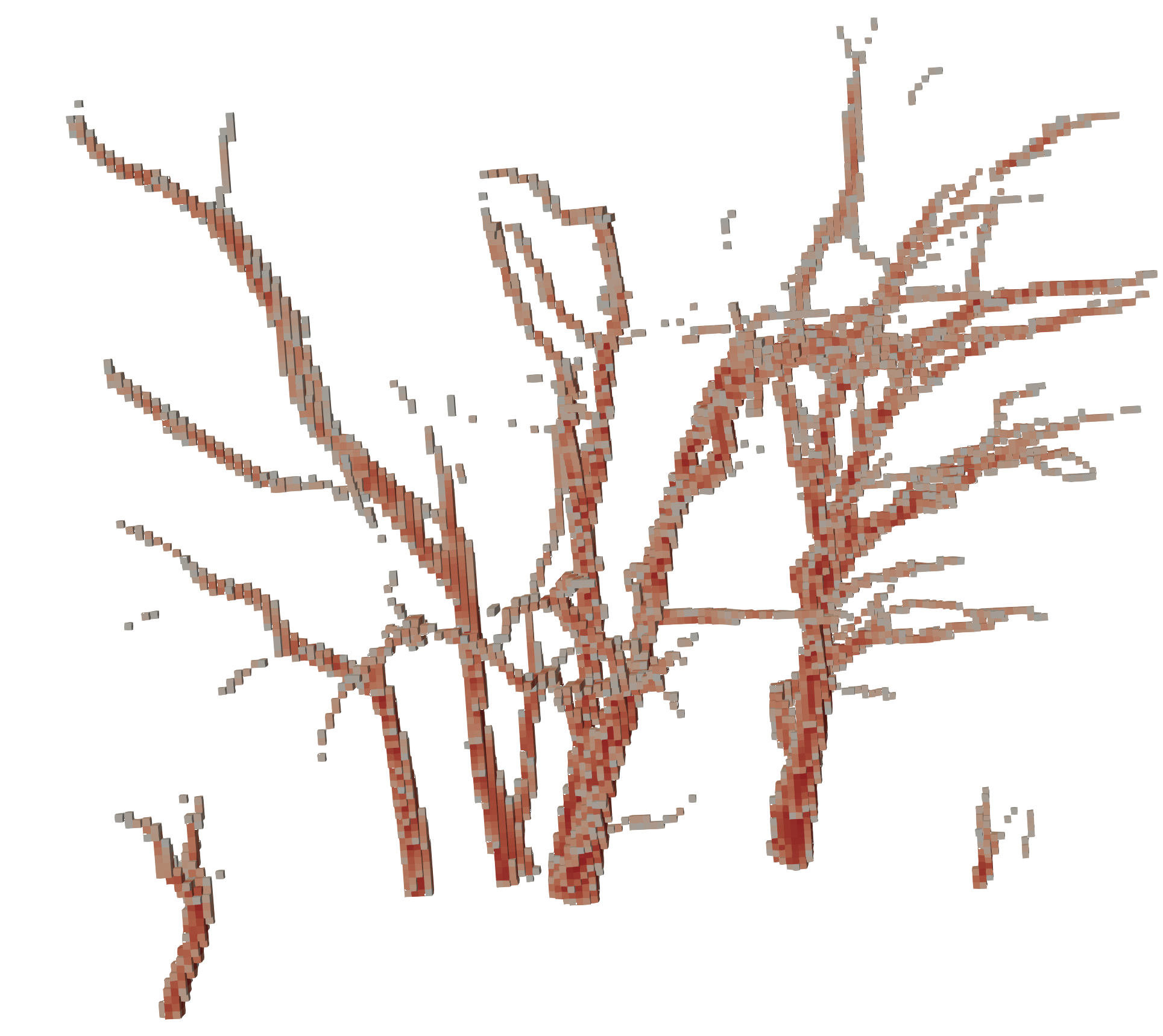} & 
    \includegraphics[height=\picheight cm]{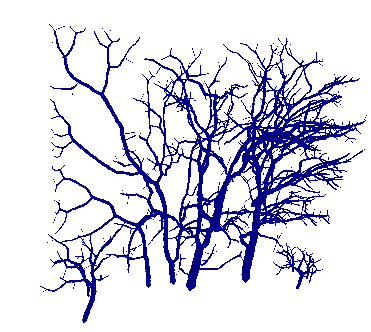} & 
    \includegraphics[height=\picheight cm]{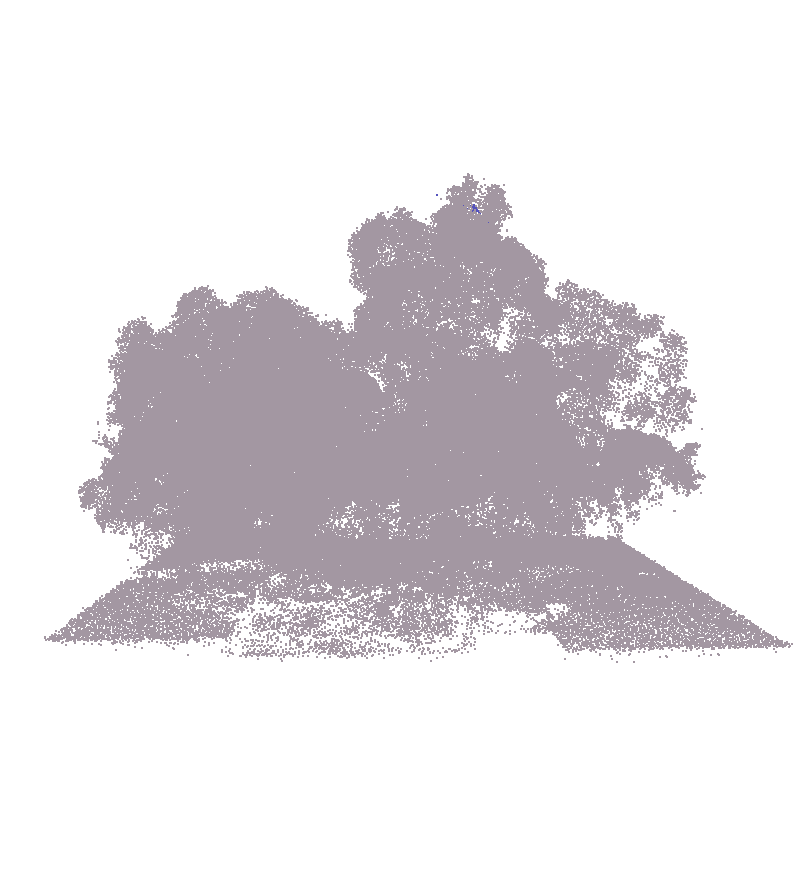} \\
    \shortstack{10\\Canopy reaches\\almost to ground} &
    \includegraphics[height=\picheight cm]{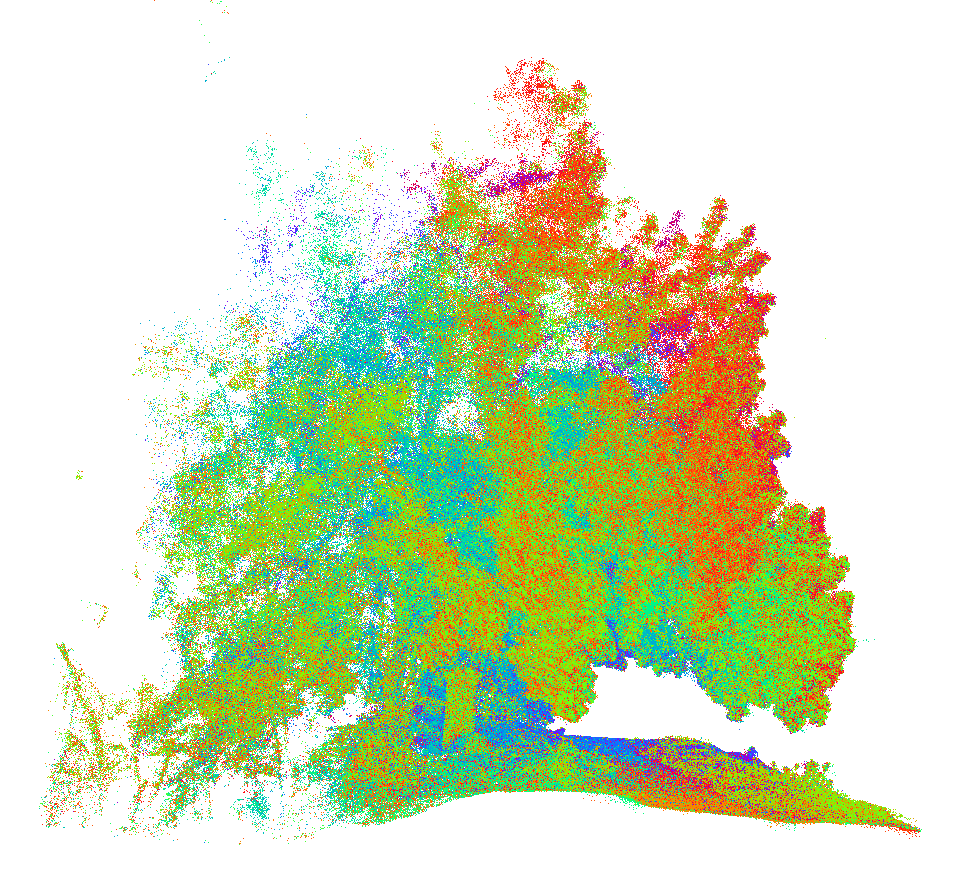} & 
    \includegraphics[height=\picheight cm]{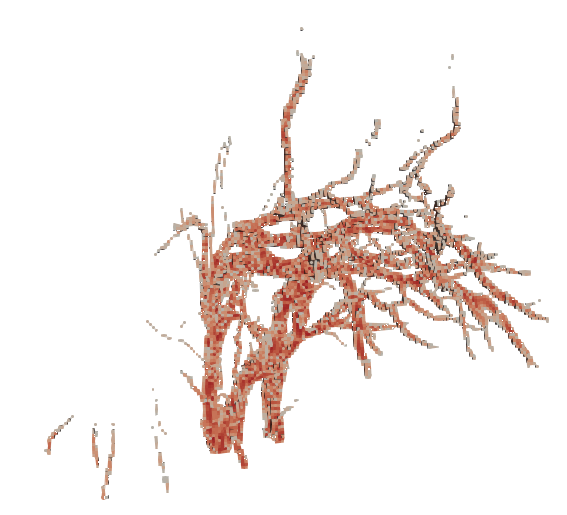} & 
    \includegraphics[height=\picheight cm]{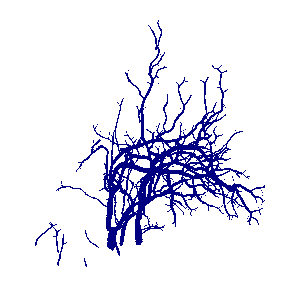} &
    \includegraphics[height=\picheight cm]{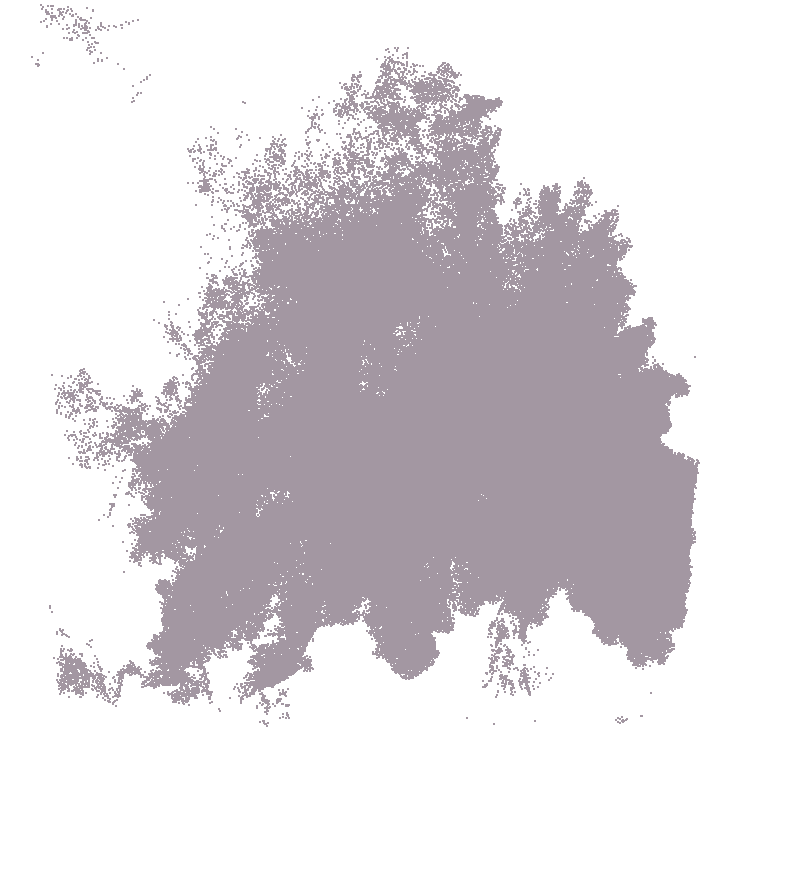} \\
    \shortstack{11\\Overlapping\\vegetation} & 
    \includegraphics[height=\widepicheight cm]{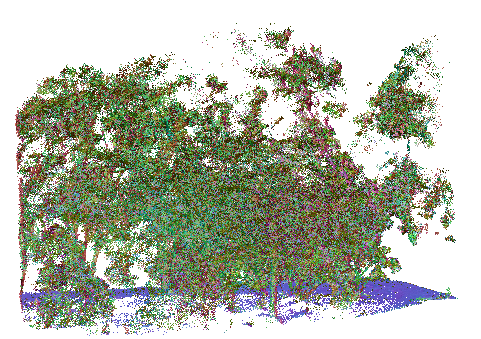} & 
    \includegraphics[height=\widepicheight cm]{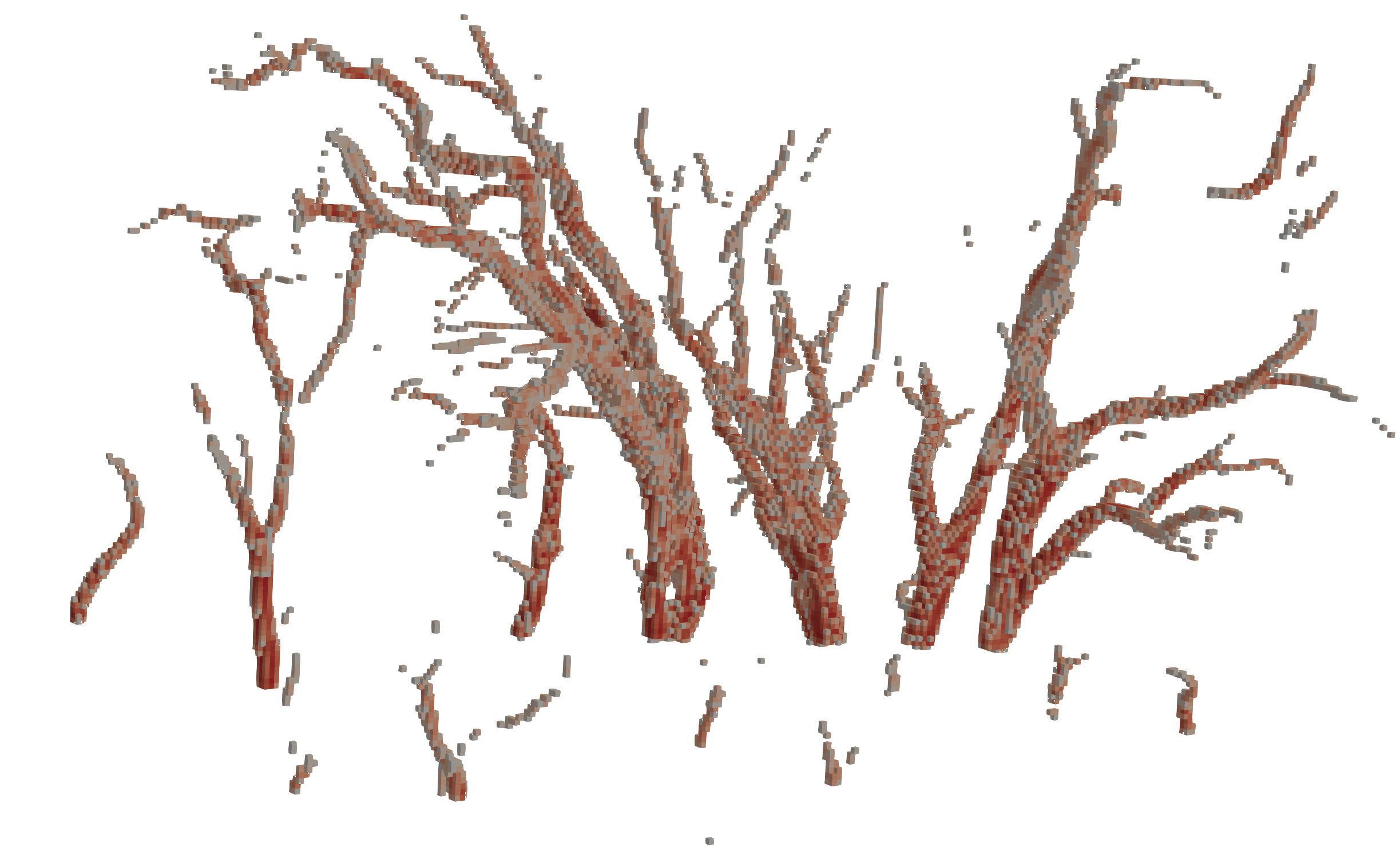} & 
    \includegraphics[height=\widepicheight cm]{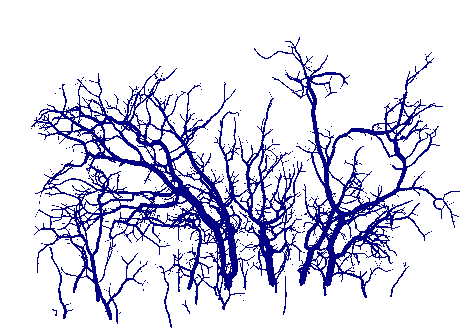} &
    \includegraphics[height=\widepicheight cm]{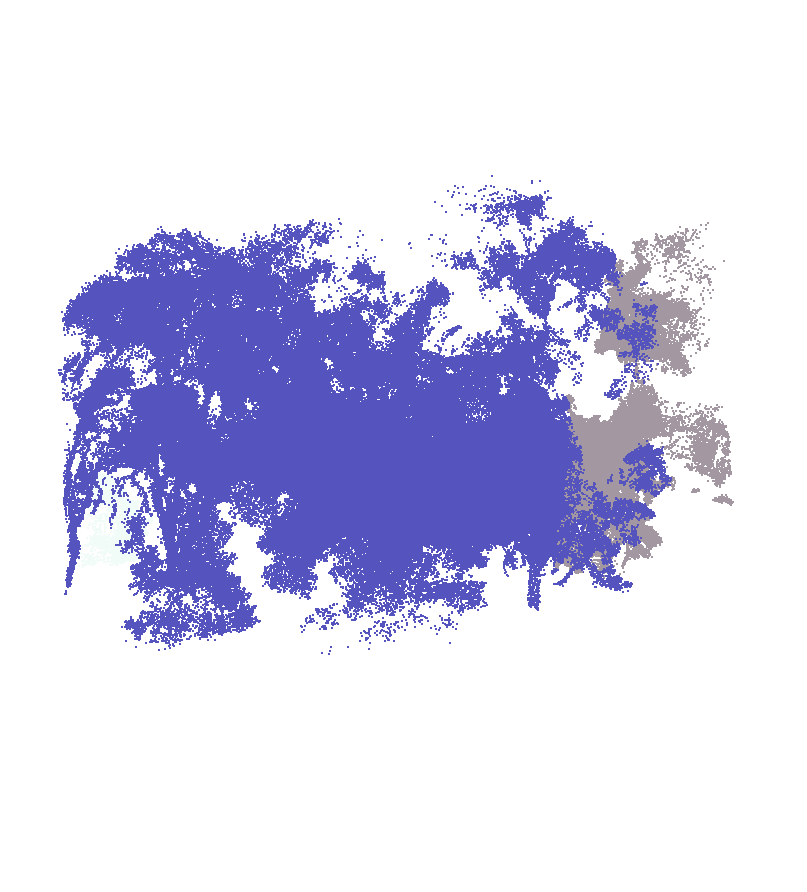} \\
    \shortstack{12\\Overlapping,\\close proximity\\trees} & 
    \includegraphics[height=\picheight cm]{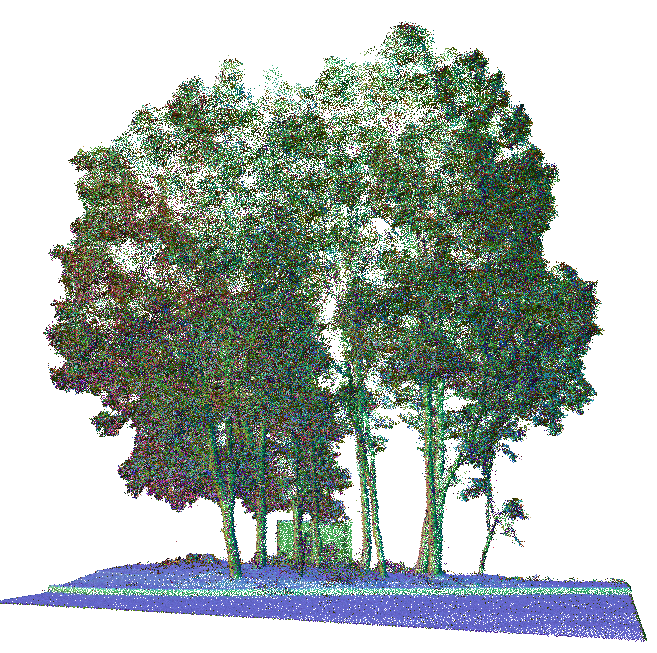} & 
    \includegraphics[height=\picheight cm]{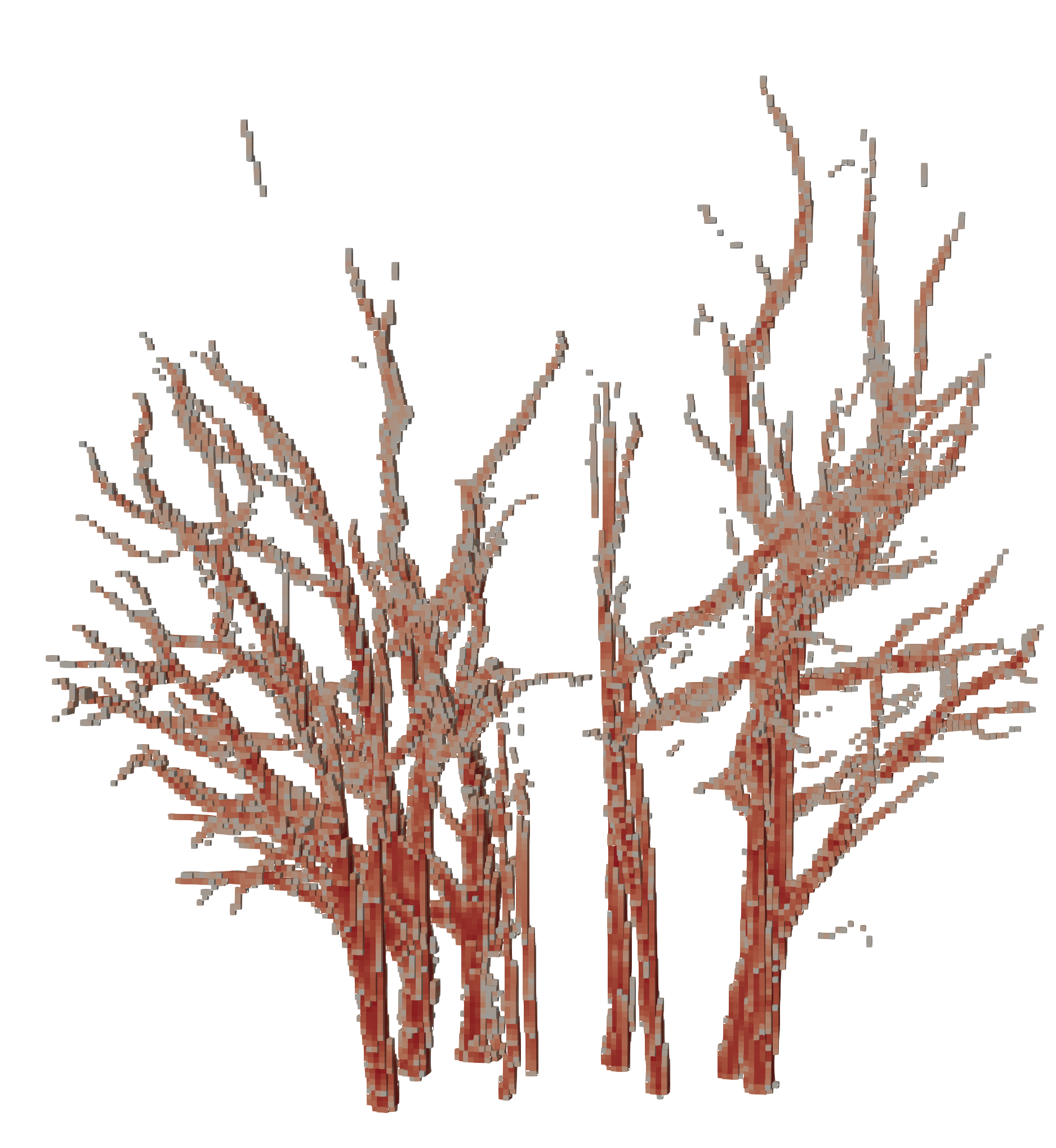} & 
    \includegraphics[height=\picheight cm]{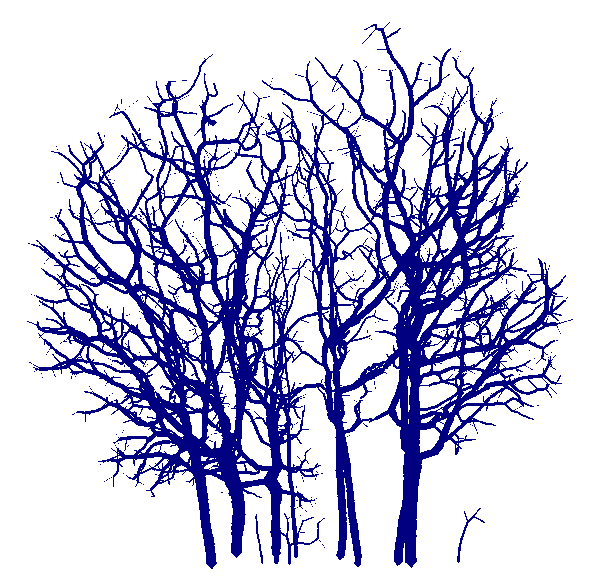} &
    \includegraphics[height=\picheight cm]{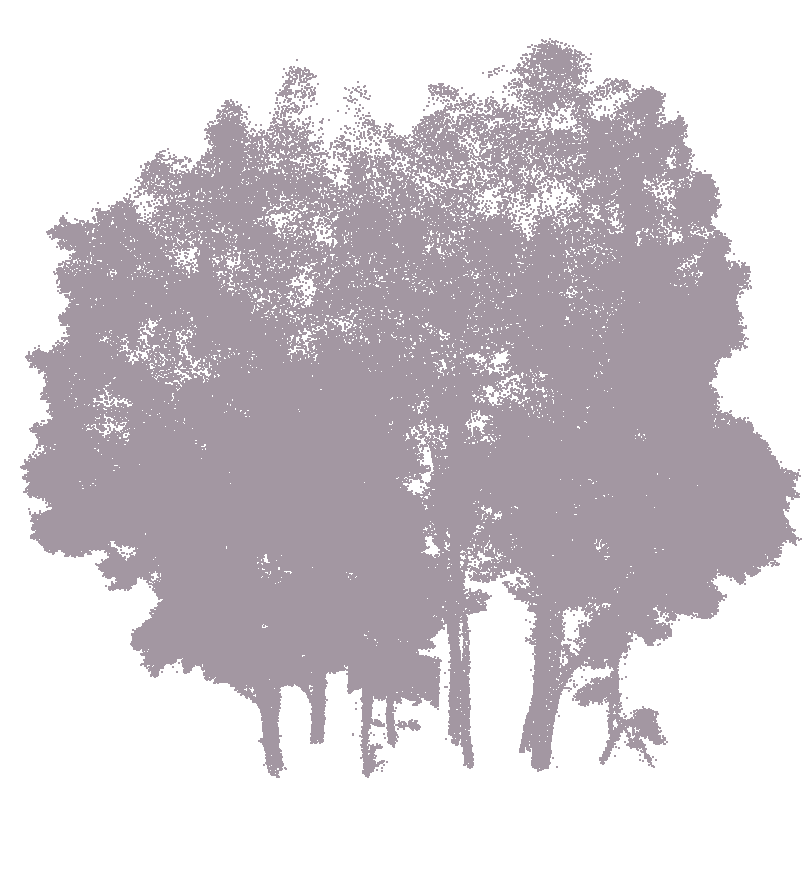} \\
    \shortstack{13\\non-singular,\\split trunk} & 
    \includegraphics[height=\picheight cm]{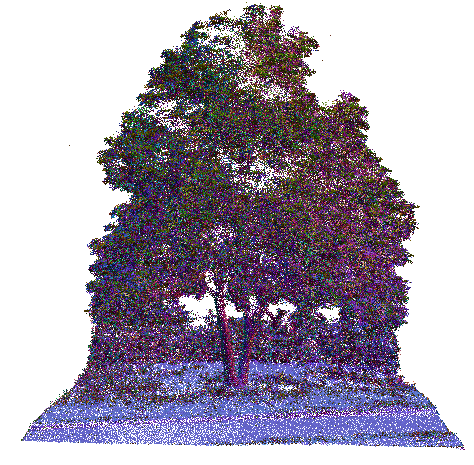} & 
    \includegraphics[height=\picheight cm]{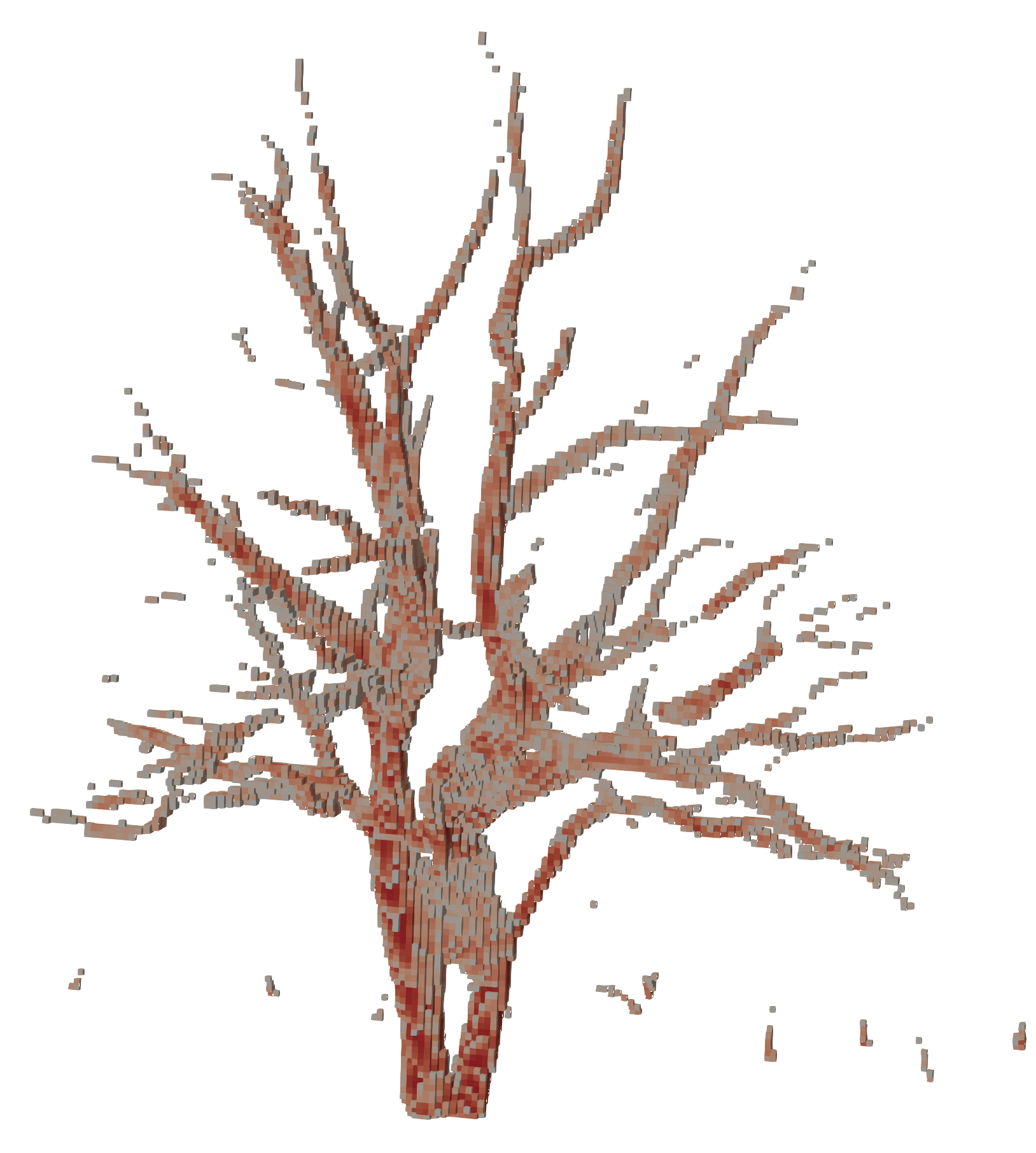} & 
    \includegraphics[height=\picheight cm]{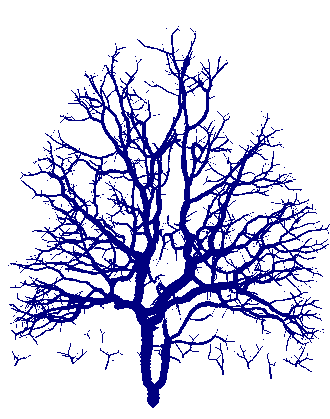} &
    \includegraphics[height=\picheight cm]{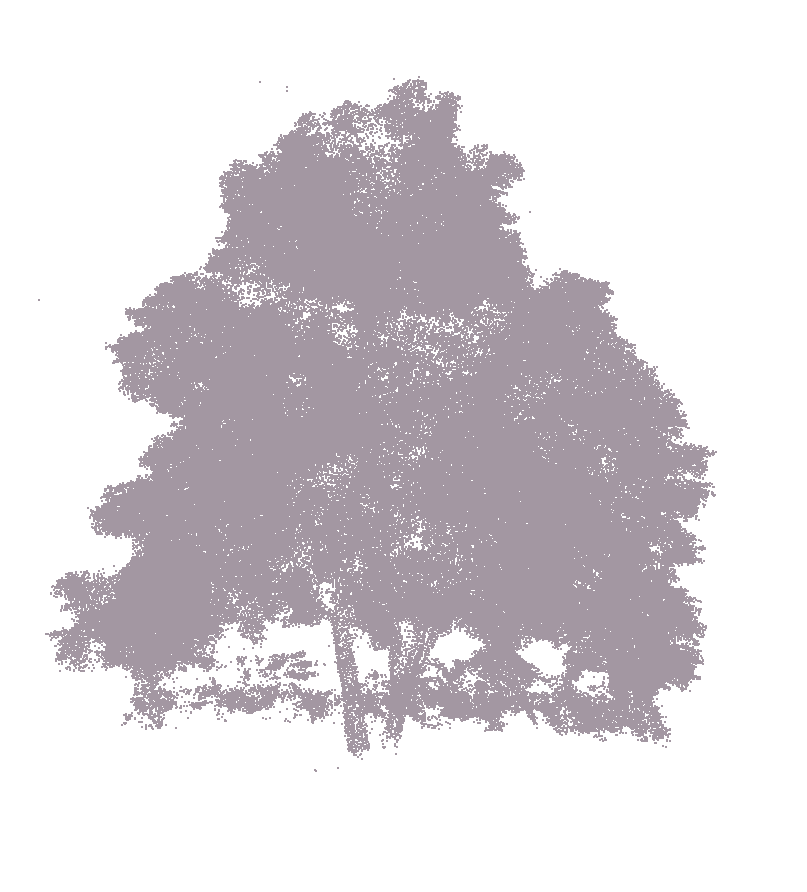} \\
    \end{tabular}
    \endgroup
    \label{tab:Qualitativeresults2}
    \end{center}
    \end{adjustwidth}
\end{table*}

\begin{table*}[]
\begin{adjustwidth}{-3cm}{-3cm}
\begin{center}
\caption{Quantitative comparison. For each case in Tables~\ref{tab:Qualitativeresults1} and \ref{tab:Qualitativeresults2} and  we present the Surface Error (SE) of our method to its point cloud, and compare the principle number of stems reconstructed with the number of segments produced from the popular Treeseg and TreeSeparation algorithms. Failed functions are marked with \xmark.} 
\begingroup
\setlength{\tabcolsep}{3pt} 
\begin{tabular}{c|c|c|c|c|c|c|c}
    Case & $w$ (cm) & SE (cm) & SE/$w$ & estimated cloud \#segs & our method \#segs & Treeseg \#segs & TreeSeparation \#segs \\
    \hline
    1 & 6 & 9.9 & 1.7 & 1 & 1 & \xmark & 1\\
    2 & 13 & 18.9 & 1.5 & 2 & 4 & 2 & 2\\
    3 & 16 & 18.6 & 1.2 & 3 & 4 & 1 & 2\\
    4 & 14 & 16.4 & 1.2 & 2 & 3 & 2 & 4\\
    5 & 13 & 21.0 & 1.6 & 2 & 4 & 1 & 3\\
    6 & 5 & 12.7 & 2.5 & 10 & 11 & \xmark & 2 \\
    7 & 10 & 17.8 & 1.8 & 7 & 8 & \xmark & 2\\
    8 & 9 & 16.6 & 1.8 & 5 & 4 & \xmark & 2\\
    9 & 5 & 10.6 & 2.1 & 4 & 5 & \xmark & 1\\
    10 & 5 & 13.6 & 2.7 & 6 & 8 & \xmark & 1\\
    11 & 6 & 16.9 & 2.8 & 10 & 8 & \xmark & 2\\
    12 & 6 & 13.1 & 2.2 & 9 & 9 & \xmark & 1\\
    13 & 4 & 10.7 & 2.7 & 1 & 1 & \xmark & 1\\
    \hline
    mean & 8.6 cm & 15.1 & 2.0 & 4.8 & 5.4 & - & 1.8 \\
    \end{tabular}
    \endgroup
    \label{tab:comparison}
    \end{center}
\end{adjustwidth}    
\end{table*}

\begin{table*}[]
\begin{adjustwidth}{-3cm}{-3cm}
\begin{center}
\caption{Ablation study: comparing results for two values of the branch angle parameter $\alpha$ on dense foliage that heavily restricts branch visibility}
\setlength{\tabcolsep}{3pt} 
\renewcommand{\arraystretch}{0.0} 
\begin{tabular}{c|c|c|c}
    Problem Case & Point Cloud & BSG $\alpha=1$ & BSG $\alpha=0$ \\
    \shortstack{14\\Dense canopy\\occludes all\\branches} & 
    \includegraphics[width=5.0 cm]{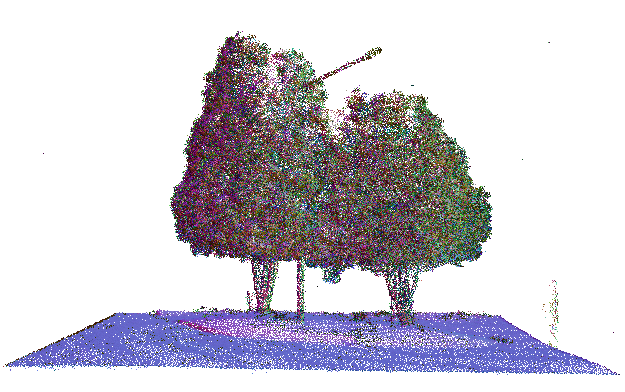} &  
    \includegraphics[height=\hardpicheight cm]{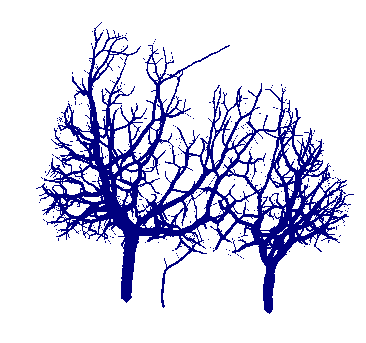} &
    \includegraphics[height=\hardpicheight cm]{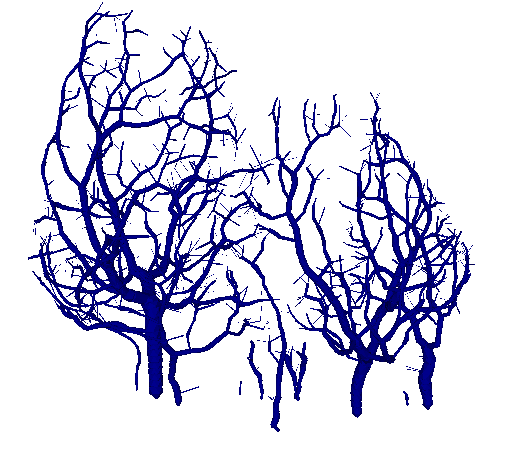}\\
    \shortstack{15\\Flat top} & 
    \includegraphics[width=5.0 cm]{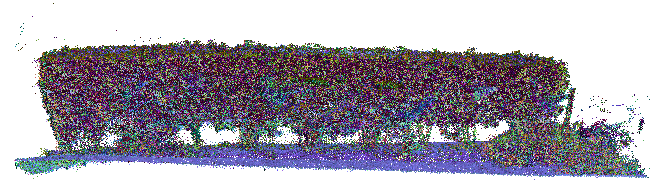} &  
    \includegraphics[width=5.0 cm]{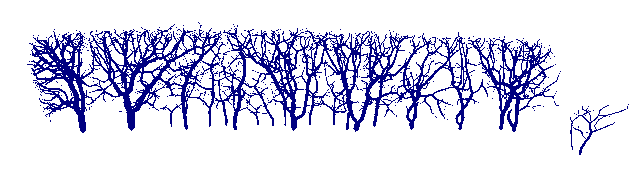}&
    \includegraphics[width=5.0 cm]{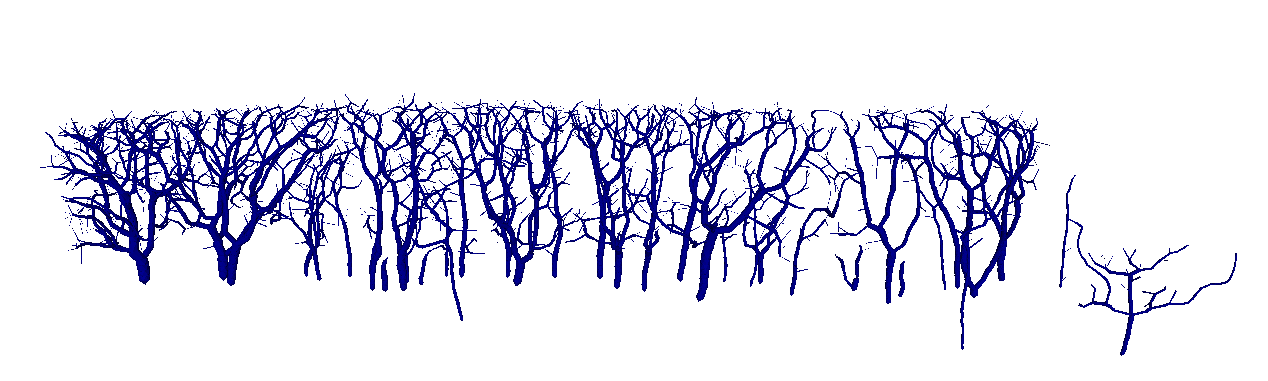}\\
    \shortstack{16\\Thickly \\ overlapping\\including palm} &
    \includegraphics[height=\hardpicheight cm]{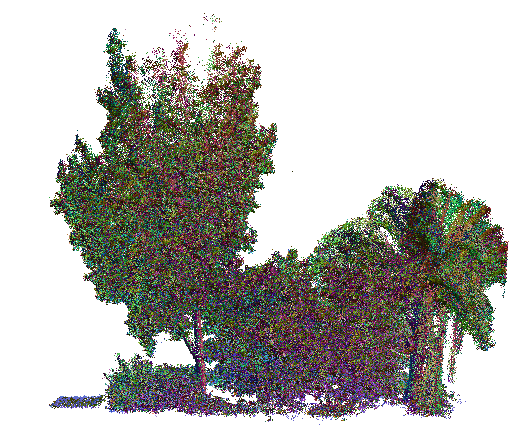} &  
    \includegraphics[height=\hardpicheight cm]{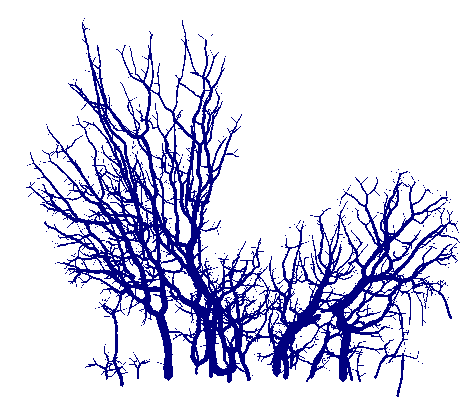}&
    \includegraphics[height=\hardpicheight cm]{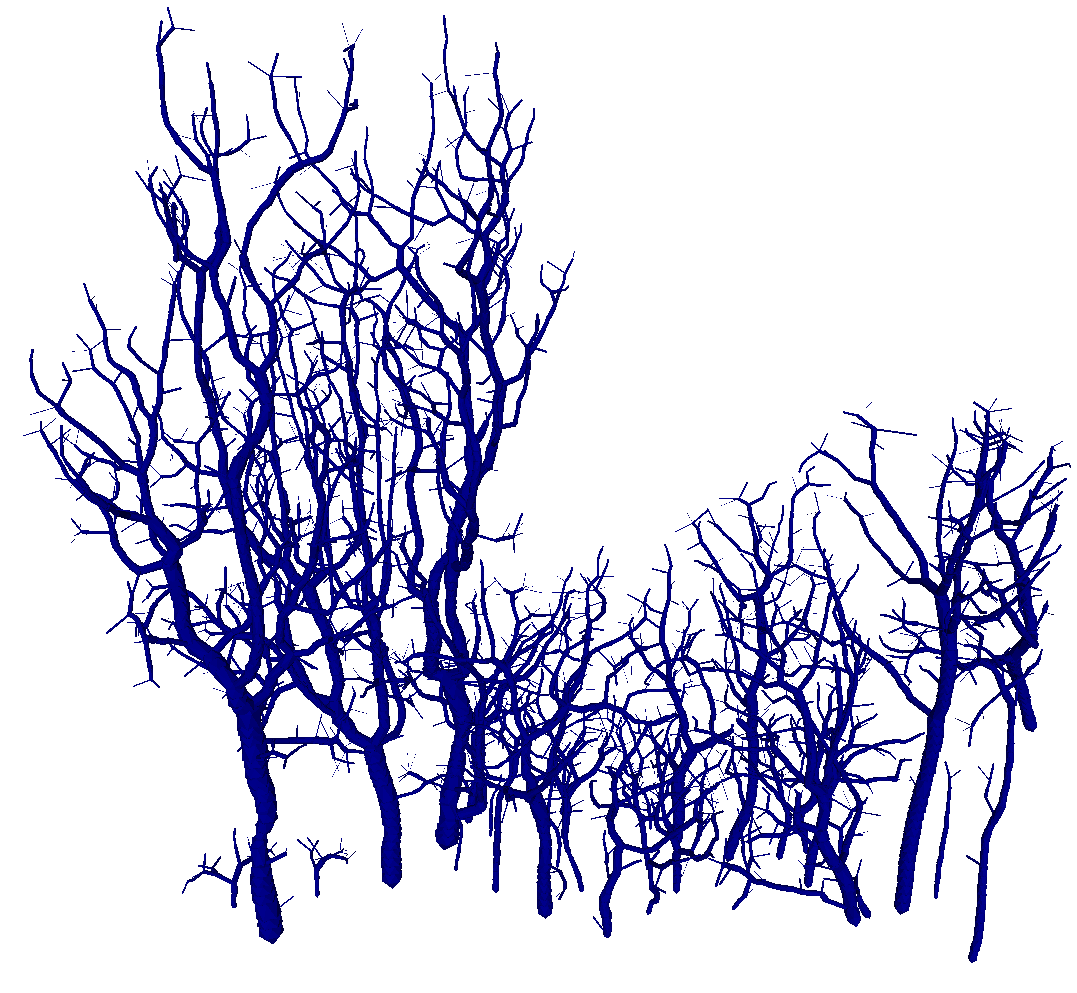} \\
    \shortstack{17\\Thick foliage\\occludes branches} &
    \includegraphics[height=\hardpicheight cm]{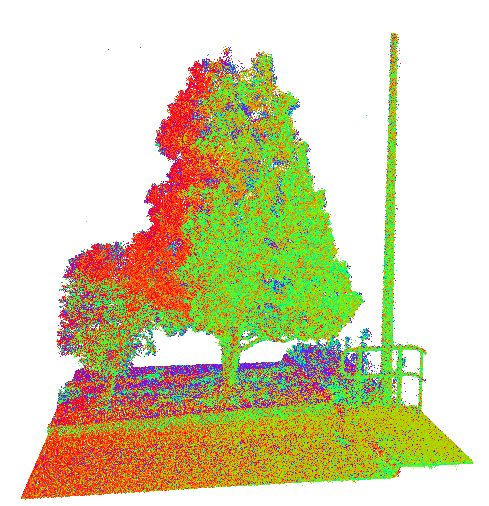} &  
    \includegraphics[height=\hardpicheight cm]{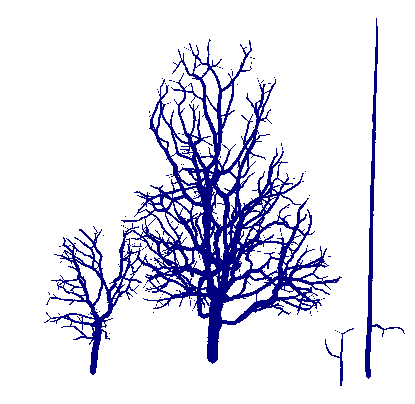}&
    \includegraphics[height=\hardpicheight cm]{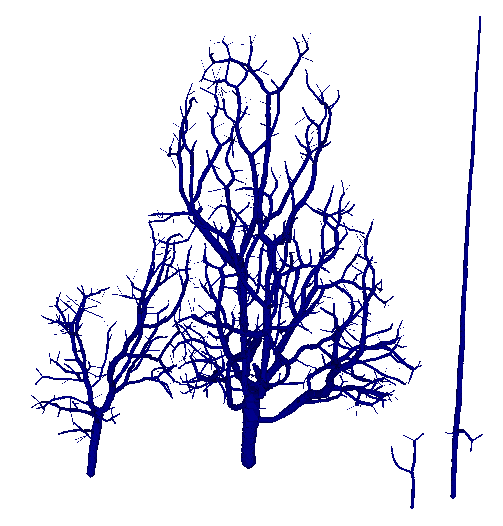}\\
    \shortstack{18\\tree tucked\\under others.\\Wall visible} & 
    \includegraphics[height=\hardpicheight cm]{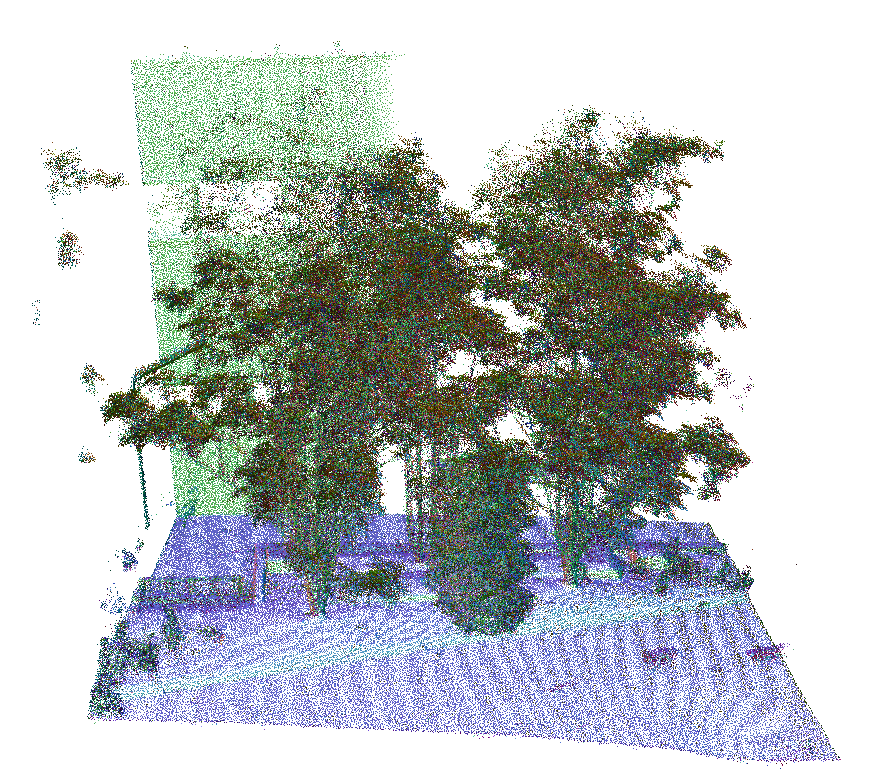}  &  
    \includegraphics[height=\hardpicheight cm]{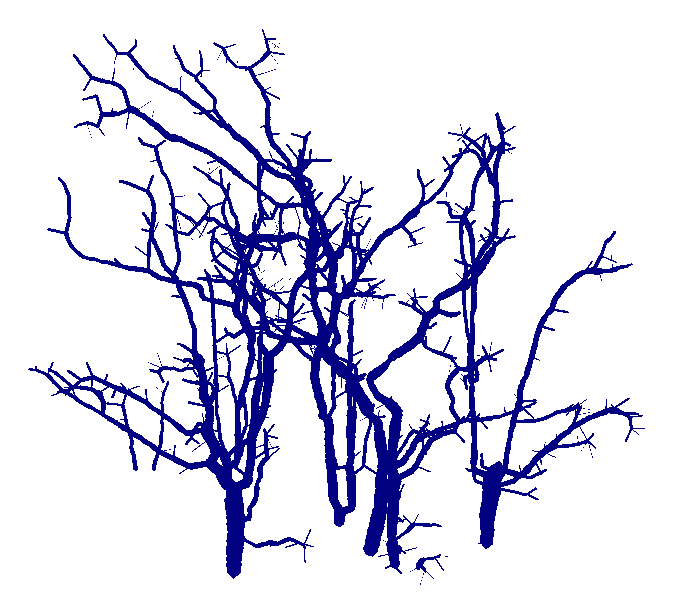} &
    \includegraphics[height=\hardpicheight cm]{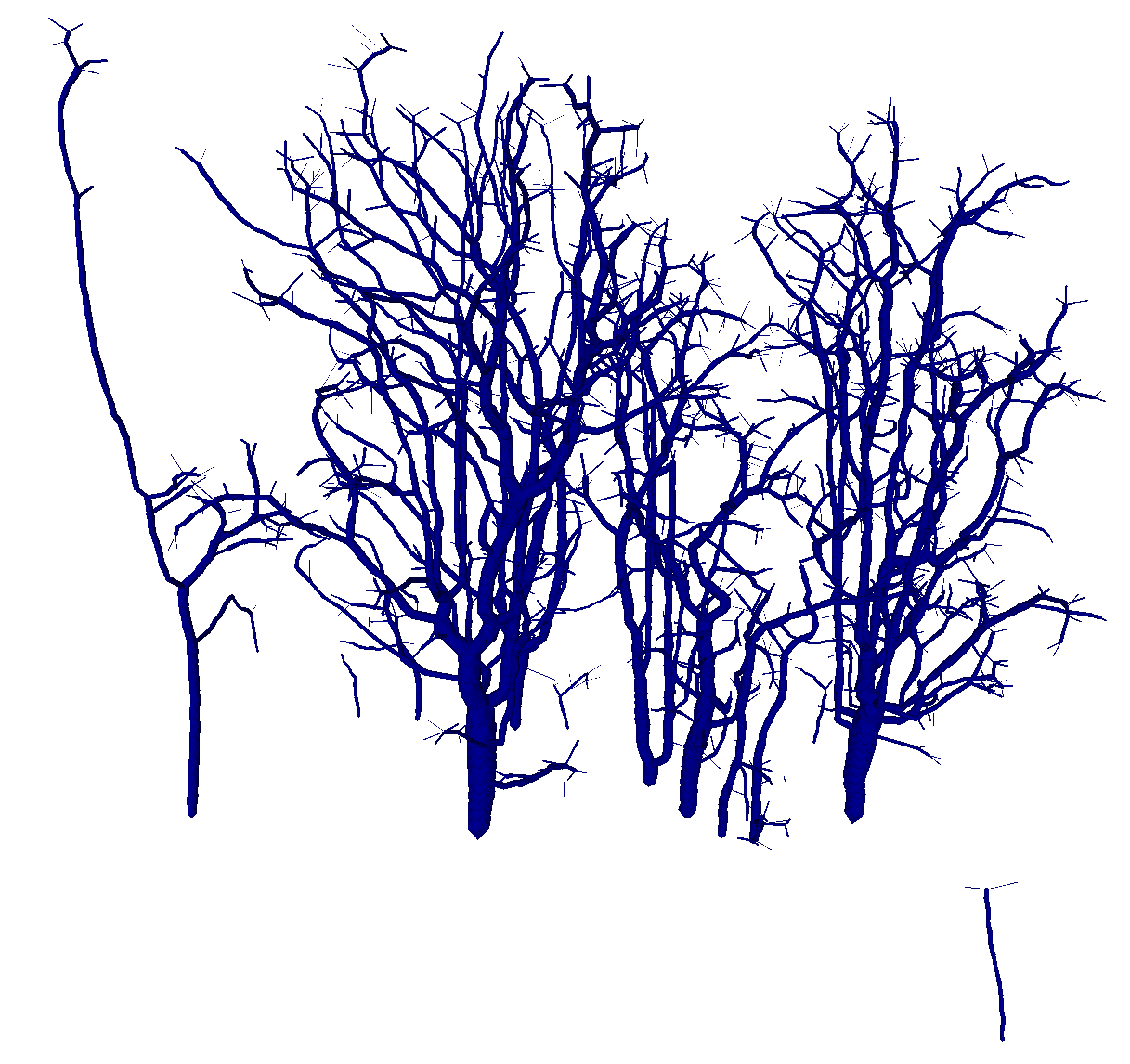}\\
    \shortstack{19\\Steep ground,\\undergrowth} & 
    \includegraphics[height=\hardpicheight cm]{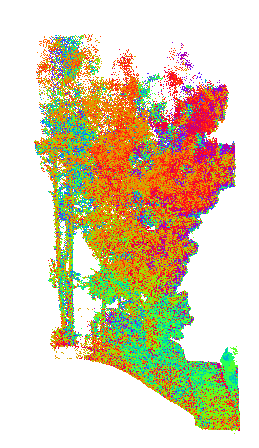}  &  
    \includegraphics[height=\hardpicheight cm]{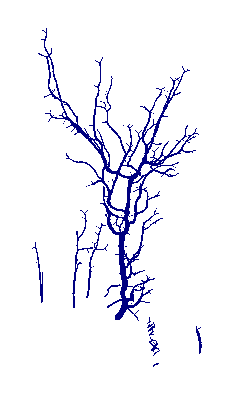}&
    \includegraphics[height=\hardpicheight cm]{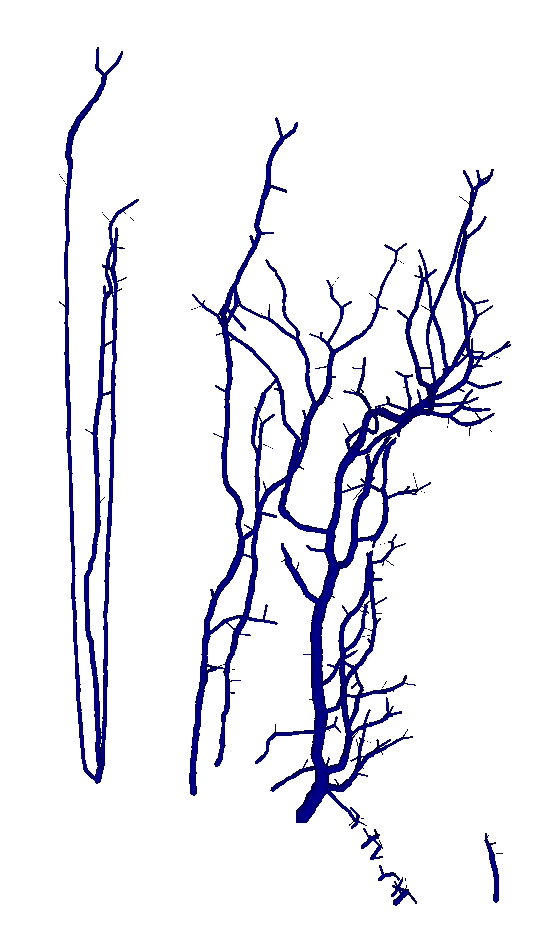}
    \\
    \end{tabular}
    \label{tab:hardcases2}
    \end{center}
\end{adjustwidth}    
\end{table*}

\begin{table*}[]
\begin{adjustwidth}{-3cm}{-3cm}
\caption{Aerial comparison: of a tree reconstructed from high-density ground data and sparse aerial data of a Moreton Bay fig tree}
\begin{center}
\begin{tabular}{c|c|c|c}
    Problem Case & Point Cloud & BSG & Superimposed \\
    \shortstack{20\\Thick Foliage\\Ground Scan} & 
    \includegraphics[height=\picheight cm]{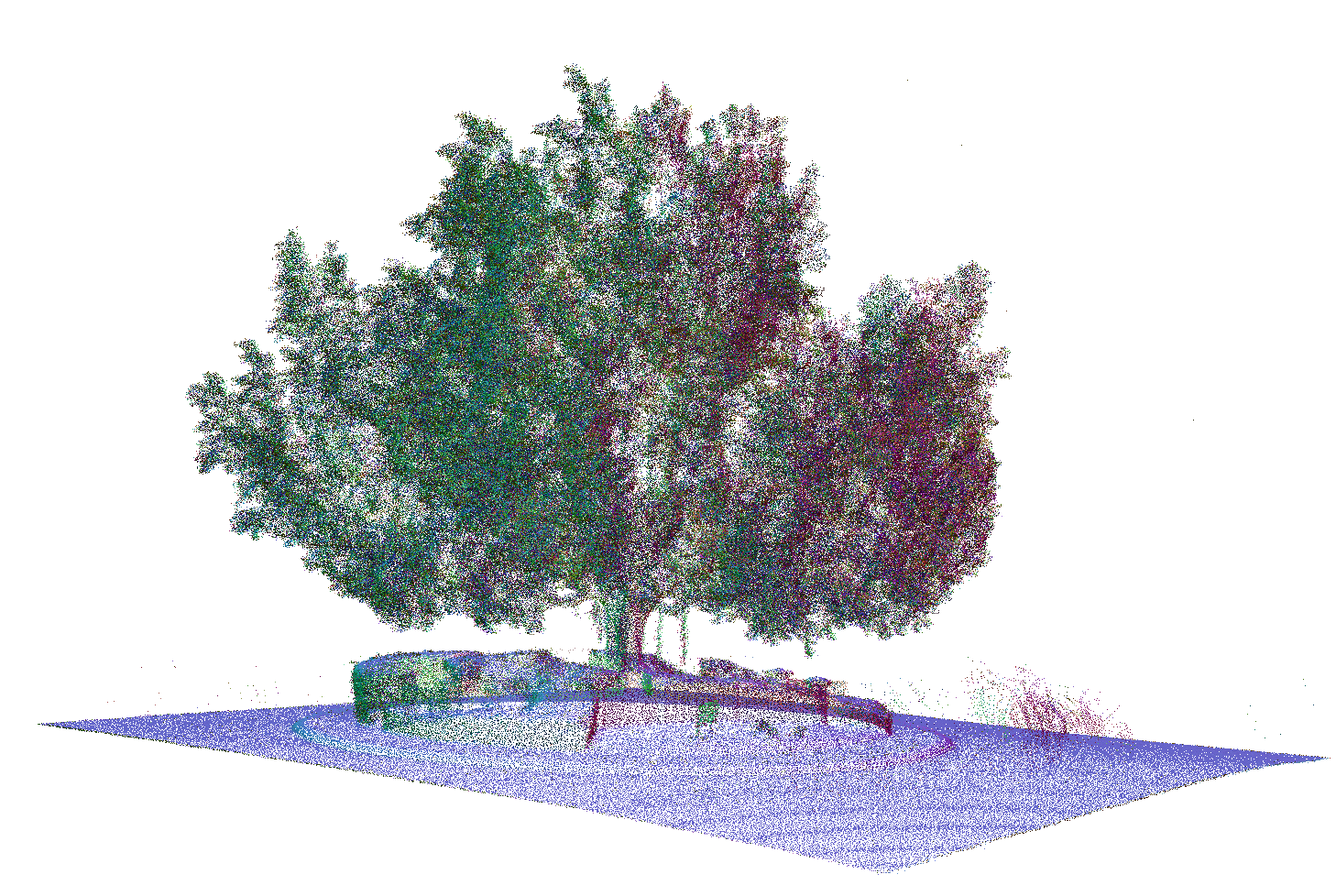} & 
    \includegraphics[height=\picheight cm]{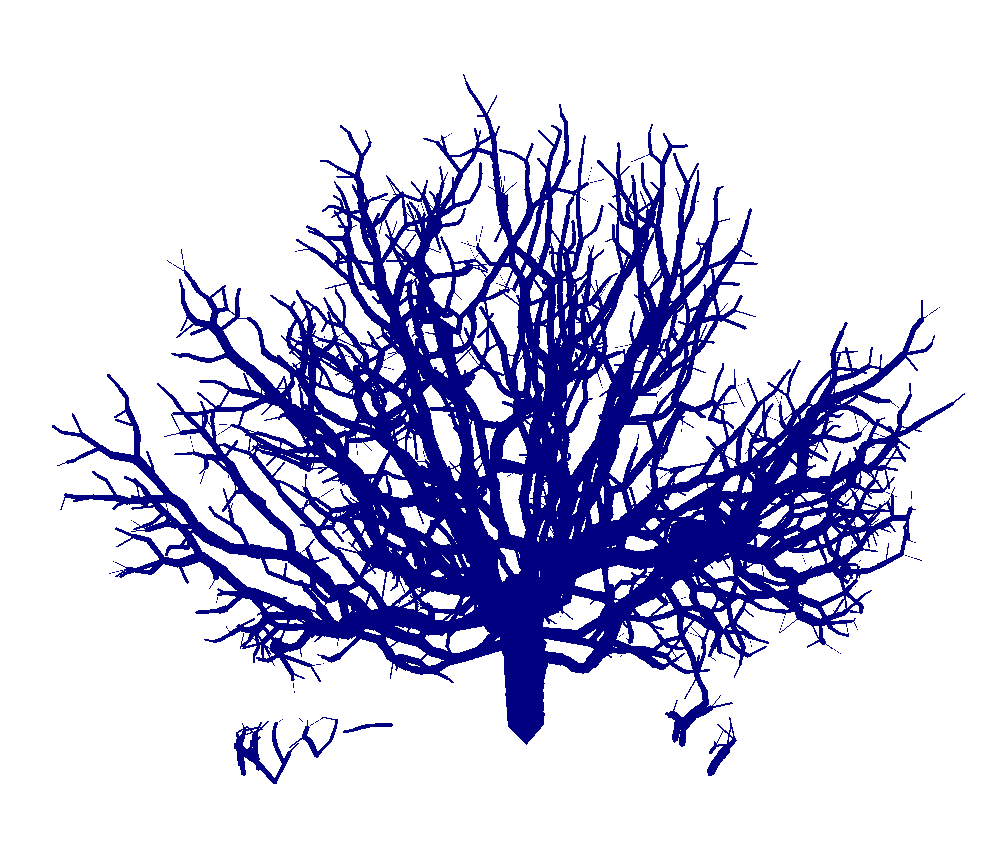} & 
    \includegraphics[height=\picheight cm]{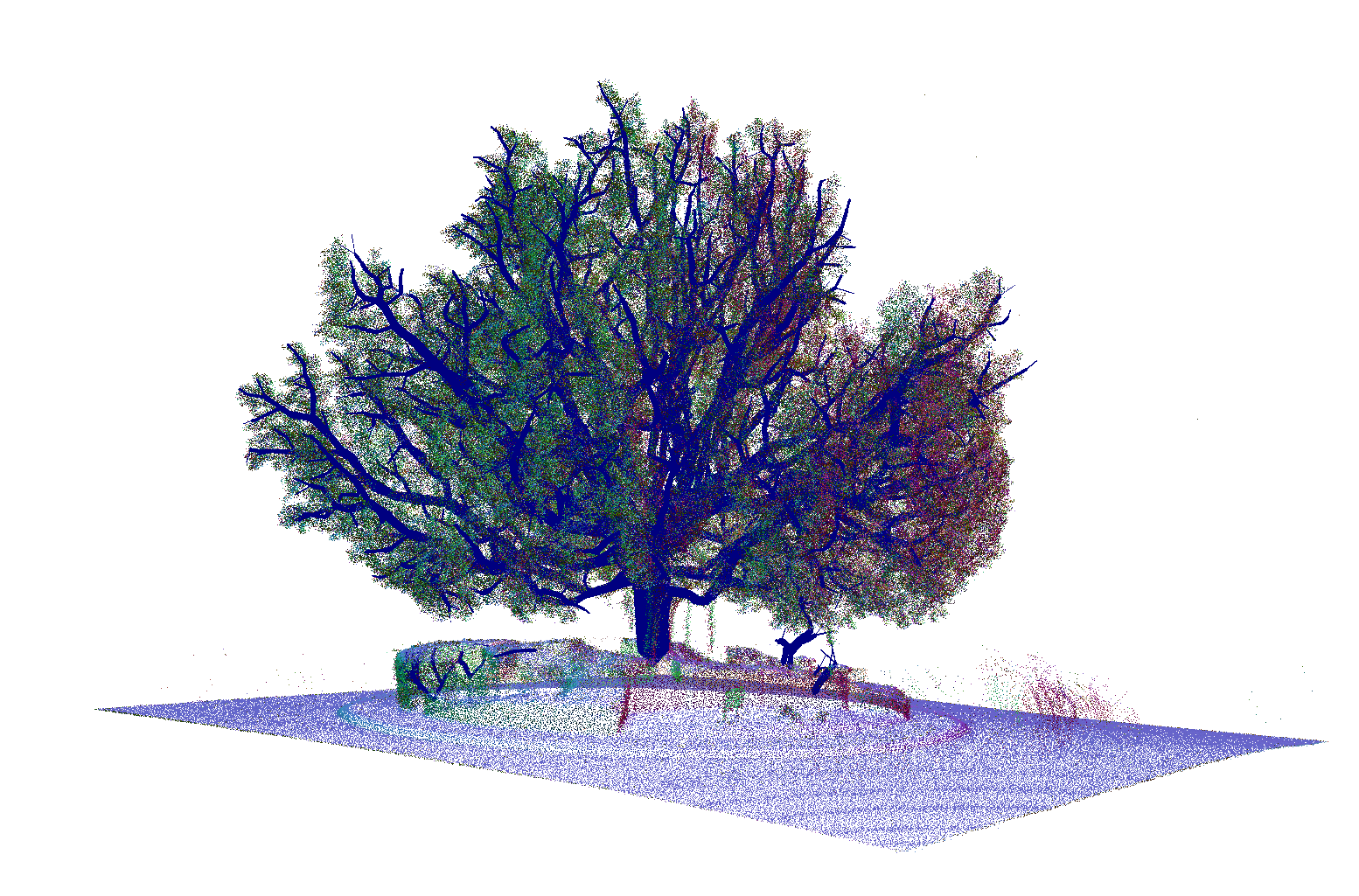} \\
    \shortstack{21\\Thick Foliage\\Aerial Scan} &
    \includegraphics[height=\picheight cm]{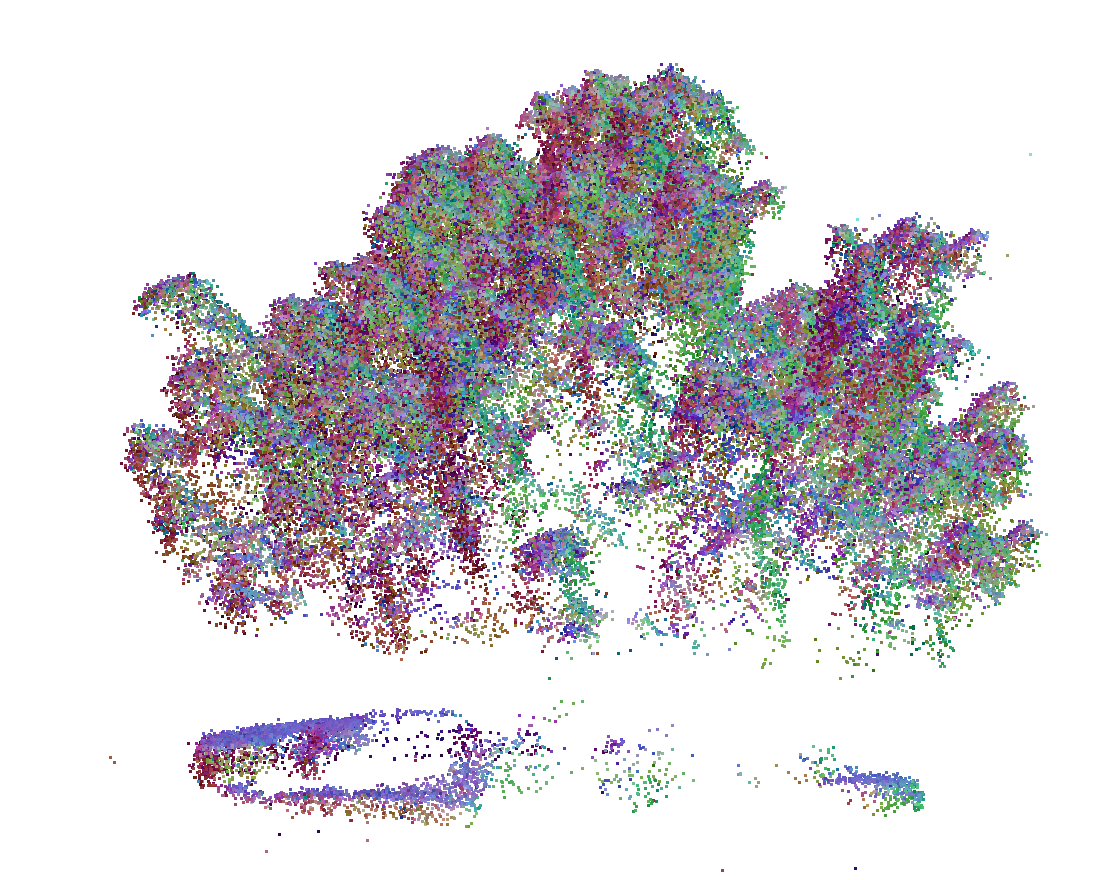} & 
    \includegraphics[height=\picheight cm]{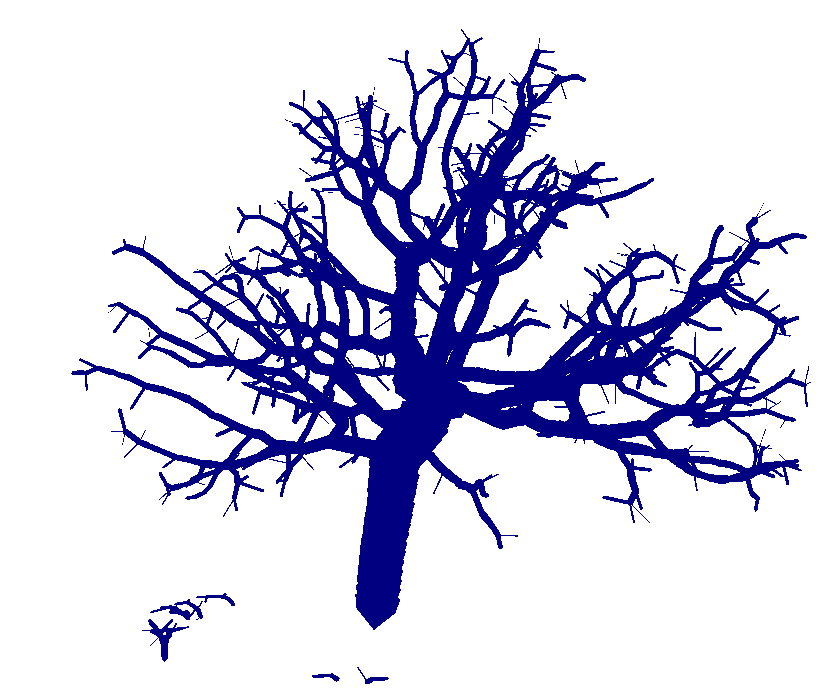} & 
    \includegraphics[height=\picheight cm]{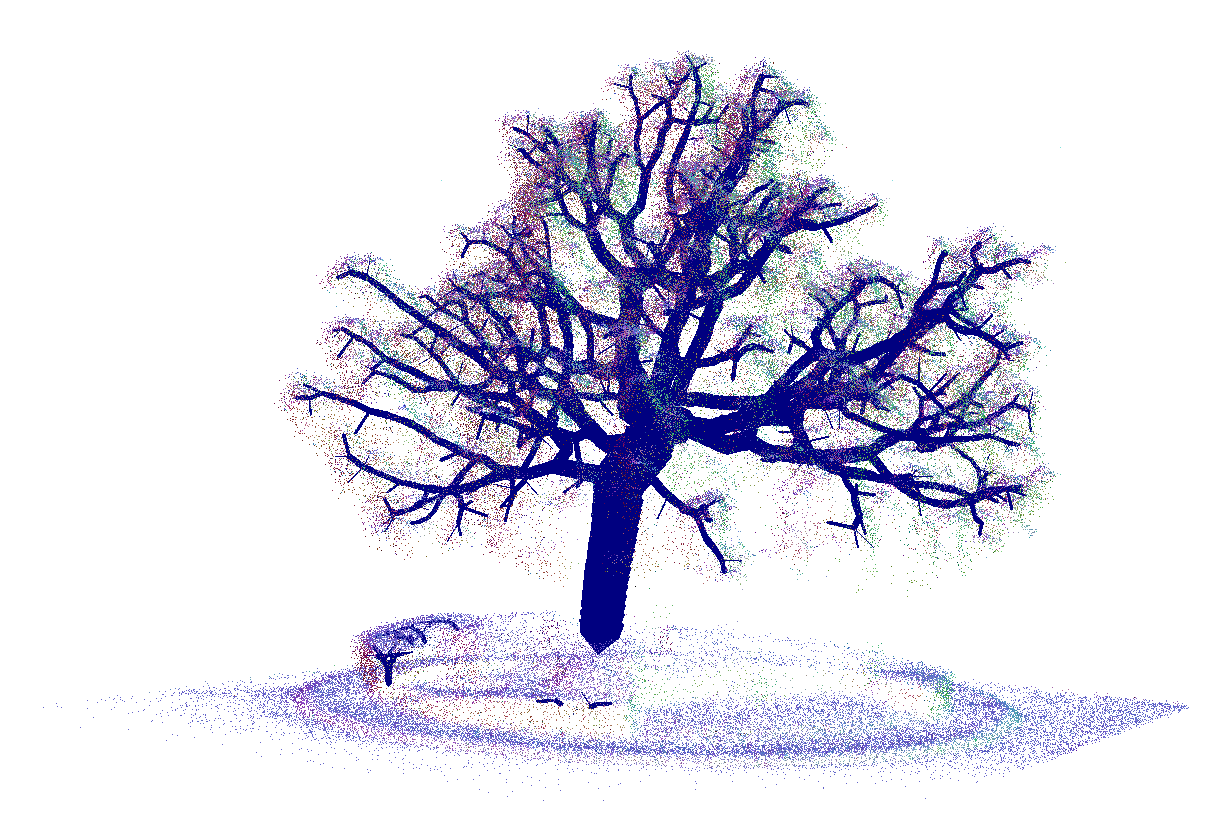}\\
    \end{tabular}
    \label{tab:aerialscan}
    \end{center}
\end{adjustwidth}    
\end{table*}

\begin{table*}[]
\begin{adjustwidth}{-3cm}{-3cm}
\begin{center}
\caption{Occlusion study: the optimisation procedure is repeated after removing spheres ranging in diameter from 1/8 to 1/2 of each trees height (red) from their geometric centre, replicating an occlusion. Even with a significant proportion of point removed from the cloud, little change is observed in the reconstructed trees (blue).}
\begin{tabular}{c|c|c|c|c|c}
    & & \multicolumn{4}{c}{Removed Sphere Diameter (relative to tree height)}\\
    Test Case & \shortstack{Original\\  Point Cloud} & 1/8 & 2/8 & 3/8 & 4/8\\
    \hline
    SE (cm) & 20.3 & 20.9 & 24.5 & 32.7 & 49.0 \\
    \hline
    SE/$w$ & 0.9 & 0.9 & 1.1 & 1.4 & 2.1\\
    \hline
    &\includegraphics[height=\spherepicheight cm]{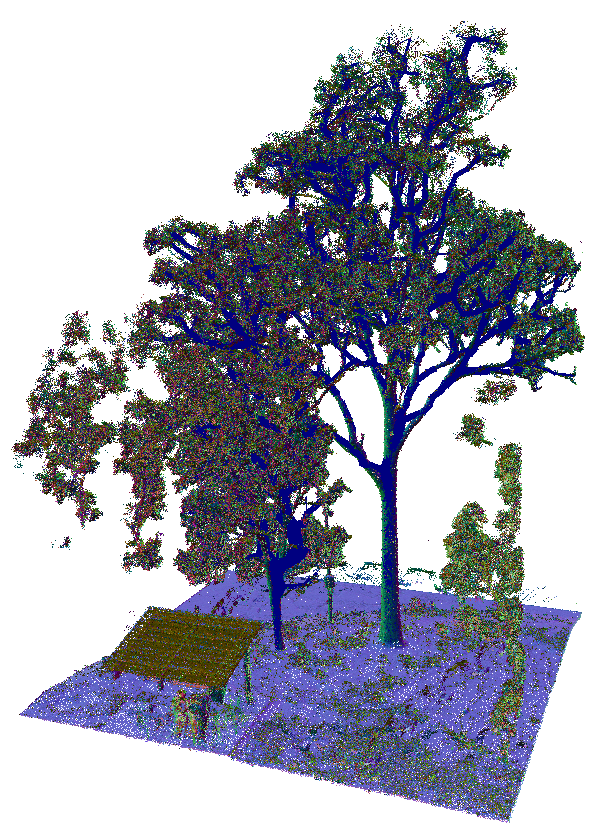} & \includegraphics[height=\spherepicheight cm]{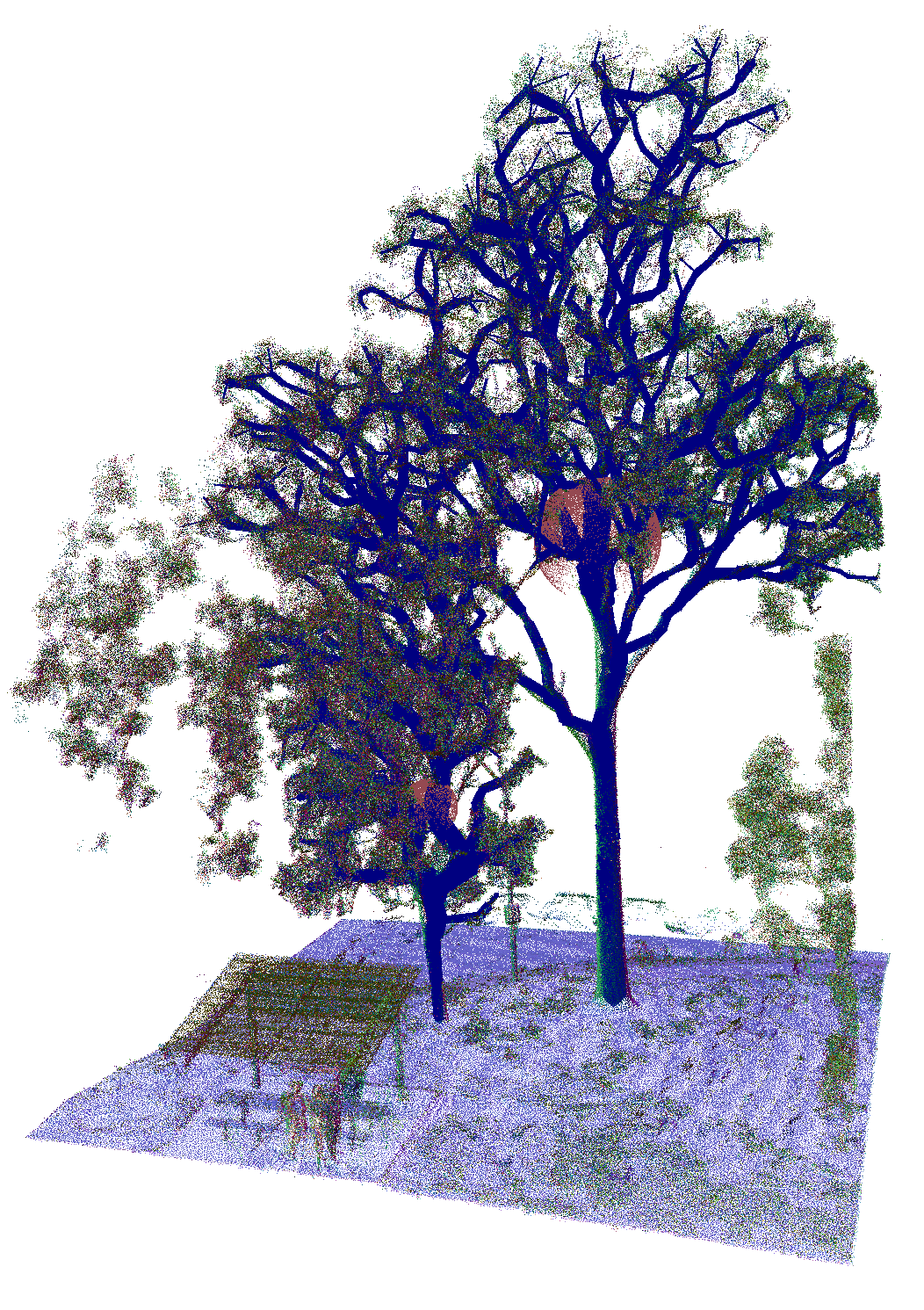} & \includegraphics[height=\spherepicheight cm]{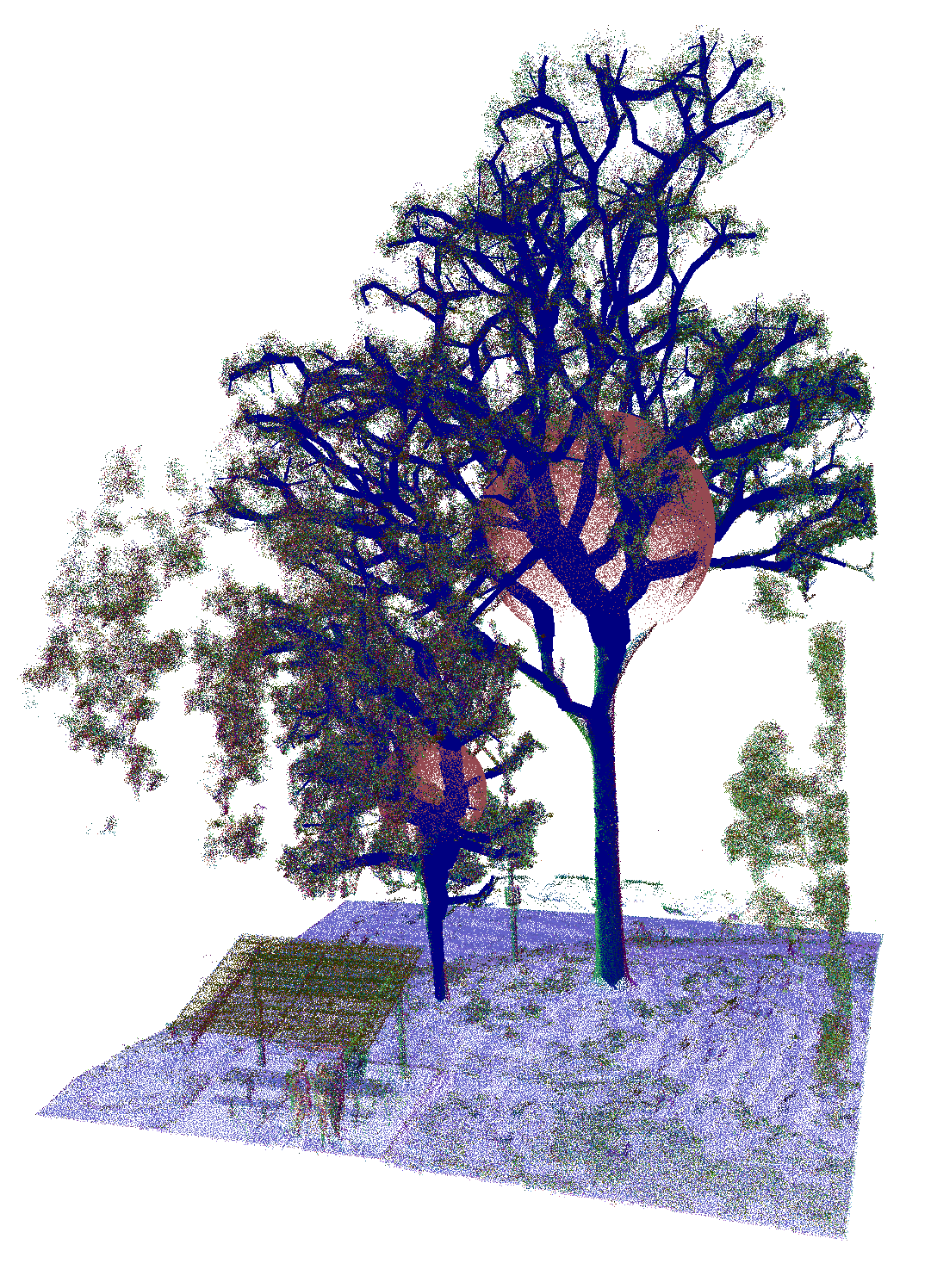} & \includegraphics[height=\spherepicheight cm]{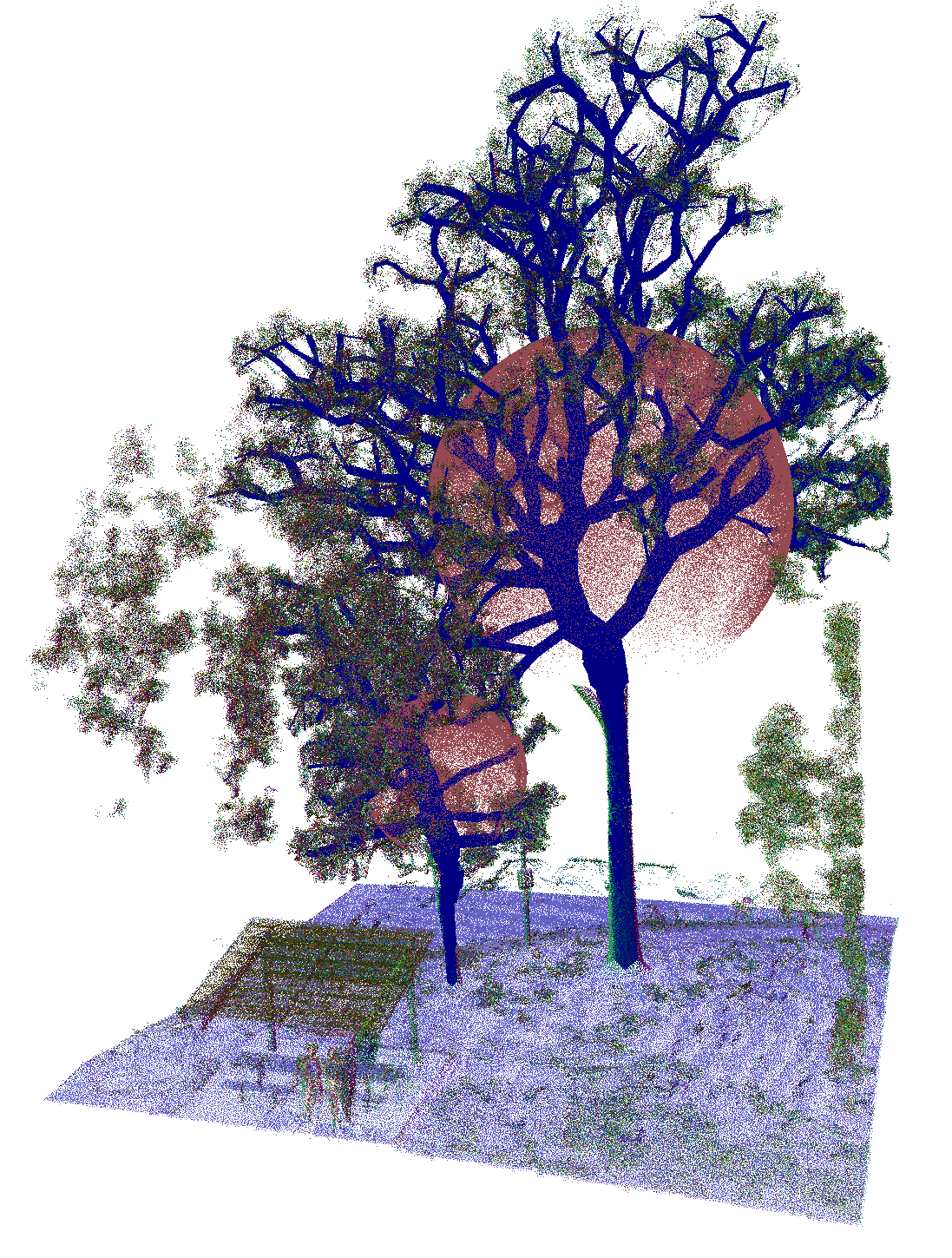} & \includegraphics[height=\spherepicheight cm]{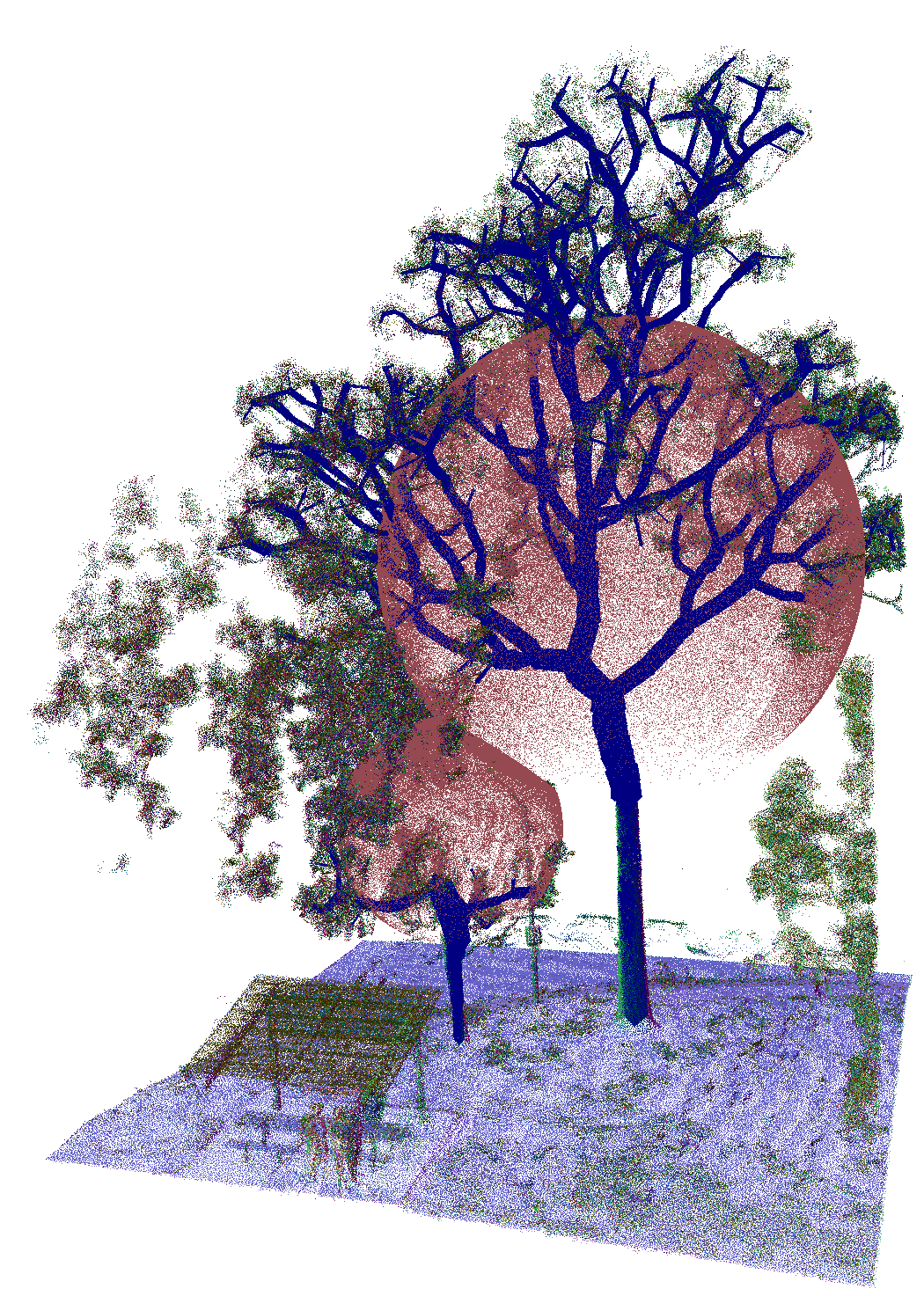} \\
\end{tabular}
\label{tab:sphereexperiment}
\end{center}
\end{adjustwidth}
\end{table*}

\section{Results and Discussion}\label{sec:results}
\subsection{Qualitative Study: Tables~\ref{tab:Qualitativeresults1} and \ref{tab:Qualitativeresults2}}
\label{sec:qualitaive}
The quality of the tree reconstruction is first evaluated qualitatively, by visually presenting the topology optimised density (voxel) grid and reconstructed BSG for thirteen datasets.
The visual representation enables a detailed assessment of the whole dataset as well as local features, which may be lost when comparing quantitatively.
These results are visually compared to the state of the art TreeSeparation segmentation method~\cite{wang2018scalable}.
The thirteen sets of trees shown are cases which are relatively simple to reconstruct, their features are generally visible and have low levels of occlusion.
These same thirteen datasets are quantitatively evaluated in Section~\ref{sec:quantitaive}.

\begin{figure}[ht]
\begin{adjustwidth}{-3cm}{-3cm}
\centering
\subfigure[]{\includegraphics[height=\monkeypicheight cm]{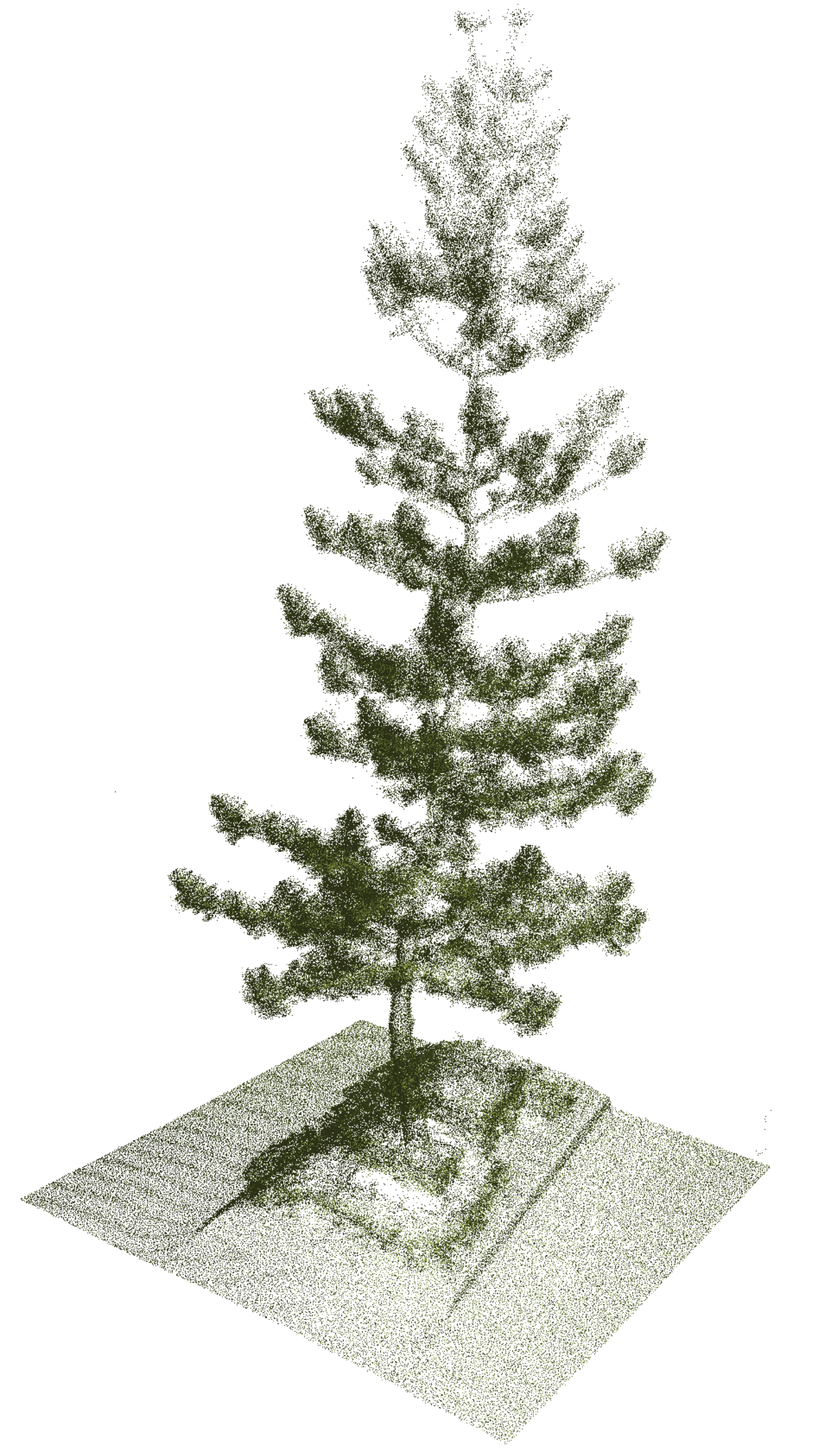}
\label{fig:case1-closeup}}
\subfigure[]{\includegraphics[height=\monkeypicheight cm]{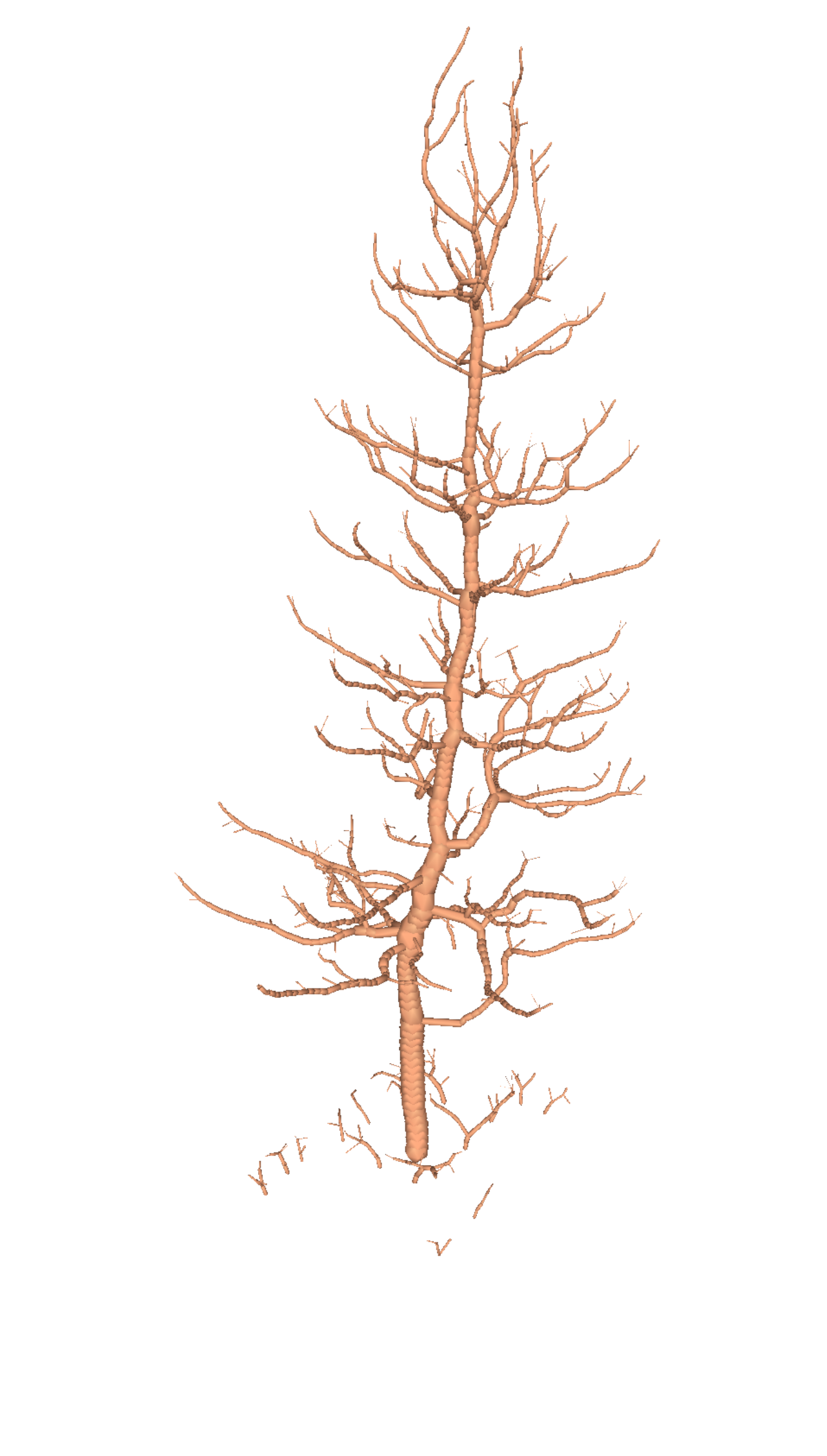}}
\subfigure[]{\includegraphics[height=\monkeypicheight cm]{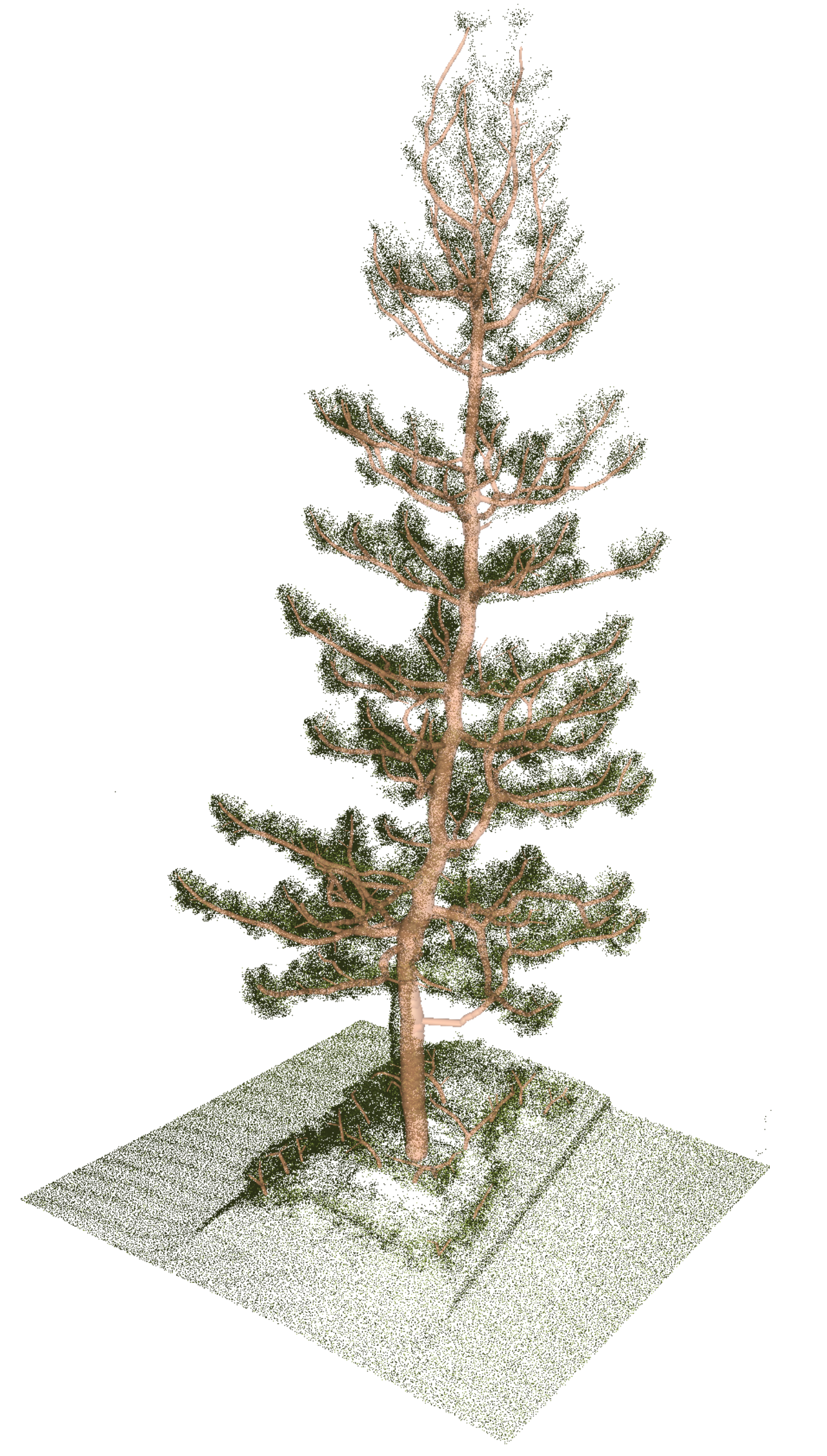}}
\caption{Enlarged view of lone pine tree (case \#1). a) Closeup view of point cloud  b) Reconstructed BSG c) Overlaid view of BSG on point cloud}
\label{fig:case1_enlarged}
\end{adjustwidth}
\end{figure}

The results illustrate the efficacy of each stage of reconstruction, showing the distance between the density grid and the input point cloud, and the reconstructed BSG to the density grid. In the first 5 cases, in which there is minimal occlusion, the BSG reconstructs the observed trees with very high fidelity, capturing both large and small branches.
Notably, it is also possible to see more thin branches in the BSG than in the thresholded density grid, demonstrating that the BSG reconstructs even low density details.  This can be seen in Figure~\ref{fig:case1_enlarged}, which presents an enlarged view of case \#1, its point cloud, BSG, and an overlaid view of the two. We highlight case \#1 because the isolated pine tree is representative of managed plantations, in which tall and narrow trees are spaced on a regular grid, and because its features are clearly visible in the point cloud.

Several noteworthy features can be identified from the figure: firstly the square base in which the tree is planted is effectively removed by the optimisation. Whilst the optimisation treats everything in the raycloud as trees, its attempts to fit trees to the squat shape results in a set of small trunks near the ground. Secondly, the trunk and branches closely follow the visible limbs and have a diameter comparable to that in the point cloud.
However, a few flaws are also noticeable: the lowest branch on the right of the tree emerges from the trunk far below the foliage, and at the top of the tree branches curve upwards erroneously to follow the surface, rather than the branches. These flaws likely arise from the low density of rays below the branches in the tree, the dense foliage at the top of the tree, and setting the branch angle parameter $\alpha$ too low for the tree species.  

In cases \#2, \#3 and \#4 of Table~\ref{tab:Qualitativeresults1}, the denser foliage and close spacing of the Australian Eucalypts prevent TreeSep correctly segmenting them. In comparison, our method is able to both segment and reconstruct the trees to closely match the point clouds. 
In cases \#3 and \#4 part of another tree's canopy is included in the dataset (and to a lesser extent in case \#2). In cases \#2 and \#4 the partial canopy is clearly separated from other trees, and results in a spurious tree being constructed. 
The space where these tree trunks have formed should be filled with rays and therefore $x_i^\mathrm{max}=0$ which would forbid woody material. The reason they form is that the ray cloud has been cropped to a width that is not a multiple of the voxel width $w$, and so the boundary voxels on the six faces of the grid are not fully filled with rays, which in these two cases has allowed sufficient woody material to form along the face to connect to the partial canopy. In case \#3, the partial canopy is close to the foliage of the central tree. The topology optimisation produces a low-density connection to the main tree, which is cropped out during the BSG generation as the thin connection is below the minimum radius.

To explore these features in detail, we encourage readers to view the 3D point clouds and meshes available in the accompanying dataset.

Our method treats every point as part of a tree, and therefore it is clear from cases that include additional objects (cases \#5, \#7, \#9, and lamp posts (\#7, \#14, \#17) that non-tree objects would need to be classified and removed for best results.

Whilst these objects do not cause catastrophic failures, their presence does adversely affect the reconstruction. In the best case, these objects form spurious trees without affecting the real ones, as occurs with the shelter in case \#5 and the lamp post in case \#7. However where the objects are close to, or overlapping with trees, they may become enmeshed in the tree's BSG. This is most obvious in case \#14, in which the lamp post is partially enclosed by the tree and is seen as part of the tree during optimisation. 

It is evident that the algorithm generates tree-like structures that fit to this wide variety of point clouds, even in poorly distinguished foliage such as case \#11. It is also clear from cases \#7, \#8 and \#11 that our method is not reliant on a clearly distinguished trunk, and in case \#6 the trunks are completely occluded by the overpass wall.

A key feature of this algorithm is that it uses the ray information to prevent trunks forming in observed free space. 
 Given a sufficiently high density of rays, the voxel grid is implicitly classified as one of: known free space (where rays have passed without hitting anything), known occupied space (where the rays hit an object) or unknown space (when an occlusion prevents rays reaching the voxel). The optimisation is contrained to prevent material being located where voxels are determined to be free space.  
This is evident in case \#10 where the free-space information prevents a trunk forming despite the canopy sitting very close to the ground (right hand side of the image).

As it does not rely on species specific features, our algorithm is able to reconstruct a broad range of tree types and features. For example in case \#13, codominant stems are captured in the topology optimisation, however during the BSG generation, most branches are joined back to the left stem.
To be clear, whilst the method is capable of reconstructing a broad range of features, it performs best on a dataset with uniform branching angles, a well tuned $\alpha$ and a dense raycloud. 

The segmentation in the TreeSeparation algorithm is shown to approximately segment the main trees for the first four cases. However the remainder of cases are more densely overlapping trees, and the method severely underestimates the number of separate trees in each case. This is because TreeSeparation relies on a thinning or gap between neighbouring trees, which is barely present in cases \#5 up to \#13.

\subsection{Quantitative Comparison: Table~\ref{tab:comparison}}
\label{sec:quantitaive}
In this section we present a quantitative comparison for each case in Tables~\ref{tab:Qualitativeresults1} and \ref{tab:Qualitativeresults2}, firstly as a Surface Error from the input point cloud, and secondly as a comparison of the number of identified trees to the two tree segmentation methods Treeseg~\cite{burt2019extracting} and TreeSeparation~\cite{wang2018scalable}.

The voxel width $w$ for each case is based on the number of voxels $N$ and rounded to the nearest cm. As with any voxelisation of a point cloud, $w$ can be varied, but is most effective when it is at a similar scale to the point density. Larger voxels underfit to the tree geometry, and smaller voxels overfit to gaps between the rays. This effect can be seen in the rightmost tree in case \#4, in which the low ray density allows branches to form inccorectly. The average in these experiments is $w=8.6$ cm which is approximately the point density at the more distant parts of the trees.

Across the 13 cases evaluated, SE averages to 15 cm or approximately two voxel widths. The lowest absolute errors are in the datasets optimised with a small voxel size $w$. However, the lowest relative errors ($SE/w$) occur in the simpler tree cases. That is, the cases with fewer trees, more spacing between trees, an easily identifyable skeleton and minimal occlusions. Cases \#1-5 all have an SE of less than 1.7 voxel widths, with the smallest being 1.2 for cases \#3 and \#4. In constrast, the complex and challenging cases with multiple overlapping trees, dense foliage, and other occlusions have a SE of 2.5-2.8 voxels. In terms of the number of trees produced, our algorithm matches the manual count within 13\%, with our method erring on the side of too many stems. 

While Treeseg and TreeSeparation both produce reasonable numbers of segments for the first five (well-separated) point clouds, TreeSeparation (which relies on gap features between trees) significantly underestimates the number of segments in the densely overlapping cases (\#6 to \#13) and Treeseg (which relies on trunk features) fails to process on the remaining datasets, which generally lack clear trunks. 

\subsection{Ablation Study: Table~\ref{tab:hardcases2}}
This is a qualitative ablation study for comparing results with and without the convolution kernel, $\alpha=1$ and $\alpha=0$ respectively. 

Here we can see the effect of the branch angle parameter $\alpha$ on the reconstruction of heavily occluded trees. 
When $\alpha = 0$ the branches adhere close to the foliage surface in case \#14 as that is the shortest path which joins the observed points to ground. The result is an unnatural looking set of surface branches. However, at $\alpha=1$, the larger branching angle forces branches to occupy the interior of the trees volume. 

In cases \#15, \#16 and \#18 the $\alpha=0$ reconstruction generates a larger number of stems than the $\alpha=1$ reconstruction. This is consistent with the expected behaviour of the branch angle parameter as demonstrated in two dimensions in Figure~\ref{fig:gilbert}.

In case \#18 three trees surround a small shrub in the centre on the image. At $\alpha=1$, trunks and primary branches of the three trees appear correctly located and sized, despite the absence of a central leader branch. However, the smaller shrub is poorly reconstructed. Because of the dense foliage, the reconstruction is unable to separate it from the surrounding trees.

In addition, at both $\alpha=0$ and $\alpha=1$, the presence of a wall in the raycloud causes 'trees' to be generated to fit those points, resulting in the long branches on the left of the figure.

In case \#16 the palm tree on the right of the figure is incorrectly reconstructed for both values of $\alpha$. This suggests that the branching structure of the palm fronds, which radiate from a central point, is inconsistent with the constant branch angle assumption of our method. 

In case \#19 the results for $\alpha = 0$ fit to the ray cloud better, which has long vertical stems on the left hand side. This is consistent with the data in Figure~\ref{fig:gilbert} that show that small $\alpha$ better models vertical stems and narrow branch angles. Therefore best results are expected when $\alpha$ is set appropriately for the expected branch angle.

\subsection{Aerial Comparison: Table~\ref{tab:aerialscan}}
This is a qualitative comparison of a single tree that has been reconstructed from two different angles: a ground-based dense scan and sparse aerial scan.

In both cases a full internal branch structure has been found that fits to the shape of the foliage. However the sparser aerial scan is biased towards a higher up branch structure, reflecting the relatively higher density on the top side of the tree. Also the aerial scan is taken from a 45$^o$ downwards angle, giving a density distribution that favours an off-centre trunk. In both cases the internal branch structure is not observed in the scan.

\subsection{Occlusion Study: Table~\ref{tab:sphereexperiment}}
In this study, the sensitivity of the reconstruction to occlusion is estimated by simulating a spherical occlusion of varying sizes. 
Case \#5 is selected for the study as it is representative of the unoccluded trees whilst containing two close trees, which confounds reconstruction.

An occlusion is simulated by placing spheres at the geometric centre of each of the two trees in the dataset, removing all points from the point cloud which lie within the spheres. The 'occluded' trees are then reconstructed and their accuracy evaluated with varying sphere diameters.

The sphere diameters used are proportional to the trees' height, and range from 1/8th of each tree's height up to 1/2. The diameter of both tree's spheres are increased simultaneously. In this experiment, the optimisation uses 1 million voxels, equating to a voxel size of $23$ cm.

Small occlusions have little effect on the reconstruction, with a sphere 1/8th of the tree's height leading to 3\% increase in error. However, even with an occluding sphere diameter 3/8 of the tree height, which is approximately the diameter of each tree, the results show only a 61\% error increase. Visually, we can see the unconstrained branch angle is greater than for the real tree, so smaller values of $\alpha$ are likely to perform better for this tree species. 

\begin{figure}[h]
\centering
\includegraphics[width=0.6\columnwidth]{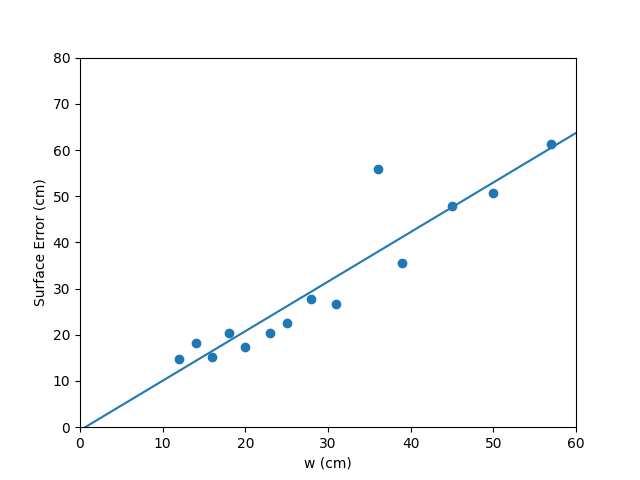}
\caption{Sensitivity of Surface Error to the voxel size $w$ for case \#5 with non-tree objects removed.}
\label{fig:voxel_sensitivity}
\end{figure}

\subsection{Voxel Size Sensitivity: Figure~\ref{fig:voxel_sensitivity}}
\label{sec:voxelsensitivity}
The sensitivity of the reconstructed BSG surface error to the voxel size $w$ is presented in Figure~\ref{fig:voxel_sensitivity}. As one would expect, the error increases roughly in proportion to the voxel size. There is a single outlier due to the BSG reconstruction phase missing a portion of one tree at the large voxel size of $36$ cm. Other than this, the relationship is almost linear, with a best fit line at $SE=1.07w-0.66$ cm. There is an apparent slight levelling off at the smallest voxel sizes, which is expected behaviour when the voxel size becomes too small to be well-represented by lidar points. The levelling appears to occur from 20 cm wide voxels and lower, which corresponds to an average of 8 lidar points per occupied voxel. 

Overall, the results demonstrate a well-behaved trade off between accuracy and solver resolution, with Surface Error approximately proportional to voxel width.

\section{Physical Interpretation of Method}
In this paper we have used topology optimisation for the purely geometric purpose of interpolating lidar points with solid acyclic structures that reach the ground. However Eq~\eqref{eq:formulation} also represents a physical system, where a quantity $f$ distributes isotropically through each uniform element under the steady state equation $\mat{K}\mat{U}=\mat{F}$. It is therefore possible to interpret the optimisation as a number of different physical processes that may be relevant in the formation of real trees:
\begin{itemize}
    \item when $f$ is rate of heat transfer and $u$ is the change in temperature, $E$ represents the thermal conductivity of the material. The optimisation therefore finds the structure that best conducts heat or cold away from the lidar points, this is the thermal model of topology optimisation. 
    \item Eq~\eqref{eq:formulation} could be interpreted as approximating resistance to wind loading. The scalar load $f$ represents an isotropic wind force, assuming the wind occurs equally in all directions over extended time periods. The parameter $E$ is the Young's modulus of the wood and $u$ is the magnitude of position change. It therefore finds the structure that bends and moves least due to long term wind forces. This interpretation is aided by the fact that lidar points represent locations that a distant lidar can observe along straight rays. Which corresponds quite well with the set of locations that wind acts upon, namely the tree outer surfaces.
    \item in high foliage scans Eq~\eqref{eq:formulation} could approximate resistance to water loss. The load $f$ represents the tensile force on water in the tree xylem due to transpiration at the leaf. The stiffness represents how easily water can be transported from the roots to the branches, and $u$ represents the change in water content. It therefore finds the structure that minimises water loss from transpiration.  
\end{itemize}
There are other similar approximations such as optimising the transport of sugars to the roots. In each case the physical process represents some form of load or effort at the lidar points that must be conducted efficiently to the ground. 

The factors that affect tree structure remains an open topic of research, but we know that competition for light and resources favours trees that distribute their material efficiently, and this is a form of optimisation. Previously researched factors include wind load~\cite{eloy2017wind,eloy2011leonardo}, efficiency and safety of hydraulic transport~\cite{midgley2003bigger}, and others such as stress minimisation in trees~\cite{mattheck2004mechanical}. 

It is useful to consider such botanical interpretations because any future research that clarifies the optimising factors in tree structure can be used to extend the formulation in Eq~\eqref{eq:formulation}, and so provide more accurate interpolation of branches in unobserved regions.

\section{Limitations}
The presented method has three limitations that remain to be addressed. Firstly, the total volume of the structure must be estimated a priori. We use the volume of occupied voxels as an estimator for this value but its estimation is still crude. Further research is required to better adapt the volume of material to the dataset being used.   

Secondly, setting nodal load in proportion to number of lidar points is biased towards areas that are scanned for longer, or are observed at closer range. This bias is evident in the difference between the same tree, scanned in Table~\ref{tab:aerialscan} from different directions. It is also evident in case 19, where the higher density of points on the right hand side cause the rightmost tree to `capture' the canopy of all the trees. A simple improvement is to spatially subsample the ray cloud, and a more advanced technique could normalise the nodal loads to the mean load within a local neighbourhood. 

Lastly, the four assumptions of our method in Section~\ref{sec:introduction} are also forms of limitation. If clouds contain non-tree structures, cyclic structures, or trees with unobserved branches that are winding or at varying branch angles, then the method will not reconstruct these structures as they are. Instead, it will generate the tree structures that best fit its optimisation model.

It would be an instructive topic of future research to assess the severity of errors in the reconstruction due to data that breaks the above assumptions, and to assess the effect of improved nodal load distributions. 

The focus on this paper has been on the method itself rather than its computational efficiency. We would therefore like to investigate the most optimal solver methods, software and parameters for solving Eq~\eqref{eq:formulation} as our next topic of research. 

\section{Conclusion}
Methods of tree reconstruction in vegetated environments are hampered by the extreme variability of input data, and the lack of consistent features on which to segment trees or branches. Consequently, the majority of existing reconstruction techniques are limited to constrained environments, and suffer from catastrophic failures where the relied-on features are missing. 

We have taken a radically different approach, avoiding feature-based reconstruction entirely, and instead treating the tree as the structure that matches the observed data while minimising compliance to loads. The results demonstrate the resilience of this simple minimisation rule under a wide range of datasets. 

Our results suggest that topological optimisation is a valuable method for robust tree reconstruction, and deserves a prominent role in this field. They suggest that the research question should not focus on what are the features of a tree, but on what a tree optimises.

\blfootnote{This research was supported by Victorian Government's Powerline Bushfire Safety Program R\&D fund, Powercor, Sylvanus and CSIRO Data61. \\

The datasets generated and used in the current study are available in the `Tree Reconstructions from Pointclouds Scanned in Pullenvale QLD' repository \\ \url{https://doi.org/10.25919/yt2m-9373}}

\newpage
\bibliography{bibliography.bib}

\begin{thebibliography}{10}

\bibitem{trochta20173d}
J.~Trochta, M.~Krucek, T.~Vrska, and K.~Kral, ``3d forest: An application for
  descriptions of three-dimensional forest structures using terrestrial
  lidar,'' {\em PloS one}, vol.~12, no.~5, p.~e0176871, 2017.

\bibitem{aiteanu2014hybrid}
F.~Aiteanu and R.~Klein, ``Hybrid tree reconstruction from inhomogeneous point
  clouds,'' {\em The Visual Computer}, vol.~30, no.~6, pp.~763--771, 2014.

\bibitem{du2019adtree}
S.~Du, R.~Lindenbergh, H.~Ledoux, J.~Stoter, and L.~Nan, ``Adtree: Accurate,
  detailed, and automatic modelling of laser-scanned trees,'' {\em Remote
  Sensing}, vol.~11, no.~18, p.~2074, 2019.

\bibitem{livny2010automatic}
Y.~Livny, F.~Yan, M.~Olson, B.~Chen, H.~Zhang, and J.~El-Sana, ``Automatic
  reconstruction of tree skeletal structures from point clouds,'' in {\em ACM
  SIGGRAPH Asia 2010 papers}, pp.~1--8, 2010.

\bibitem{li2010segmentation}
H.~Li, X.~Zhang, M.~Jaeger, and T.~Constant, ``Segmentation of forest terrain
  laser scan data,'' in {\em Proceedings of the 9th ACM SIGGRAPH Conference on
  Virtual-Reality Continuum and its Applications in Industry}, pp.~47--54,
  2010.

\bibitem{zhong2016segmentation}
L.~Zhong, L.~Cheng, H.~Xu, Y.~Wu, Y.~Chen, and M.~Li, ``Segmentation of
  individual trees from tls and mls data,'' {\em IEEE Journal of Selected
  Topics in Applied Earth Observations and Remote Sensing}, vol.~10, no.~2,
  pp.~774--787, 2016.

\bibitem{liu2021point}
Q.~Liu, W.~Ma, J.~Zhang, Y.~Liu, D.~Xu, and J.~Wang, ``Point-cloud segmentation
  of individual trees in complex natural forest scenes based on a trunk-growth
  method,'' {\em Journal of Forestry Research}, pp.~1--12, 2021.

\bibitem{burt2019extracting}
A.~Burt, M.~Disney, and K.~Calders, ``Extracting individual trees from lidar
  point clouds using treeseg,'' {\em Methods in Ecology and Evolution},
  vol.~10, no.~3, pp.~438--445, 2019.

\bibitem{li2012new}
W.~Li, Q.~Guo, M.~K. Jakubowski, and M.~Kelly, ``A new method for segmenting
  individual trees from the lidar point cloud,'' {\em Photogrammetric
  Engineering \& Remote Sensing}, vol.~78, no.~1, pp.~75--84, 2012.

\bibitem{ayrey2017layer}
E.~Ayrey, S.~Fraver, J.~A. Kershaw~Jr, L.~S. Kenefic, D.~Hayes, A.~R.
  Weiskittel, and B.~E. Roth, ``Layer stacking: a novel algorithm for
  individual forest tree segmentation from lidar point clouds,'' {\em Canadian
  Journal of Remote Sensing}, vol.~43, no.~1, pp.~16--27, 2017.

\bibitem{heinzel2018constrained}
J.~Heinzel and M.~O. Huber, ``Constrained spectral clustering of individual
  trees in dense forest using terrestrial laser scanning data,'' {\em Remote
  Sensing}, vol.~10, no.~7, p.~1056, 2018.

\bibitem{wang2018scalable}
J.~Wang, R.~Lindenbergh, and M.~Menenti, ``Scalable individual tree delineation
  in 3d point clouds,'' {\em The Photogrammetric Record}, vol.~33, no.~163,
  pp.~315--340, 2018.

\bibitem{luo2021individual}
H.~Luo, K.~Khoshelham, C.~Chen, and H.~He, ``Individual tree extraction from
  urban mobile laser scanning point clouds using deep pointwise direction
  embedding,'' {\em ISPRS Journal of Photogrammetry and Remote Sensing},
  vol.~175, pp.~326--339, 2021.

\bibitem{Krisanski2021}
S.~Krisanski, M.~S. Taskhiri, S.~G. Aracil, D.~Herries, and P.~Turner,
  ``{Sensor agnostic semantic segmentation of structurally diverse and complex
  forest point clouds using deep learning},'' {\em Remote Sensing}, vol.~13,
  no.~8, 2021.

\bibitem{tagliasacchi20163d}
A.~Tagliasacchi, T.~Delame, M.~Spagnuolo, N.~Amenta, and A.~Telea, ``3d
  skeletons: A state-of-the-art report,'' in {\em Computer Graphics Forum},
  vol.~35, pp.~573--597, Wiley Online Library, 2016.

\bibitem{schilling2014automatic}
A.~Schilling and H.-G. Maas, ``Automatic reconstruction of skeletal structures
  from tls forest scenes.,'' {\em ISPRS Annals of Photogrammetry, Remote
  Sensing \& Spatial Information Sciences}, vol.~2, no.~5, 2014.

\bibitem{jiang2020skeleton}
A.~Jiang, J.~Liu, J.~Zhou, and M.~Zhang, ``Skeleton extraction from point
  clouds of trees with complex branches via graph contraction,'' {\em The
  Visual Computer}, pp.~1--17, 2020.

\bibitem{raumonen2013fast}
P.~Raumonen, M.~Kaasalainen, M.~{\AA}kerblom, S.~Kaasalainen, H.~Kaartinen,
  M.~Vastaranta, M.~Holopainen, M.~Disney, and P.~Lewis, ``Fast automatic
  precision tree models from terrestrial laser scanner data,'' {\em Remote
  Sensing}, vol.~5, no.~2, pp.~491--520, 2013.

\bibitem{wang2018lidar}
R.~Wang, J.~Peethambaran, and D.~Chen, ``Lidar point clouds to 3-d urban models
  $: $ a review,'' {\em IEEE Journal of Selected Topics in Applied Earth
  Observations and Remote Sensing}, vol.~11, no.~2, pp.~606--627, 2018.

\bibitem{hu2017efficient}
S.~Hu, Z.~Li, Z.~Zhang, D.~He, and M.~Wimmer, ``Efficient tree modeling from
  airborne lidar point clouds,'' {\em Computers \& Graphics}, vol.~67,
  pp.~1--13, 2017.

\bibitem{xu2007knowledge}
H.~Xu, N.~Gossett, and B.~Chen, ``Knowledge and heuristic-based modeling of
  laser-scanned trees,'' {\em ACM Transactions on Graphics (TOG)}, vol.~26,
  no.~4, pp.~19--es, 2007.

\bibitem{treeQSM}
D.~P. Raumonen, ``Treeqsm.'' \url{https://github.com/InverseTampere/TreeQSM},
  2017.

\bibitem{fan2020adqsm}
G.~Fan, L.~Nan, Y.~Dong, X.~Su, and F.~Chen, ``Adqsm: A new method for
  estimating above-ground biomass from tls point clouds,'' {\em Remote
  Sensing}, vol.~12, no.~18, p.~3089, 2020.

\bibitem{Ikonen2018b}
T.~J. Ikonen, G.~Marck, A.~S{\'{o}}bester, and A.~J. Keane, ``{Topology
  optimization of conductive heat transfer problems using parametric
  L-systems},'' {\em Structural and Multidisciplinary Optimization}, vol.~58,
  no.~5, pp.~1899--1916, 2018.

\bibitem{Pinskier2020}
J.~Pinskier, B.~Shirinzadeh, M.~Ghafarian, T.~K. Das, A.~Al-Jodah, and
  R.~Nowell, ``{Topology optimization of stiffness constrained flexure-hinges
  for precision and range maximization},'' {\em Mechanism and Machine Theory},
  vol.~150, p.~103874, 2020.

\bibitem{Pinskier2018b}
J.~Pinskier and B.~Shirinzadeh, ``{Topology optimization of leaf flexures to
  maximize in-plane to out-of-plane compliance ratio},'' {\em Precision
  Engineering}, vol.~55, pp.~397--407, 2019.

\bibitem{Clark2018}
L.~Clark, B.~Shirinzadeh, J.~Pinskier, Y.~Tian, and D.~Zhang, ``{Topology
  optimisation of bridge input structures with maximal amplification for design
  of flexure mechanisms},'' {\em Mechanism and Machine Theory}, vol.~122,
  pp.~113--131, 2018.

\bibitem{Pinskier2022}
J.~Pinskier and D.~Howard, ``{From Bioinspiration to Computer Generation:
  Developments in Autonomous Soft Robot Design},'' {\em Advanced Intelligent
  Systems}, vol.~4, no.~1, p.~2100086, 2022.

\bibitem{Zhang2018f}
H.~Zhang, A.~S. Kumar, J.~Y.~H. Fuh, and M.~Y. Wang, ``{Design and Development
  of a Topology-Optimized Three-Dimensional Printed Soft Gripper},'' {\em Soft
  Robotics}, vol.~5, no.~5, pp.~650--661, 2018.

\bibitem{Sigmund2001}
O.~Sigmund, ``{Design of multiphysics actuators using topology optimization -
  Part I: One-material structures},'' {\em Computer Methods in Applied
  Mechanics and Engineering}, vol.~190, no.~49-50, pp.~6605--6627, 2001.

\bibitem{Kumar2020}
P.~Kumar, J.~S. Frouws, and M.~Langelaar, ``{Topology optimization of fluidic
  pressure-loaded structures and compliant mechanisms using the Darcy
  method},'' {\em Structural and Multidisciplinary Optimization}, vol.~61,
  no.~4, pp.~1637--1655, 2020.

\bibitem{Munk2017}
D.~J. Munk, T.~Kipouros, G.~A. Vio, G.~P. Steven, and G.~T. Parks, ``{Topology
  optimisation of micro fluidic mixers considering fluid-structure interactions
  with a coupled Lattice Boltzmann algorithm},'' {\em Journal of Computational
  Physics}, vol.~349, pp.~11--32, 2017.

\bibitem{Zhang2022a}
H.~K. Zhang, J.~Zhou, W.~Fang, H.~Zhao, Z.~L. Zhao, X.~Chen, H.~P. Zhao, and
  X.~Q. Feng, ``{Multi-functional topology optimization of Victoria cruziana
  veins},'' {\em Journal of the Royal Society Interface}, vol.~19, no.~191,
  2022.

\bibitem{Bendsoe2003}
M.~P. Bends{\o}e and O.~Sigmund, {\em {Topology optimization: theory, methods,
  and applications}}.
\newblock Springer, 2003.

\bibitem{Zhang2018e}
X.~Zhang and B.~Zhu, {\em {Topology Optimization of Compliant Mechanisms}}.
\newblock Springer, 2018.

\bibitem{Sigmund2013}
O.~Sigmund and K.~Maute, ``{Topology optimization approaches: A comparative
  review},'' {\em Structural and Multidisciplinary Optimization}, vol.~48,
  no.~6, pp.~1031--1055, 2013.

\bibitem{Munk2015}
D.~J. Munk, G.~A. Vio, and G.~P. Steven, ``{Topology and shape optimization
  methods using evolutionary algorithms: a review},'' {\em Structural and
  Multidisciplinary Optimization}, vol.~52, no.~3, pp.~613--631, 2015.

\bibitem{VanDijk2013}
N.~P. {Van Dijk}, K.~Maute, M.~Langelaar, and F.~{Van Keulen}, ``{Level-set
  methods for structural topology optimization: A review},'' {\em Structural
  and Multidisciplinary Optimization}, vol.~48, no.~3, pp.~437--472, 2013.

\bibitem{Pansfem2019}
T.~Yuta, ``Pansfem.'' \url{https://github.com/PANFACTORY/PANSFEM2}, 2019.

\bibitem{andreassen2011efficient}
E.~Andreassen, A.~Clausen, M.~Schevenels, B.~S. Lazarov, and O.~Sigmund,
  ``Efficient topology optimization in matlab using 88 lines of code,'' {\em
  Structural and Multidisciplinary Optimization}, vol.~43, no.~1, pp.~1--16,
  2011.

\bibitem{Guest2004}
J.~K. Guest, J.~H. Pr{\'{e}}vost, and T.~Belytschko, ``{Achieving minimum
  length scale in topology optimization using nodal design variables and
  projection functions},'' {\em International Journal for Numerical Methods in
  Engineering}, vol.~61, no.~2, pp.~238--254, 2004.

\bibitem{Liu2021}
C.-H. Liu, Y.~Chen, and S.-Y. Yang, ``{Topology Optimization and Prototype of a
  Multimaterial-Like Compliant Finger by Varying the Infill Density in 3D
  Printing},'' {\em Soft Robotics}, no.~October, 2021.

\bibitem{Fleury1989}
C.~Fleury, ``{CONLIN: An efficient dual optimizer based on convex approximation
  concepts},'' {\em Structural Optimization}, vol.~1, no.~2, pp.~81--89, 1989.

\bibitem{yan2018non}
S.~Yan, F.~Wang, and O.~Sigmund, ``On the non-optimality of tree structures for
  heat conduction,'' {\em International Journal of Heat and Mass Transfer},
  vol.~122, pp.~660--680, 2018.

\bibitem{lowe2021raycloudtools}
T.~Lowe and K.~Stepanas, ``Raycloudtools: A concise interface for analysis and
  manipulation of ray clouds,'' {\em IEEE Access}, 2021.

\bibitem{calders2015nondestructive}
K.~Calders, G.~Newnham, A.~Burt, S.~Murphy, P.~Raumonen, M.~Herold,
  D.~Culvenor, V.~Avitabile, M.~Disney, J.~Armston, {\em et~al.},
  ``Nondestructive estimates of above-ground biomass using terrestrial laser
  scanning,'' {\em Methods in Ecology and Evolution}, vol.~6, no.~2,
  pp.~198--208, 2015.

\bibitem{dijkstra1959note}
E.~W. Dijkstra {\em et~al.}, ``A note on two problems in connexion with
  graphs,'' {\em Numerische mathematik}, vol.~1, no.~1, pp.~269--271, 1959.

\bibitem{Ramezani2022}
M.~Ramezani, K.~Khosoussi, G.~Catt, P.~Moghadam, J.~Williams, P.~Borges,
  F.~Pauling, and N.~Kottege, ``{Wildcat: Online Continuous-Time 3D
  Lidar-Inertial SLAM},'' vol.~XX, pp.~1--13, 2022.

\bibitem{Jinhu-Wang}
J.~Wang, ``Treeseparation.''
  \url{https://github.com/Jinhu-Wang/TreeSeparation}, 2020.

\bibitem{andrew_burt_2021_4884923}
A.~Burt, T.~Peter, and jgrn307, ``apburt/treeseg: v0.2.2,'' may 2021.

\bibitem{westling2021simtreels}
F.~Westling, M.~Bryson, and J.~Underwood, ``Simtreels: Simulating aerial and
  terrestrial laser scans of trees,'' {\em Computers and Electronics in
  Agriculture}, vol.~187, p.~106277, 2021.

\bibitem{esmoris2022virtual}
A.~M. Esmor{\'\i}s, M.~Yermo, H.~Weiser, L.~Winiwarter, B.~H{\"o}fle, and F.~F.
  Rivera, ``Virtual lidar simulation as a high performance computing challenge:
  Toward hpc helios++,'' {\em IEEE Access}, vol.~10, pp.~105052--105073, 2022.

\bibitem{eloy2017wind}
C.~Eloy, M.~Fournier, A.~Lacointe, and B.~Moulia, ``Wind loads and competition
  for light sculpt trees into self-similar structures,'' {\em Nature
  communications}, vol.~8, no.~1, pp.~1--12, 2017.

\bibitem{eloy2011leonardo}
C.~Eloy, ``Leonardo’s rule, self-similarity, and wind-induced stresses in
  trees,'' {\em Physical review letters}, vol.~107, no.~25, p.~258101, 2011.

\bibitem{midgley2003bigger}
J.~J. Midgley, ``Is bigger better in plants? the hydraulic costs of increasing
  size in trees,'' {\em Trends in Ecology \& Evolution}, vol.~18, no.~1,
  pp.~5--6, 2003.

\bibitem{mattheck2004mechanical}
C.~Mattheck and I.~Tesari, ``The mechanical self-optimisation of trees,'' {\em
  WIT Transactions on Ecology and the Environment}, vol.~73, 2004.

\end{thebibliography}
\bibliographystyle{ieeetr}

\newpage
\appendix
\section{Sensitivity}\label{appendix:sensitivity}
From Eq~\eqref{eq:formulation} the compliance for the smoothed densities $\bar{\vec{x}}$ can be expressed as a sum over all elements $i$: 
\begin{align}
c(\bar{\vec{x}})=\sum_i \tilde{E}\vec{u}^T_i \vec{k}_i \vec{u}_i
\end{align}
where $\tilde{E}$ is defined in Eq~\eqref{eq:multiplier} as the product of two terms. Using the product rule, the differential with respect to each smoothed element density $\bar{x}_i$ can then be written as the sum of two terms:
\begin{align}
\dfrac{\partial c(\bar{\vec{x}})}{\partial \bar{x}_i}=a_i\vec{u}^T_i \vec{k}_i \vec{u}_i + \sum_j b_{ij} \vec{u}^T_j \vec{k}_j \vec{u}_j
\end{align}
where 
\begin{align}
a_i = -A(\bar{x}_i)^\alpha p\bar{x}_i^{p-1}(E_{S}-E_{V})
\end{align}
and 
\begin{align}
b_{ij} = -\dfrac{\alpha A(\bar{x}_j)^{\alpha-1}\left(E_{V}+\bar{x}_j^p(E_{S}-E_{V})\right)}{\mathrm{max}(r_{ij},1)}
\end{align}
The sensitivity with respect to the unsmoothed design variable $\vec{x}$ is then:
\begin{align}
\dfrac{\partial c(\bar{\vec{x}})}{\partial x_i}=\dfrac{\sum_{j\in N_e}W_{ij}\hat{x}_i(\beta)\dfrac{\partial c(\bar{\vec{x}})}{\partial \bar{x}_j}}{\sum_{j \in N_e} W_{ij}}
\end{align}
where 
\begin{equation}
\hat{x}_i(\beta)=\beta\dfrac{1-\tanh^2({\beta(\tilde{x_i}-0.5}))}
{2\tanh{(0.5\beta)}}
\end{equation}
\end{document}